%% file: main.tex
\documentclass[runningheads]{llncs}

\usepackage[mobile]{eccv}

\usepackage{eccvabbrv}

\usepackage{graphicx}
\usepackage{booktabs}

\usepackage[accsupp]{axessibility}  % Improves PDF readability for those with disabilities.

\usepackage{hyperref}

\usepackage{orcidlink}

\input{preamble}
\begin{document}

% ---------------------------------------------------------------
% TODO REVIEW: Replace with your title
%\title{\ours: Interpretable Semantic Hierarchies in Vision-Language Encoders} 
\title{CFM: Language-aligned Concept Foundation Model for Vision}

% TODO REVIEW: If the paper title is too long for the running head, you can set
% an abbreviated paper title here. If not, comment out.
% \titlerunning{Abbreviated paper title}

% TODO FINAL: Replace with your author list. 
% Include the authors' OCRID for the camera-ready version, if at all possible.
\author{Kai Wittenmayer\orcidlink{0009-0002-0521-8466} \and
Sukrut Rao$^\dagger$\orcidlink{0000-0001-8896-7619} \and
Amin Parchami-Araghi$^\dagger$\orcidlink{0000-0003-0424-7812}\and \\
Bernt Schiele\orcidlink{0000-0001-9683-5237} \and
Jonas Fischer\orcidlink{0000-0002-6459-5053}}

% TODO FINAL: Replace with an abbreviated list of authors.
\authorrunning{K.~Wittenmayer et al.}
% First names are abbreviated in the running head.
% If there are more than two authors, 'et al.' is used.

% TODO FINAL: Replace with your institution list.
\institute{Max Planck Institute for Informatics, Saarland Informatics Campus \\ Saarbrücken, Germany \\
\email{\{kai.wittenmayer,sukrut.rao,amin.parchami\}@mpi-inf.mpg.de}\\
\email{\{schiele,jonas.fischer\}@mpi-inf.mpg.de}
}

\maketitle

\begin{center}
    \vspace{-10pt}
    \scriptsize $\dagger$~Equal contribution
\end{center}

\begin{center}
    \captionsetup{type=figure}
    \includegraphics[width=\textwidth, trim={5cm 6.5cm 4cm 6cm}, clip]{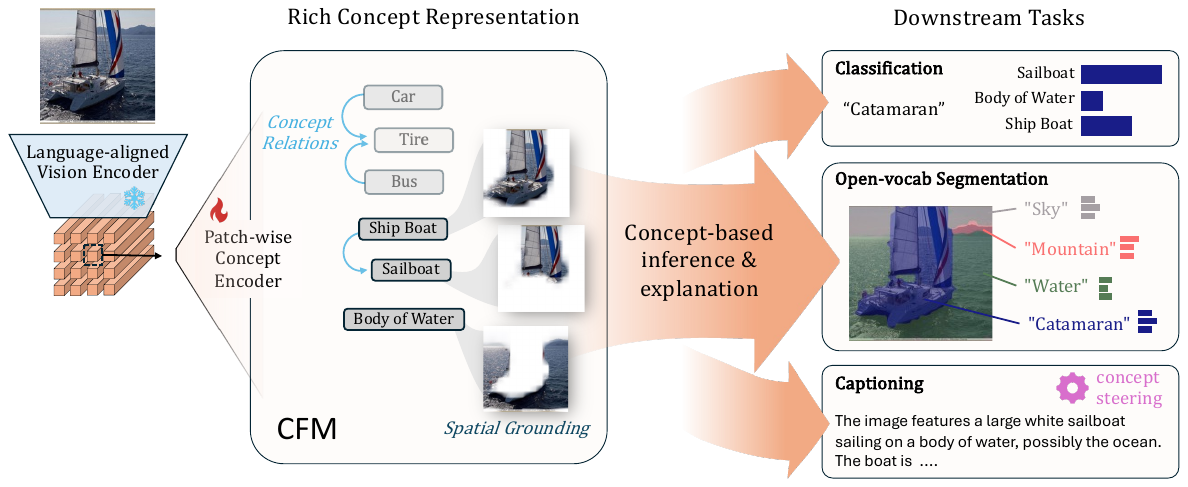}
    % \fbox{\rule{0pt}{2.5in} \rule{.95\linewidth}{0pt}}
    \captionof{figure}{\textbf{\ours provides rich conceptual explanations for vision foundation model tasks.} Our model provides a \textbf{hierarchical concept representation} (\eg `Sailboat' concept is a child of `Ship Boat') with \textbf{local, spatially grounded} concepts that are automatically \textbf{named}. These concepts enable \ours to provide concept-based explanations for decision-making across vision tasks.
    \label{fig:teaser}} 
\end{center}

\input{sec/0_abstract} 
\input{sec/1_intro}

\input{sec/2_related}
\input{sec/3_method}
\input{sec/5_results}
\input{sec/6_conclusion}

% \clearpage

\section*{Acknowledgements}
Sukrut Rao and Bernt Schiele were funded in part by the Deutsche Forschungsgemeinschaft (DFG, German Research Foundation) -- GRK 2853/1 “Neuroexplicit Models of Language, Vision, and Action” - project number 471607914.

\clearpage

% ---- Bibliography ----
%
% BibTeX users should specify bibliography style 'splncs04'.
% References will then be sorted and formatted in the correct style.
%
\bibliographystyle{splncs04}
\bibliography{main}

\clearpage

\appendix
\input{sec/X_suppl}

\end{document}

%% file: preamble.tex
\usepackage{xspace}
\newcommand{\ours}{\textsc{CFM}\xspace}

\usepackage[table]{xcolor}
\usepackage{amssymb}
\usepackage{dsfont}
\usepackage{pifont}
\usepackage{svg}
\newcommand{\dncbm}{DN-CBM\xspace}
\newcommand{\msae}{MSAE\xspace}
\newcommand{\patchsae}{PatchSAE\xspace}
\newcommand{\salfcbm}{SALF-CBM\xspace}
\newcommand{\cfcbm}{CF-CBM\xspace}

\newcommand{\dcbm}{D-CBM\xspace}
\usepackage{pifont} 
\newcommand{\cmarkg}{\textcolor{green!60!black}{\ding{51}}}  % green check
\newcommand{\xmarkr}{\textcolor{red}{\ding{55}}}             % red cross
\newcommand{\ymarko}{\textcolor{orange}{\textbullet}}        % 

\usepackage{etoolbox}
\newtoggle{comment}
\toggletrue{comment}
\newcommand{\myparagraph}[1]{\vspace{3pt}\noindent{\bf #1}}

\usepackage{changepage} % Used for the appendix

\usepackage{multirow}
\usepackage{enumitem}
\usepackage{fvextra}

%% file: sec/0_abstract.tex
\begin{abstract}
Language-aligned vision foundation models perform strongly across diverse downstream tasks. Yet, their learned representations remain opaque, making interpreting their decision-making difficult. 
Recent work decompose these representations into human-interpretable concepts, but provide poor spatial grounding and are limited to image classification tasks.
In this work, we propose \ours, a \textit{language-aligned concept foundation model for vision} that provides fine-grained concepts, which are human-interpretable and spatially grounded in the input image.
When paired with a foundation model with strong semantic representations, we get explanations for \textit{any of its downstream tasks}. Examining local co-occurrence dependencies of concepts allows us to define concept relationships through which we improve concept naming and obtain richer explanations.
On benchmark data, we show that \ours provides performance on classification, segmentation, and captioning that is competitive with opaque foundation models while providing fine-grained, high quality concept-based explanations. Code at \href{https://github.com/kawi19/CFM} {github.com/kawi19/CFM}. Interactive visualizations at \href{https://concept-foundation-model.mpi-inf.mpg.de}{concept-foundation-model.mpi-inf.mpg.de}.
\end{abstract}

%% file: sec/1_intro.tex
\section{Introduction}
\label{sec:intro}

Language-aligned vision foundation models such as CLIP~\cite{radford2021learning,zhai2023sigmoid,tschannen2025siglip} serve as powerful feature extractors that perform strongly across tasks such as zero-shot classification and open-vocabulary semantic segmentation.
Such models also serve as vision encoders for large vision-language models~\cite{chen2024internvl,liu2024improved,li2025llava} and generative models~\cite{rae}. However, their representations are opaque, which hinders interpreting their decision-making. Interpretability is, however, key for human users, especially for safety critical applications, and explanations of decision-making can help to understand whether a model is right for the right reasons, increase trust in a model's prediction, investigate why a model failed, and steer it away from harmful outputs.

Concept Bottleneck Models have emerged as a widely successful approach to make model predictions more interpretable,
with recent works~\cite{zaigrajew2025interpreting,bhalla2024interpreting,rao2024discover,lim2024sparse,benou2025show} aiming to decompose representations of foundation models into human-interpretable concepts and achieve transparency at an unprecedented scale.
However, extracted concepts are not often properly grounded in the input image~\cite{zaigrajew2025interpreting,bhalla2024interpreting,rao2024discover}, harming interpretability. %
Moreover, the existing approaches only provide a limited range of granularity of concepts~\cite{rao2024discover,bhalla2024interpreting,lim2024sparse,benou2025show} and their spatial coverage~\cite{rao2024discover,bhalla2024interpreting,zaigrajew2025interpreting}, and inter-concept dependencies are not examined even when using hierarchical priors~\cite{zaigrajew2025interpreting}. Textual concept labels, when assigned~\cite{rao2024discover,bhalla2024interpreting}, often stem from a limited vocabulary that is not tailored for large-scale and fine-grained conceptual diversity.

Besides these limitations of what concepts can capture, such models have not been used in the foundational setting, i.e. such concept representations have only been used for global tasks, typically classification.
The lack of granular, well-grounded, and non-spurious concepts hinders application to local downstream tasks, such as segmentation or large vision-language model reasoning,
whereas the underlying foundation model backbone would be capable of this.

In this work, we address these issues and propose \ours, a \textit{language-aligned concept foundation model} that provides a unified concept representation applicable to local image regions, thus \textit{enabling any downstream vision foundation model task previously not possible with concept-based explanations}. The discovered concepts are at varying granularities, spatially grounded in the image, and organized into hierarchies with explicit inter-concept dependencies. To achieve this we employ a semantically smoothened image encoding through DINOised CLIP features~\cite{wysoczanska2024clipdinoiser} and learn a sparse autoencoder with a matryoshka objective encouraging hierarchical structures of the concept representations~\cite{zaigrajew2025interpreting,bussmann2025learning}. By learning the concept representation across local image regions (e.g., patches) we achieve spatially localized and human-aligned concepts from coarse to fine granularities and avoid spurious concepts common in the existing global concept bottlenecks. By examining co-occurrences of local concept activations we discover parent-child relationships between concepts that provide further insights into the learned semantics. We use these relations to refine concept labeling towards more accurate and meaningful textual concept labels. Our model enables interpretable decision-making across foundation model downstream tasks typical for language-aligned vision foundation models (see \cref{fig:teaser}).
Our contributions can be summarized as follows:
\begin{enumerate}
    \item We propose a \textbf{concept foundation model} (\ours), lifting concept bottleneck models to language-aligned vision foundation models \textbf{applicable to any downstream task} beyond just image classification through unified local concept representations. 
    \item We show how to extract \textbf{explicit hierarchical relations} between foundational concepts that further improve the explanations of model decisions.
    \item We propose an \textbf{improved concept labeling scheme} that explicitly takes these concept relationships into account.
    \item Putting this together, \ours yields \textbf{human-interpretable concepts} at \textbf{diverse granularities} that are \textbf{spatially localized and well-grounded} in the input.
\end{enumerate}

On benchmark data and through a human user study we show that in contrast to state-of-the-art, \ours yields concepts that are more quantitatively and qualitatively consistent, spatially more localized (\cref{fig:interpret_local}), well organized into hierarchies (\cref{fig:qualfamily}), and meaningfully named (\cref{fig:name_accuracy}). %
We further show that this added interpretability does not come with a degradation of performance across various downstream tasks.
On image classification tasks \ours yields performances competitive with the state-of the-art concept-based and opaque baseline models (\cref{fig:classification}). On an
image captioning task, \ours performs similar to its opaque backbone and enables steering (\cref{fig:captioning}).
\ours also enables interpretable open-vocabulary segmentation that is on par with opaque foundation model alternatives (\cref{table:main_bg}, \cref{fig:qualSegmentation}).

%% file: sec/2_related.tex
\begin{figure*}[!t]
\includegraphics[width=\textwidth, trim={0cm 3.8cm 2cm 2.4cm}, clip]{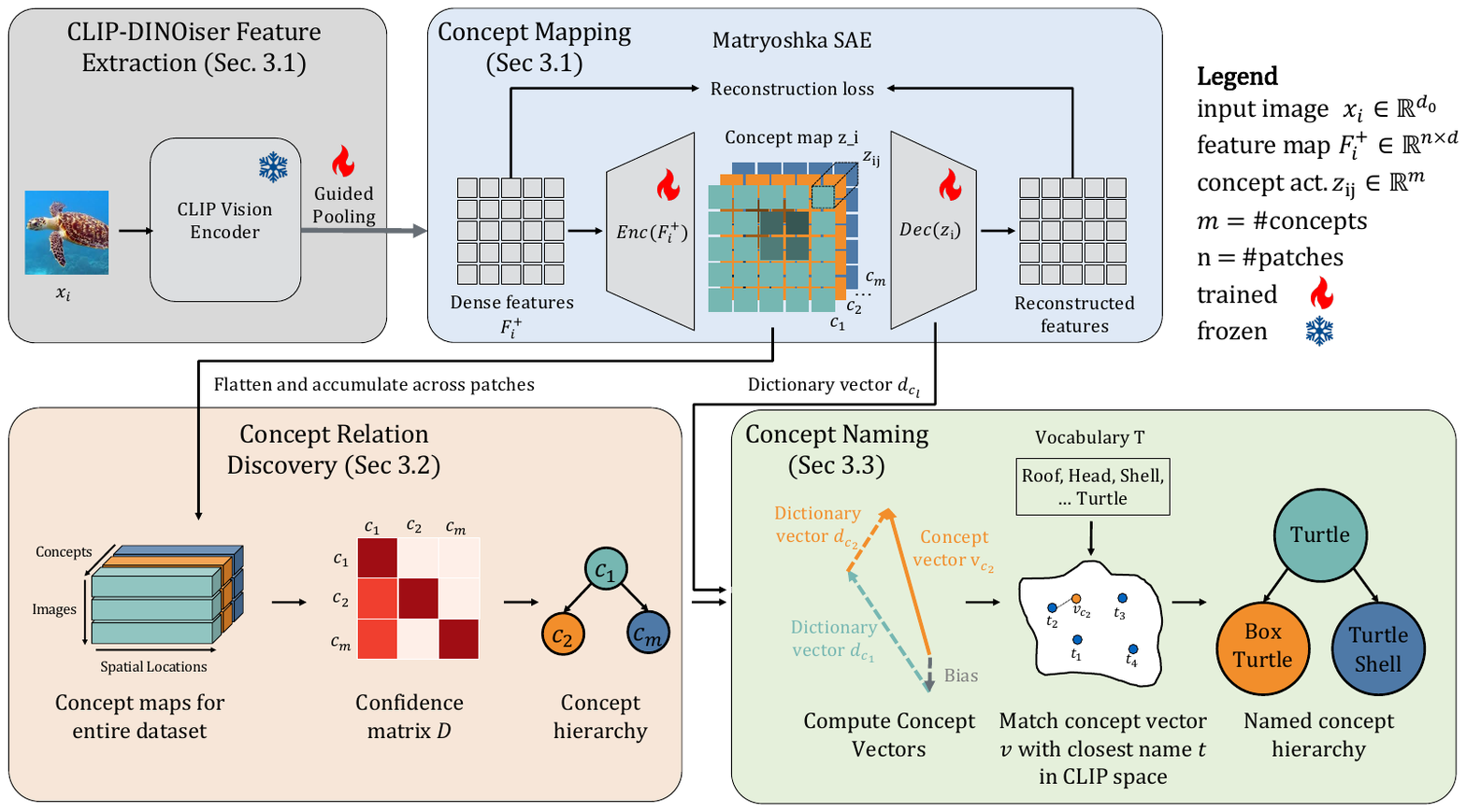}
\caption{\textbf{Overview of \ours.} We visualize our approach to represent human-interpretable, well localized, and consistent concepts of different granularities in a language-aligned vision foundation model. As backbone, we use pretrained CLIP model, train a DINOising~\cite{wysoczanska2024clipdinoiser} layer on top to obtain semantically smooth patch features and extract concepts patch-wise by training a Matryoshka Sparse Autoencoder (SAE) without any supervision. From these learned concept representations, we discover concept relations through examining the co-occurrences of concepts within patches that allow more fine-grained interpretation of decision-making on downstream tasks. We further name the concepts with a Matryoshka-aware matching between concept representations and a large vocabulary of labels, taking the discovered concept relations into account.}\label{fig:method}%
\end{figure*}

\section{Related Work}
\label{sec:related}

\myparagraph{Vision-language models (VLMs)} such as CLIP~\cite{radford2021learning,cherti2023reproducible,ilharco_gabriel_2021_5143773}, ALIGN~\cite{jia2021scaling}, CoCa~\cite{yu2022coca}, and SigLIP~\cite{zhai2023sigmoid,tschannen2025siglip} learn vision and text encoders that share a semantic latent space. Such models are typically trained on billions of paired image-text data through contrastive learning, yielding powerful but opaque language-aligned visual representations enabling tasks such as open-vocabulary classification and semantic segmentation. Our proposed method \ours is such a language-aligned vision encoder, but trained to provide a human-interpretable concept representation space.

\myparagraph{Concept extraction methods} aim to decompose learned representations from encoders into human-interpretable concepts through, e.g., clustering~\cite{omahony2023disentangling,kowal2024visual,ghorbani2019towards}, non-negative matrix factorization~\cite{fel2023craft} or a learned concept dictionary that is image-specific~\cite{bhalla2024interpreting} or shared across images~\cite{rao2024discover,zaigrajew2025interpreting,lim2024sparse, parchami2025fact}. The latter, in particular, have been applied to VLMs, but (1) lack spatial grounding of concepts to the image~\cite{rao2024discover,zaigrajew2025interpreting,bhalla2024interpreting}, (2) learn a flat hierarchy of concepts~\cite{rao2024discover,bhalla2024interpreting,lim2024sparse}, or (3) use a limited vocabulary and imprecise names~\cite{rao2024discover}. Our \ours, in contrast, learns a rich set of spatially grounded concepts with explicit parent-child relationships that are named using a large fine-grained vocabulary taking such relationships into account, yielding a significantly more accurate, localized, and interpretable concept decomposition.

\myparagraph{Sparse autoencoders (SAEs)}~\cite{bricken2023monosemanticity,cunningham2023sparse} are models that aim to learn a dictionary of human-interpretable concepts from a latent space. These consist of an affine encoder and decoder that map features to a sparse high-dimensional space of concepts typically trained through a reconstruction and sparsity loss. SAEs have been shown a useful tool to understand large language models (LLMs)~\cite{bricken2023monosemanticity,cunningham2023sparse}, vision models~\cite{rao2024discover,zaigrajew2025interpreting}, and vision-language models~\cite{pach2025sparse}. Recent works suggested alternative sparsity objectives~\cite{rajamanoharan2024improving,rajamanoharan2024jumping,gao2024scaling,bussmann2024batchtopk} and architectural priors for concept hierarchies~\cite{zaigrajew2025interpreting,bussmann2025learning} to improve SAE representations. Here, we combine the best of both worlds and use Matryoshka SAEs~\cite{bussmann2025learning} with a BatchTopK objective~\cite{bussmann2024batchtopk} to extract high quality, localized, and hierarchical concepts.

\myparagraph{Concept-based classification models} such as concept bottleneck models (CBMs)~\cite{koh2020concept,yuksekgonul2022post,oikarinen2023label,rao2024discover,benou2025show} and visual prototype-based %
methods~\cite{chen2019looks,donnelly2022deformable,xue2022protopformer,nauta2023pip} learn concepts from a latent feature space to yield interpretable classification predictions. Where CBMs extract labeled, global concepts that have been shown to often be incorrectly grounded in the image~\cite{huang2024concept}, prototypes are local visual features that have to be investigated by a user to understand their conceptual meaning. Our approach bridges the gap between the two and helps construct inherently interpretable concepts that are labeled, can be global as well as local, and are well grounded in the input. Moreover, in the spirit of foundation models, \ours enables downstream tasks beyond just classification.% 

%% file: sec/3_method.tex
\section{\ours: A Language-aligned Concept Foundation Model for Vision}
\label{sec:method}

To lift concept bottleneck models to be applicable to any vision foundation model downstream task we need accurate, spatially local concept representations.
Here, we propose a language-aligned vision foundation model with explicit concept representations of local image regions, such as patches or pixels, that enable transparent, human-interpretable reasoning.
In a nutshell, we learn a \textit{local} mapping from dense, semantically smoothened image representations to a human-interpretable concept space. This concept space is
encouraged to represent \textit{concepts at varying levels of granularity}, so that we are able to discover locally consistent \textit{concept hierarchies} describing 
part- or specification-relationships between concepts based on patch-wise concept co-occurrences  (\cref{fig:method}).

Specifically, as backbone we leverage CLIP-DINOised image embeddings~\cite{wysoczanska2024clipdinoiser},
a CLIP architecture that encodes images such that semantically similar regions, informed through DINO~\cite{caron2021emerging}, share similar representations.
To map this latent CLIP encoding to meaningful concepts, we train a Sparse Autoencoder (SAE) on patch embeddings to learn \textit{local} interpretable concept representations in an unsupervised manner at scale (\cref{sec:msae}).
Through a Matryoshka objective, we further encourage to learn concepts at different levels of granularity.
From these interpretable concept representations, we show how to extract relations between concepts that reflect both specifications as well as part-relations through an analysis of local co-occurrence patterns (\cref{sec:coocc}).
Finally, we extend the existing concept vocabularies used for labeling, show how to improve concept labeling by leveraging our discovered concept relationships, and show that an architecture-aware text encoding is necessary for accurate labels (\cref{sec:labeling}).

\subsection{From Global to Spatially Localized Concept Representations}
\label{sec:msae}

We first extract human interpretable, spatially localized, and visually grounded concepts from latent features of the CLIP~\cite{radford2021learning} vision encoder. We do this using sparse autoencoders (SAEs)~\cite{bricken2023monosemanticity}, which we discuss first, and then describe our approach.

\subsubsection{Background: Sparse Autoencoders (SAEs).}
Sparse autoencoders provide an efficient and unsupervised method to learn a concept mapping, crucial at the scale of foundation models. These consist of a  light-weight linear layer encoder, $\pi^{enc}:\mathbb{R}^d\rightarrow\mathbb{R}^{m}$ that describes an affine mapping from a $d$-dimensional input space of CLIP features to a sparse representation where each dimension ideally encodes a single interpretable concept. An affine mapping $\pi^{dec}:\mathbb{R}^{m}\rightarrow\mathbb{R}^d$
then maps back to the input space. The SAE is then trained through a reconstruction loss such that $\pi^{dec}\pi^{enc}(x) \approx x, ~~x\in\mathbb{R}^d$
and a sparsity regularization term on the encoder representations, typically via an $\ell_1$~\cite{bricken2023monosemanticity}, TopK~\cite{gao2024scaling}, or BatchTopK~\cite{bussmann2024batchtopk} constraint. 
Recent work~\cite{zaigrajew2025interpreting} proposed Matryoshka SAEs to learn concepts across granularities.
Specifically, let $C=[1\ldots m]$ be the index set for the $m$ neurons in the SAE concept layer and $l_j\in C$ indices until which we define a matryoshka shell. Then
\begin{equation}
\mathcal{L}_{rec} (F_p) = \sum_j \left\Vert \pi^{dec}(~\pi^{enc}(F_p)[1:l_j]~) - F_p\right\Vert_2^2, 
\end{equation}
where $F_p$ are the latent features for patch $p$ given by the backbone model and $ ~\pi^{enc}(F_p)[1:l_j]$ sets all activations of neurons after index $l_i$ to zero. Thus, each Matryoshka shell consisting of the first $l_j$ neurons is trained to reconstruct the \textit{full} latent features, which encourages that the smallest subset of neurons learns coarse, high-level information and higher-index neurons learn the \textit{residual} fine-grained information on top of this coarse representation.

\subsubsection{Extracting Semantic, Spatially Localized, and Grounded Concepts.}
While useful for learning concepts across granularities, concepts from such an SAE are not spatially localized, visually grounded to the input, and may not map well to human interpretable semantic features. To address this, we propose to use the following two approaches (\cref{fig:method}, top).

\myparagraph{Local Concept Extraction.} Instead of encoding pooled visual features, we encode \textit{each patch token} from the final layer of the CLIP vision encoder through the Matryoshka SAE, which helps encode spatially localized concepts. This also allows visual grounding, since concept strengths across spatial regions in the image can now be viewed as a heat map.
We also use BatchTopK~\cite{bussmann2024batchtopk} instead of TopK for sparsity regularization as they provide more flexibility across samples.

\myparagraph{DINOised CLIP Features.} It has been shown that~\cite{zhou2022extract} fine-grained local representations can be extracted from CLIP by repurposing the patch token value vectors from the CLS attention as dense local embeddings. However, despite preserving a degree of spatial locality, they are noisy, which hurts concept extraction. To address this, we adopt the guided pooling approach introduced by CLIP-DINOiser~\cite{wysoczanska2024clipdinoiser}, which smoothens the representations of semantically similar patches using affinity patterns of self-supervised features such as those from DINO~\cite{caron2021emerging}.
Intuitively, this pooling acts as a ``voting system'' that enforces consistency among semantically similar patches while ``attenuating noisy features''~\cite{wysoczanska2024clipdinoiser}.
For efficient inference via a single forward pass,
a small 3$\times$3 convolutional layer
is trained on patch tokens to approximate the DINO affinity map.
In our case, we DINOise the final layer patch tokens from CLIP before using them as inputs to the Matryoshka SAE;
we refer to \cref{app:sec:training} for more details.

As compared to using pooled CLIP features, our approach helps obtain localized and better human-aligned concepts, since these DINOised local embeddings are pooled from semantically coherent regions of the image which frequently correspond to meaningful object parts or localized visual attributes. Moreover, these extracted local embeddings remain well aligned with CLIP’s language space, allowing them to still be directly associated with natural language descriptions through CLIP’s pretrained text encoder.

\input{tables/combined_locality_metrics_res}  

\subsection{Discovering Concept Relationships}
\label{sec:coocc}

The concept extraction procedure from \cref{sec:msae} provides localized concepts at diverse granularities, but does not reveal inter-concept relationships. For example, a concept could be a specialization (\eg `leatherback turtle') or part (\eg `turtle shell') of a more general parent concept (\eg `turtle'). Identifying such relationships would help interpretability and allow for more effective steering of model decisions.
Despite being trained to encode concepts at multiple granularities, extracting explicit relations from Matryoshka SAE concepts has so far not been explored. Other prior works rely on predefined concept hierarchies or leverage LLMs~\cite{sun2024eliminating, panousis2024coarse}
to order concepts in a hierarchical manner, but these may not reflect the structure that emerges in the data or the model’s reasoning process. In contrast, by discovering concept relations directly from data, we avoid mismatches between imposed taxonomies and the representations that the model actually uses.

To represent hierarchies, we identify pairs of 
parent and child concepts (\cref{fig:method}, bottom left), where parents represent more general concepts
and children capture more fine-grained or part-based specializations. In such a relation, the parent concept must also be active (\ie, above a threshold, see \cref{app:sec:training}) whenever the specialized child concept is active.
To identify such pairs, we construct a co-occurrence matrix $C\in [0,1]^{V\times V}$, such that $C_{ij}$ denotes the weighted conditional probability of concept $i$ being active given that concept $j$ is active, \ie

\begin{equation}
    C_{ij} = \frac{\sum_{p: i,j\in B(p)}A_{jp}}{\sum_{p: j\in B(p)}A_{jp}}
    \label{hierarchy_matrix}
\end{equation}
where $A_{jp}$ denotes the activation strength of concept $j$ for patch $p$ and $B(p)$ is the set of concepts active for patch $p$. We then use a threshold $\tau$ to prune weak associations, \ie, a concept $u$ is assigned to be a parent of concept $v$ if $C_{uv}\ge\tau$, and construct our hierarchy.

For this to work, our design with \emph{local, well-grounded} concepts is crucial as otherwise spatially unrelated concepts (e.g., \emph{field} and \emph{cow}) may incorrectly be assigned a relationship merely due to spurious correlations in their image-level activations.
The discovered relations refine explanations through inter-concept dependencies and serve as a prior that allows to learn better concept labels discussed next.

\begin{figure*}[!t]
\centering
\includegraphics[width=.9\linewidth]{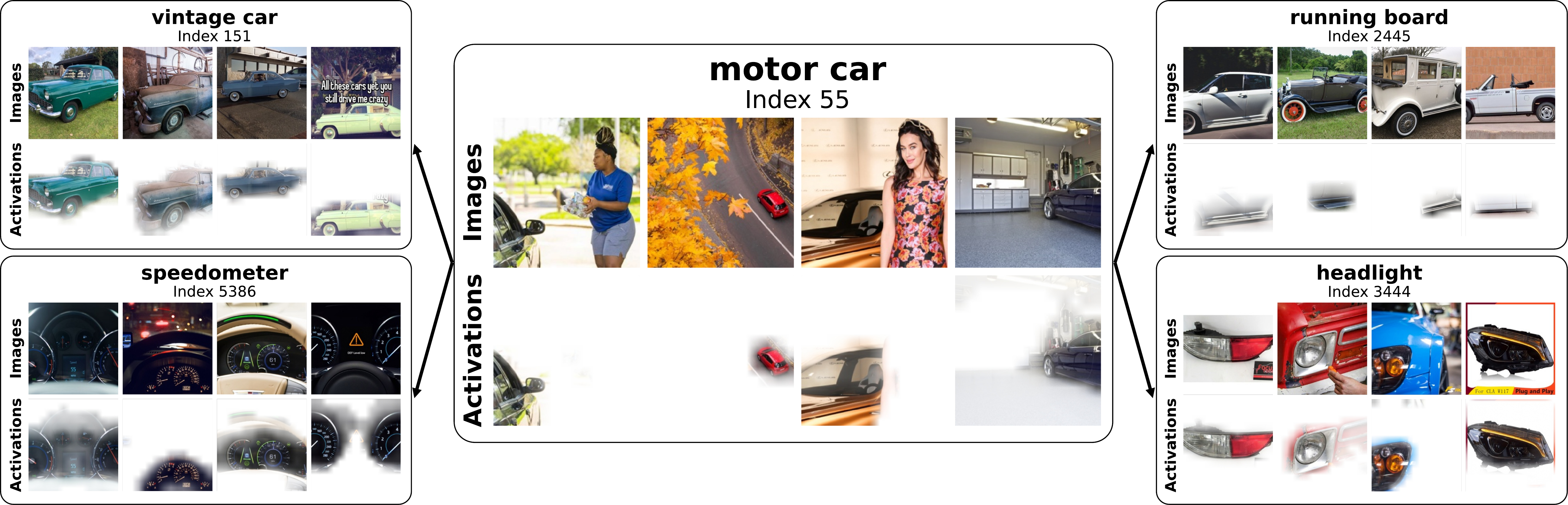}
\caption{\textbf{A concept relationship subgraph.} The parent concept, `motor car', is shown in the center with children concepts arranged around it. For each concept, we provide the assigned label (top) and the top-4 most activating images (first row) along with their spatial localization (second row) based on concept activation strength per patch.}\label{fig:qualfamily}
\end{figure*}

\subsection{Hierarchy-aware Concept Labeling}
\label{sec:labeling}

For human-understandable textual explanations of decision-making, prior work~\cite{rao2024discover} assign labels for each concept neuron in an automated manner. 
They leverage the language-alignment of CLIP to assign labels $l_i$ to a concept $i$, by finding the label $v$ from a pre-defined vocabulary $\mathcal{V}$ whose CLIP text embeddings $\mathcal{T}(v)$ have the highest cosine similarity to the concept dictionary vector, given by the SAE decoder weights $\pi^{dec}_i$, \ie
\begin{equation}
l_{i} = \text{argmax}_{v \in \mathcal{V}} \cos \left( \pi^{dec}_i, \mathcal{T}(v) \right).
\end{equation}
While this performs well for general concepts in the first Matryoshka group of the SAE, it fails for subsequent groups with fine-grained concepts since it does not take into account inter-concept dependencies (see \cref{app:sec:training}).
To account for this, we propose a hierarchy-aware naming scheme (\cref{fig:method}, bottom right) that offsets child concept dictionary vectors with those of their parents, \ie
\begin{equation}
l_{j} = \text{argmax}_{v \in \mathcal{V}} \cos \left( \left(\sum_{(i\to j)\in G} \pi^{dec}_{i} \right)+\pi^{dec}_{j} + b, \mathcal{T}(v) \right),
\end{equation}
where $b$ is the SAE decoder bias and $(i\to j)$ denotes a parent-child relationship in the extracted hierarchy $G$. 

Putting all pieces together (\cref{fig:method}), we obtain \ours, a concept foundation model which extracts language-aligned, spatially grounded, visually localized human-interpretable concepts across granularities for bringing interpretability to diverse downstream tasks.

\input{tables/name_accuracy}

%% file: tables/combined_locality_metrics_res.tex
\begin{figure}[!t]
\begin{minipage}{0.57\textwidth}
\centering
\scriptsize

\begin{tabular}{lcc|cc|cc|c}
\toprule
& \multicolumn{2}{c|}{\textbf{Loc.} $\uparrow$} & \multicolumn{2}{c|}{\textbf{Cons.} $\uparrow$} &\multicolumn{2}{c}{ \textbf{Impur.} $\downarrow$} & \bf {C$^2$-Sc.} $\uparrow$\\
\textbf{Methods}  & Part & Stuff & Part & Stuff & Part & Stuff & IMN\\ \midrule
\salfcbm \cite{benou2025show}      & 16.4 & 19.5 & 9.9 & 6.7 & 0.841 & 1.143        & 0.298 \\
\patchsae \cite{lim2024sparse} & 32.2 & 34.6 & 13.1 & 12.7 & 0.353  & 0.663         & 0.295 \\
\rowcolor{ForestGreen!15!white} \ours & \textbf{44.3} & \textbf{44.5} & \textbf{15.3} & \textbf{13.1} & \textbf{0.333} & \textbf{0.539} & \bf 0.465 \\
\bottomrule
\end{tabular}
\end{minipage}
\hfill
\begin{minipage}{0.4\textwidth}
        \centering
        \includegraphics[width=.8\textwidth]{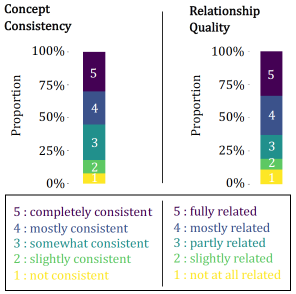}
\end{minipage}

\caption{\textbf{\ours discovers consistent, well-grounded, and well-named concepts.} \textbf{\emph{Left:}} We quantitatively evaluate the consistency and grounding of our concepts using annotation-based and annotation-free metrics. Specifically, we compare against state-of-the-art concept extraction methods on part-annotated \underline{Part}ImageNet and COCO-\underline{Stuff}, using metrics from ~\cite{pham2025escaping}, and measure \underline{loc}alization of concepts, \underline{cons}istency of localization across images, and concept \underline{impur}ity across parts. We also evaluated \underline{C$^2$-Sc}ore~\cite{parchami2025fact} over ImageNet, which measures consistency of attributions independent of human annotations. %
\textbf{\emph{Right:}} User study results on concept consistency (left) and relationship accuracy (right) for $\sim$1000 neurons each covered with 5 users in Amazon MTurk. We find our concepts to be highly consistent and with accurate relations.
}
\label{fig:interpret_local}
\end{figure}

%% file: tables/name_accuracy.tex
\begin{figure}[!t]
    \centering
    \begin{subfigure}[b]{0.25\textwidth}
        \centering
        \includegraphics[width=\textwidth]{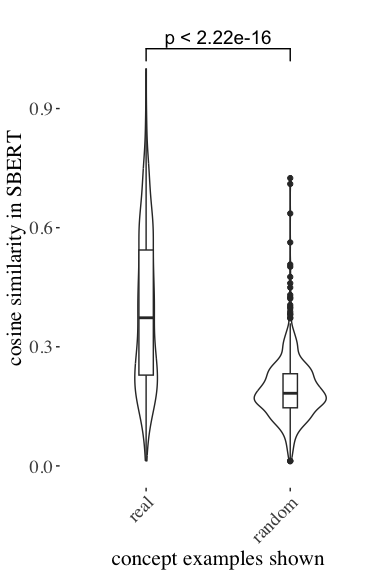}
    \end{subfigure}
    \hfill
    \begin{subfigure}[b]{0.33\linewidth}
        \centering
        \includegraphics[width=\linewidth]{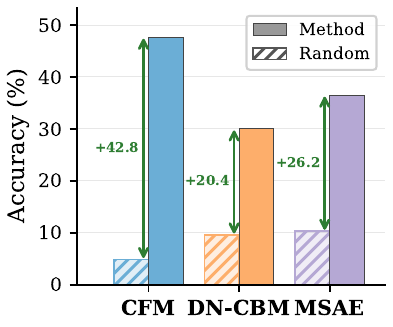}
    \end{subfigure}
    \hfill
    \begin{subfigure}[b]{0.35\textwidth}
        \centering
        \includegraphics[width=\textwidth]{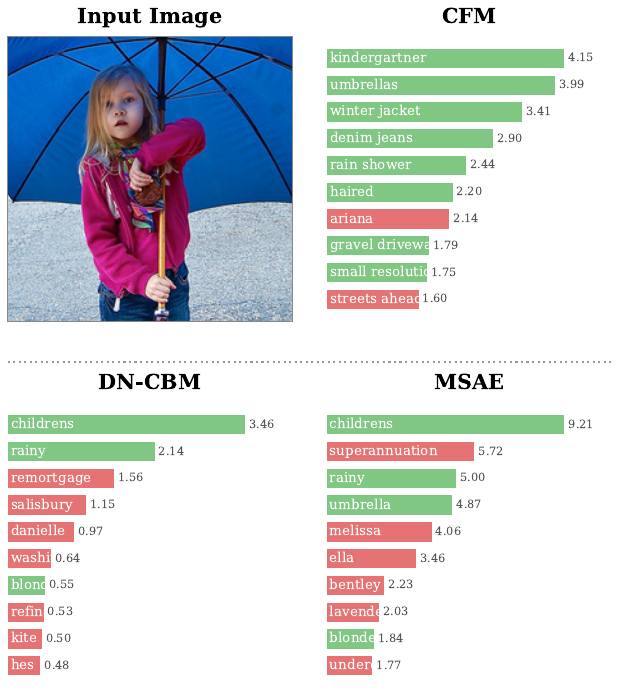}
    \end{subfigure}
    \caption{\textbf{\ours assigns concept names accurately.} \textbf{\emph{Left:}} We ask human annotators to provide a 1-2 word label for each concept, and compute cosine similarities with the concept name assigned by \ours using SBERT~\cite{reimers-2019-sentence-bert}, and find a significant improvement over a random baseline. \textbf{\emph{Middle:}} We use gpt-5-mini~\cite{singh2025openai} as a judge to evaluate whether the top-10 concepts predicted by \ours and baselines are present in the image. \ours provides the most accurate concept names and has the biggest improvement over the random baseline. \textbf{\emph{Right:}} A qualitative example of concept name accuracy. Green (red) indicate concepts present (absent) in the image as per the judge.}
    \label{fig:name_accuracy}
\end{figure}

%% file: sec/5_results.tex
\section{Experimental Evaluation}
\label{sec:results}

We construct a \ours for a CLIP ViT-B/16 and use CC12M~\cite{changpinyo2021conceptual} to train the SAE with 8192 concepts and $K=12$ concepts per patch for our main experiments. For more details including ablations on the number of concepts, see \cref{app:sec:training}.
In \cref{sec:conceptquality}, we evaluate the quality and interpretability of our extracted concepts, \ie how localized, consistent, and pure they are via quantitative metrics and a human user study. We also evaluate accuracy of concept relationships and assigned names from \ours and compare them against baselines. Then, in \cref{sec:foundation_model_applications}, we evaluate the utility of \ours as a foundation model for diverse downstream tasks. We first evaluate for image classification and show comparable performance to baselines, while enhancing interpretability. Next, we move to tasks that require local concept representations and show that \ours can be applied to yield meaningful image captions that are steerable through our concept layer. Finally, we use \ours for Open Vocabulary Segmentation (OVS) and show that it provides performance competitive with baselines while being the only method to provide explanations for the predicted segmentation.

\subsection{Concept Quality and Interpretability}
\label{sec:conceptquality}

\subsubsection{Locality, Consistency, Purity.} We quantitatively evaluate concepts extracted from \ours (\cref{sec:msae}) for locality, consistency, and purity using metrics adapted from Pham \etal~\cite{pham2025escaping}, on the part-annotated PartImagenet~\cite{he2022partimagenet} and COCO-Stuff~\cite{caesar2018coco} datasets, and compare against state-of-the-art grounded concept extraction methods---SALF-CBM~\cite{benou2025show}
and PatchSAE~\cite{lim2024sparse}. Specifically, to measure \textit{locality}, we evaluate how well the concepts represent a human-annotated part by finding the maximal IoU between regions where the concept is active and the part annotation, across all parts. To measure \textit{consistency}, \ie how consistent parts map to the same concepts across images, we compute the maximal sum of part-specific IoUs across all parts.
Lastly for the \textit{impurity}, we evaluate how \textit{semantically impure} a concept is by computing the per-concept entropy of normalized activation values across all parts.
To avoid bias from dataset-annotated parts, we also compute the C$^2$-Score~\cite{parchami2025fact}, which measures grounding consistency in DINOv2~\cite{oquab2024dinov2} feature space, on ImageNet~\cite{deng2009imagenet}. We find that \ours consistently outperforms (\cref{fig:interpret_local}-left) all baselines across datasets and metrics, and in particular provides a significant improvement for locality. For more details, see \cref{app:sec:experiment_details}, and for comparisons with non-grounded baselines using post-hoc attribution methods, see \cref{app:sec:results}.

While relying on human-annotated part data for quantitative evaluation is an established benchmark in the field and provides a proxy for interpretability, ultimately we are interested in how well humans perceive and understand our discovered concepts.
To do this, we conduct a large-scale user study via Amazon MTurk for concept consistency, with 1000 concepts and five annotators per task, which ask annotators to rate concepts on a five point scale. Each concept is shown via its top activating images and the regions it activates at. We show aggregated results in \cref{fig:interpret_local}-right and provide detailed study design and results in \cref{app:sec:experiment_details}. We find that users rate concepts mostly or completely consistent for more than half of the shown concepts. More than 80\% of annotators assign \textit{at least} somewhat consistent, \ie score 3 or higher. Taken together, we show that \ours \textbf{provides more interpretable and well-grounded concepts}.

\subsubsection{Relationship Quality.} We measure accuracy of our discovered hierarchy (\cref{sec:coocc}) via a similar user study as above, where we show annotators pairs of two concepts and ask them to rate how well related they are. As shown in \cref{fig:interpret_local}-right, we find that like with consistency, users rate concept hierarchies from \ours to be `mostly' or `fully' related for most pairs. Additionally, we evaluate the accuracy of the directionality of the relationship via a gpt-5.5 model judge, by providing top-5 images and their corresponding activating regions for parent-child pairs and asking the model to evaluate relationship direction. For consistent and related concepts, we obtain an accuracy of 79.79\% for pairs found by CFM. For more details, see \cref{app:sec:experiment_details}.

We also show a qualitative example of discovered concept relations in \cref{fig:qualfamily}, which shows visually well-grounded and meaningfully related concepts of varying granularities, and provide more examples in \cref{app:sec:results}. Overall, we show that \textbf{\ours organizes concepts into meaningful hierarchies} aiding interpretability. 

\input{tables/class_res}

\myparagraph{Naming Accuracy.} We perform two evaluations for the accuracy of concept names assigned by \ours (\cref{sec:labeling}). First, we perform a user study, where we ask annotators to provide a 1-2 word description for a concept given top activating images with grounding. We then measure the cosine similarity between user-provided names and names assigned by \ours using SentenceBERT~\cite{reimers-2019-sentence-bert}, and observe a strong statistically significant ($p< 2.22\times 10^{-16}$ Wilcoxon rank-sum test) difference (\cref{fig:name_accuracy}-left) between our assigned names versus randomly drawn names using the set of concept names discovered by our method.

To evaluate its utility as a foundation model that can provide concepts, we also evaluate the accuracy of concept names found by \ours against baselines for images from the COCO~\cite{lin2014microsoft} validation set. We use gpt-5-mini~\cite{singh2025openai} as a judge, and ask it to decide whether each concept name from the top-10 concepts reported by each of the methods is present in the image. We find that \ours provides far more accurate concepts (\cref{fig:name_accuracy}-middle) as compared to the random baseline of randomly sampling concepts from the concept sets. This is also reflected qualitatively in \cref{fig:name_accuracy}-right. Overall, we show that \textbf{\ours assigns names that meaningfully align with human labeling} and more accurately describe what is in an image as compared to prior work.

\subsection{Foundation Model Applications}
\label{sec:foundation_model_applications}

\subsubsection{Image Classification.} We evaluate \ours for image classification on ImageNet~\cite{imagenet} and Places365~\cite{zhou2017places} by
training a linear classification head for each.
We compare with recent CLIP ViT-B/16 based concept bottleneck models, \ie LF-CBM \cite{oikarinen2023label}, LaBo \cite{yang2023language},
CDM \cite{panousis2023sparse}, DCLIP \cite{menon2022visual},
\dncbm \cite{rao2024discover},
\dcbm \cite{prasse2025dcbm}, and
\cfcbm \cite{panousis2024coarse}. We additionally include the ResNet-50 based  
\salfcbm \cite{benou2025show}.
All baselines except \salfcbm and \ours only provide image-level concepts. We find that \textbf{\ours yields competitive classification accuracies (\cref{fig:classification}-left) while also providing more interpretable and spatially grounded concepts (\cref{fig:classification}-right)}. Notably, all the baselines compared against are trained explicitly and only for classification, while \ours can do much more, which highlights the versatility of a concept foundation model.

\input{tables/captioning}

\subsubsection{Image Captioning.}
Next, we go beyond classification and show the utility of \ours's interpretable concept-based representation for image captioning. We follow the LLaVA setup \cite{liu2023InstructTune}, by learning a projection for spatial output tokens from \ours to a language model through an MLP adapter and fine-tune both the adapter and LLM (Gemma-2-2B Instruct~\cite{gemmateam2024gemma2improvingopen}) captioner on CC12M DreamLIP Long captions~\cite{changpinyo2021conceptual,Zheng2024DreamLIP}. In \cref{fig:captioning}-left we observe that \ours maintains captioning performance similar to the opaque, dense representation of the backbone, while providing an explicit concept representation. Importantly, \textbf{this representation can then be used to steer the captioning model}, as shown in \cref{fig:captioning}-right.

\subsubsection{Open-Vocabulary Segmentation (OVS).}
\begin{figure}[!t]
    \centering
    \includegraphics[width=0.9\linewidth]{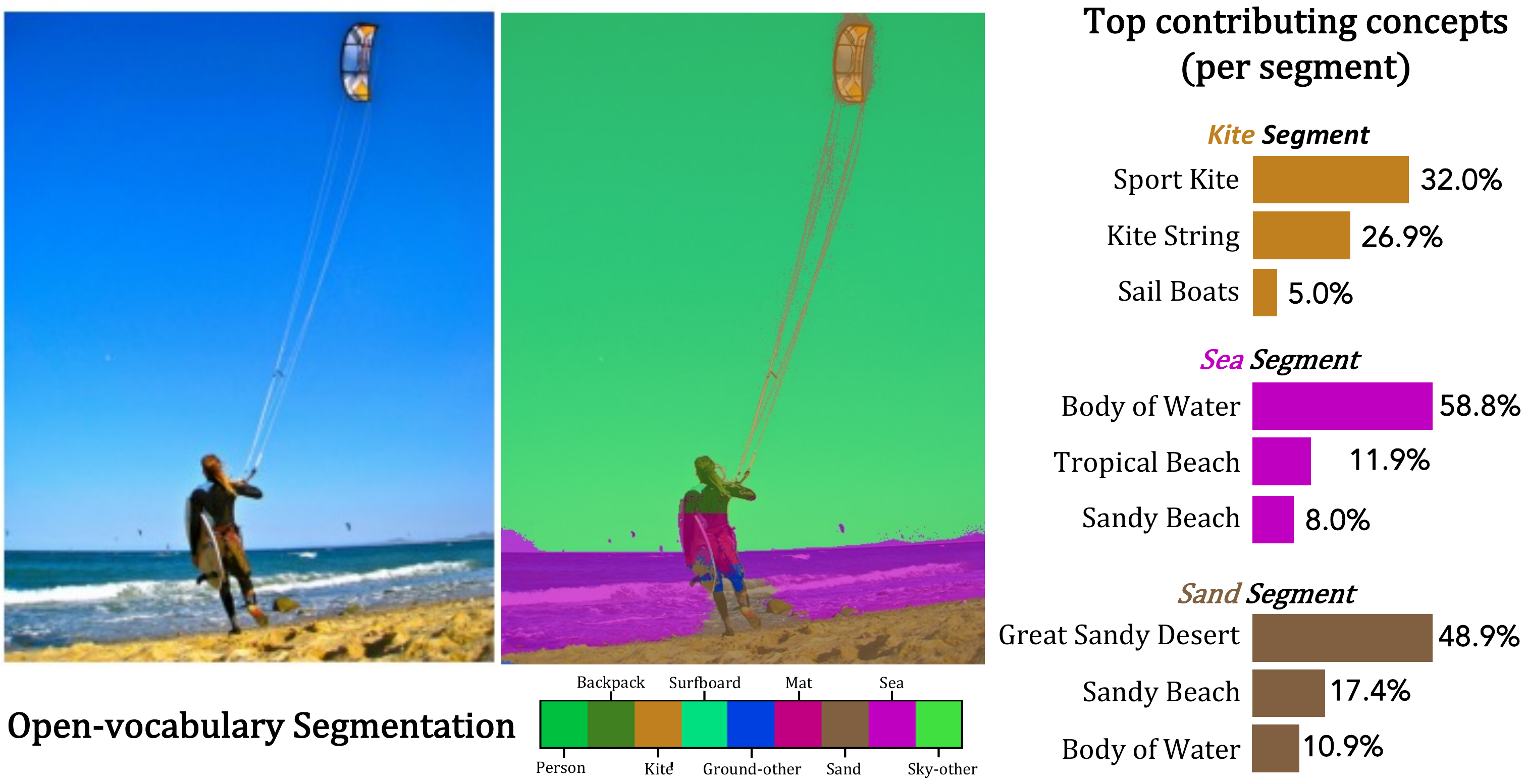}
    \caption{\textbf{Explainable Open-Vocabulary Segmentation.} $\ours^\text{+(AnyUp)}$ pixel-level open-vocabulary segmentation (left) with concept-based explanations
    for each segment (right).}
    \label{fig:qualSegmentation}
\end{figure}

\input{tables/open_vocab_seg_res_small}
Finally, we use \ours for the challenging fine-grained task of semantic segmentation. Its language alignment allows for open-vocabulary segmentation (OVS), \ie can be used as is without expensive pixel-level data annotation limited to a fixed set of classes. To do this, we compute the cosine similarity between the embedding of each word in the segmentation vocabulary and the reconstructed \textit{language aligned embedding} of the SAE. 
We consider two ways to obtain dense predictions. In one approach, we reconstruct patch-wise CLIP embeddings with the SAE and compute similarities to the vocabulary, after which predictions are upsampled to the pixel level using bilinear interpolation. Alternatively, we upsample the spatial concept activations directly to the pixel level using AnyUp~\cite{wimmer2025anyup} and reconstruct the CLIP embeddings with the SAE directly at pixel resolution before computing cosine similarities. We indicate the latter approach with +(AnyUp) in \cref{table:main_nobg,fig:qualSegmentation}. In both cases, predictions are interpretable: the reconstructed CLIP embeddings are linear combinations of concepts through the SAE decoder, allowing either patch or pixel-level segmentation decision to be explained by a linear combination of concepts. 

We compare against the reported performance of prototype-based~\cite{shin2022reco, karazija2024diffusion}, segmentation-specific~\cite{xu2022groupvit, rewatbowornwong2023zero, luo2023segclip, cha2023learning, ranasinghe2023perceptual, xu2023learning}, as well as MaskCLIP and CLIP-DINOiser versions~\cite{zhou2022extract, wysoczanska2024clipdinoiser}
on a variety of benchmark datasets, including PASCAL VOC and VOC20~\cite{everingham2011pascal}, PASCAL Context and Context59~\cite{mottaghi2014role}, COCO Object~\cite{lin2014microsoft} and Stuff~\cite{caesar2018coco}, Cityscapes~\cite{cordts2016cityscapes}, and ADE20K~\cite{zhou2019semantic}. We report mIoU in \cref{table:main_bg} with and without a background prompt following established benchmarking protocols~\cite{wysoczanska2024clipdinoiser}.

We observe that \textbf{\ours performs on par with the strongest opaque baselines}, \ie CLIP-DINOiser and OVDiff. It also outperforms existing specialized architectures by a wide margin across datasets.
In \cref{fig:qualSegmentation} we show a qualitative example of OVS on COCO-Stuff highlighting the interpretable downstream decision-making provided by \ours.

%%%%%%%%%%%%%%%%Ablations

\begin{table}[h]
\centering
\scriptsize
\setlength{\tabcolsep}{2.5pt}
\caption{\textbf{Dictionary Size Ablation.} We evaluate CFM across different dictionary capacities (4k, 8k, 16k) on ImageNet (IMN), Places365 (Places), average Open-Vocabulary Segmentation (OVS Avg), and interpretability metrics. We select 8k as our default to balance segmentation performance and computational efficiency. Best results are \textbf{bold}, second best are \underline{underlined}. Note that the classification results in this table are with the OpenCLIP CLIP ViT-B/16  backbone, in contrast results in \cref{fig:classification} are done with the OpenAI CLIP ViT-B/16 backbone for clear comparability with existing CBMs. }
\begin{tabular}{@{}l|cc|c|cc|cc|cc|c@{}}
\toprule
\multirow{2}{*}{\textbf{Dict. Size}} & \multicolumn{2}{c|}{\textbf{Class.} $\uparrow$} & \textbf{OVS} $\uparrow$ & \multicolumn{2}{c|}{\textbf{Loc.} $\uparrow$} & \multicolumn{2}{c|}{\textbf{Cons.} $\uparrow$} & \multicolumn{2}{c|}{\textbf{Imp.} $\downarrow$} & \textbf{C$^2$} $\uparrow$ \\
& IMN & Places & Avg & Part & Stuff & Part & Stuff & Part & Stuff & Score \\
\midrule
4k & 78.2 & 55.5 & 41.1 & 43.9 & 44.2 & \textbf{15.8} & 12.5 & .340 & \textbf{.540} & \textbf{0.488}\\
8k (Default) & \textbf{78.6} & \underline{55.6} & \textbf{41.3} & \underline{44.3} & \underline{44.5} & 15.3 & \underline{13.1} & \textbf{.330} & \textbf{.540} & \underline{0.465}\\
16k & \textbf{78.6} & \textbf{56.0} & 40.8 & \textbf{44.6} & \textbf{44.7} & \underline{15.7} & \textbf{13.6} & .340 & \textbf{.540} & 0.418\\
\bottomrule
\end{tabular}
\label{tab:main_dict_ablation}
\end{table}

\subsection{Ablation on Concept Set Size}
\label{sec:ablations}
To validate our hyperparameter choices, we ablate the dictionary capacity of \ours in \cref{tab:main_dict_ablation}. We evaluate dictionary sizes of 4k, 8k, and 16k on classification, average performance across datasets for open-vocabulary segmentation, and interpretability metrics. We observe that both downstream performance and concept grounding are highly stable across varying capacities. However, a trade-off emerges: classification accuracy increases with dictionary size, whereas the $C^2$ score decreases. We select 8k as our default architecture, as it achieves the highest average segmentation performance while maintaining strong classification accuracy and interpretability. 

Comprehensive structural ablations—including scaling to the 400M parameter SigLIP-2 backbone, the necessity of the guided pooling mechanism, varying sparsity factors ($k$), and comparisons against alternative SAE objectives—are detailed in \cref{app:sec:ext_ablations}.

%% file: tables/class_res.tex
\begin{figure}[!t]
\begin{minipage}{0.49\linewidth}
    \centering
    \scriptsize
    \begin{tabular}{l c cc}
    \toprule
    Model & \shortstack[c]{Spatial \\ Grounding} & IMN & Places  \\
    \midrule
    Linear Probe & - & 80.2 & 55.1 \\
    Zero Shot & - & 68.6 & 41.2 \\
    \midrule
    LF-CBM \cite{oikarinen2023label} & \xmarkr & 75.4 &  50.6  \\
    LaBo \cite{yang2023language} & \xmarkr &78.9 & -  \\
    CDM \cite{panousis2023sparse} & \xmarkr & 79.3 &  52.6  \\
    DCLIP \cite{menon2022visual} &\xmarkr & 68.0 & 40.3   \\
    \dncbm \cite{rao2024discover} & \xmarkr & \textbf{79.5} & 55.1 \\
    \dcbm \cite{prasse2025dcbm} & \xmarkr & 70.5 & 50.9  \\
    \cfcbm \cite{panousis2024coarse}& \ymarko & 78.5 & -  \\
    \salfcbm \cite{benou2025show}& \cmarkg & 76.3 & 49.4  \\
    \rowcolor{ForestGreen!15!white} \ours  & \cmarkg & 78.9 & \textbf{55.4}  \\
    \bottomrule
    \end{tabular}
\end{minipage}
\hfill
\begin{minipage}{0.49\linewidth}
    \centering
        \includegraphics[width=\linewidth,clip,trim={1cm 7cm 15.5cm 3.5cm}]{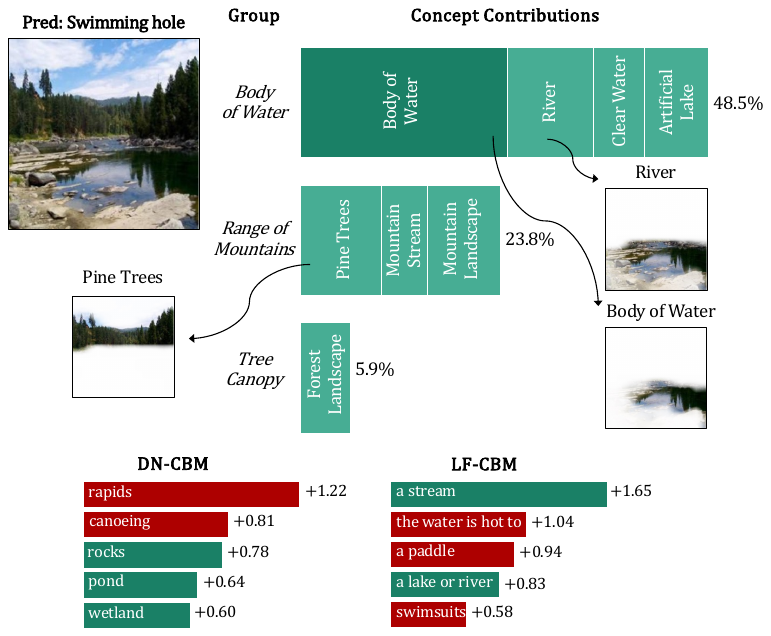}%
        \label{fig:qualClassification:a}
\end{minipage}
\caption{\textbf{\ours shows competitive classification performance while providing spatially grounded concepts.} \textbf{\emph{Left:}} We compare \ours with baselines on ImageNet (IMN) and Places365 (Places). ``Spatial Grounding'' indicates whether methods extract spatially grounded concepts (\cmarkg), partially local concepts (\ymarko), or global concepts only (\xmarkr). Note that ‘Linear Probe’ and ‘Zero-Shot’ refer to the standard CLIP~\cite{radford2021learning} baselines. We show linear probe performance of \ours on max- and mean-pooled concept activations across patches (see \cref{app:sec:probe_training} for specific aggregation details). Baseline results taken from Prasse \etal~\cite{prasse2025dcbm}, we retrained \ours on the same OpenAI-CLIP backbone, best method in \textbf{bold}. \textbf{\emph{Right:}} For an image of the ``Swimming hole'' Class of Places365  we show explanations of \ours against existing methods. For \ours, we provide the parent concept (dark green) and its contributing children (light green) with percentage of contribution and the spatial grounding of the top-3 most contributing concepts. In contrast to our work, existing methods (bottom) show spurious concepts.}
\label{fig:classification}
\end{figure}

%% file: tables/captioning.tex
\begin{figure}[!t]
\centering
\begin{minipage}{0.3\linewidth}
\scriptsize
\begin{tabular}{l c >{\columncolor{ForestGreen!15!white}}c}
\toprule
\bf Metrics $\uparrow$
& \bf \shortstack{CLIP-\\DINO}
& \bf \ours \\
\midrule
CLIP-Sc~\cite{hessel2021clipscore} & 0.72 & 0.72 \\
RefCLIP-Sc~\cite{hessel2021clipscore}  & 0.73 & 0.72 \\
Capture~\cite{Dong2024BenchmarkingAI} & 0.35 & 0.35\\
\bottomrule
\end{tabular}
\end{minipage}
\hfill
\begin{minipage}{0.66\linewidth}
\centering
\includegraphics[width=\linewidth]{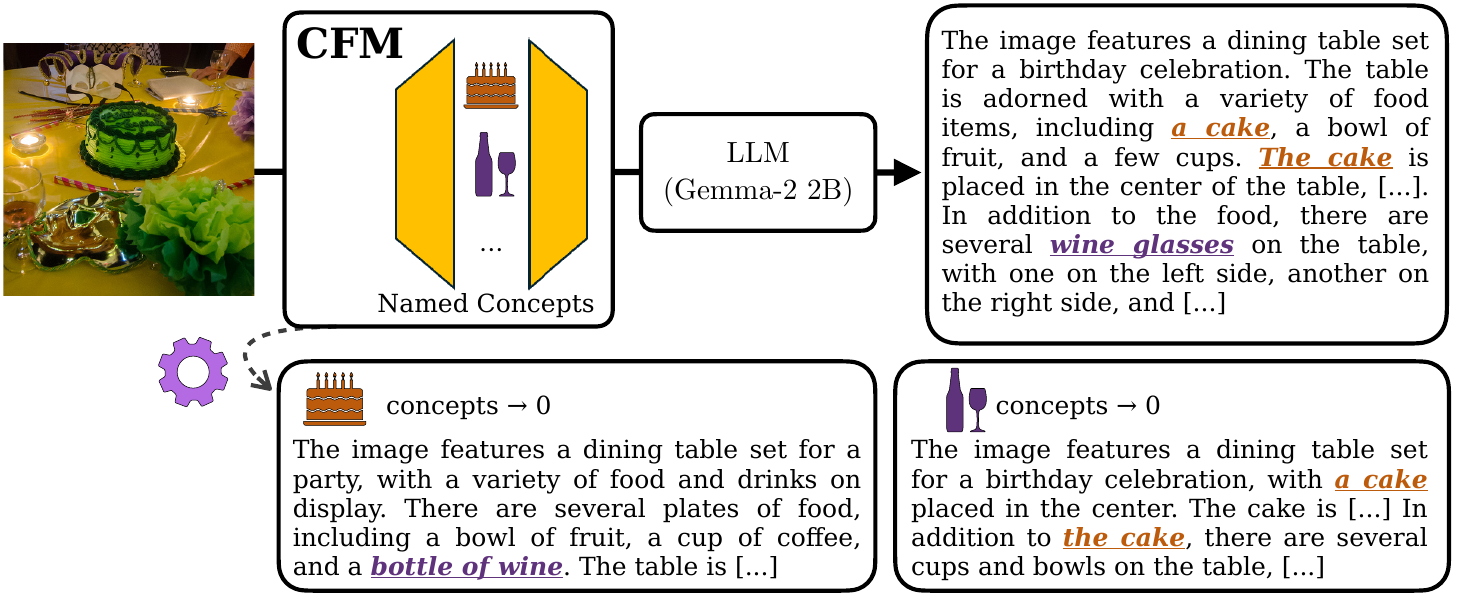}
\end{minipage}
\caption{\textbf{\ours yields competitive and steerable captioning.} \textbf{\emph{Left:}} Captioning capabilities of \ours compared to its opaque backbone. \textbf{\emph{Right:}} Through named concepts, \ours allows for concept-level intervention, steering the image captioning. For example, the `cake' concept can be removed from the caption while preserving the `drink' concept, and vice-versa. See also \cref{app:sec:training,app:sec:results} for further details and examples.}
\label{fig:captioning}
\end{figure}

%% file: tables/open_vocab_seg_res_small.tex
\newcommand{\resultsrownobg}[6]{#1 & #2 & #3 & #4 & #5 & #6 & \more}
\newcommand{\more}[3]{#1 & #2 & #3}

\begin{table}[t]
\centering
\small
\renewcommand{\arraystretch}{1.0}
\setlength{\tabcolsep}{2pt}
\scriptsize

\caption{\textbf{\ours provides strong open-vocabulary semantic segmentation.} We report mIoU between segmentations and annotated parts for each dataset considering evaluation with and without a `background' prompt as discussed in \cref{app:sec:experiment_details}. 
We took baseline results from Wysoczańska \etal~\cite{wysoczanska2024clipdinoiser} and rerun (*) for inference consistency.
First and second best method are \textbf{bold} and \underline{underlined}, respectively.
}

\vspace{-5pt}

\begin{tabular}{l|ccccc|ccc|c}
\toprule
& \multicolumn{5}{c}{\bf No background prompt} & \multicolumn{3}{c}{\bf w/ background}  \\
\textbf{Methods$^\text{+(Variants/Extra Backbones)}$} & VOC20 & C59 & Stuff & City & ADE & Context & Object & VOC & Avg. \\
\midrule

\midrule
\rowcolor{Black!10!white} \multicolumn{10}{l}{\textcolor{Black!10!white}{aaa} \textit{Build prototypes per class}} \\

\resultsrownobg
{ReCo~\cite{shin2022reco}}
{57.8} % VOC20
{22.3} % Context59
{14.8} % Stuff
{21.1} % City
{11.2} 
{19.9} % Context
{15.7} % Object 
{25.1} % VOC\\ 
& 23.5

\\

\resultsrownobg
{OVDiff~\cite{karazija2024diffusion}$^\text{+(DINO \& SD)}$}
{\underline{81.7}} % VOC20
{33.7} % Context59
{-} % Stuff
{-} % City
{14.9} 
{30.1} % Context
{\underline{34.8}} % Object
{67.1} % VOC
& -
\\ %ADE 

\midrule
\rowcolor{Black!10!white} \multicolumn{10}{l}{\textcolor{Black!10!white}{aaa} \textit{Text/image alignment \textbf{training} with captions}} \\

\resultsrownobg
{GroupViT~\cite{xu2022groupvit}}
{79.7} % VOC20
{23.4} % Context59
{15.3} % Stuff
{11.1} % City
{9.2}
{18.7} % Context
{27.5} % Object 
{50.4} % VOC
& 29.4
\\ %ADE

\resultsrownobg{ZeroSeg~\cite{chen2023exploring}}
% {\xmarkr}
% {\xmarkr}
%
% {CLIP}
{-}
{-}
{-}
{-}
{-}
{21.8} % Context
{22.1} % Object 
{42.9} % VOC
& -
\\

\resultsrownobg
{SegCLIP~\cite{luo2023segclip}}
% {\xmarkr}
% {\xmarkr}
%
% 
{-} % VOC20
{-} % Context59
{-} % Stuff
{11.0} % City
{8.7} 
{24.7} % Context
{26.5} % Object 
{52.6} % VOC
& -
\\ %ADE 

\resultsrownobg
{TCL~\cite{cha2023learning}}
% {\xmarkr}
% {\xmarkr}
%
% {CLIP}
{77.5} % VOC20
{30.3} % Context59
{19.6} % Stuff
{23.1} % City
{14.9} 
{24.3} % Context
{30.4} % Object 
{51.2} % VOC
& 33.9
\\ %ADE 

\resultsrownobg
{CLIPpy~\cite{ranasinghe2023perceptual}}
% {\xmarkr}
% {\xmarkr}
%
% {CLIP}
{-} % VOC20
{-} % Context59
{-} % Stuff
{-} % City
{13.5} 
{-} % Context
{32.0} % Object 
{52.2} % VOC
& -
\\ %ADE 

\resultsrownobg
{OVSegmentor~\cite{xu2023learning}}
% 
% 
%
% {CLIP}
{-} % VOC20
{-} % Context59
{-} % Stuff
{-} % City
{5.6} 
{20.4} % Context
{25.1} % Object 
{53.8} % VOC
& -
\\ %ADE 

\midrule

\rowcolor{Black!10!white} \multicolumn{10}{l}{\textcolor{Black!10!white}{aaa} \textit{Use \textbf{Frozen} CLIP}} \\

\resultsrownobg
{CLIP-DIY~\cite{wysoczanska2024clip}$^\text{+(DINO)}$}
{79.7} % VOC20
{19.8} % Context59
{13.3} % Stuff
{11.6} % City
{9.9} 
{19.7} % Context
{31.0} % Object 
{59.9} % VOC
& 30.6
\\ %ADE 

\resultsrownobg
{MaskCLIP~\cite{zhou2022extract}} %(OpenCLIP Laion 2B)}
{61.8} % VOC20
{25.6} % Context59
{17.6} % Stuff
{25.0} % City
{14.3} 
{22.9} % Context 
{16.4} % Object
{32.9} % VOC
& 27.1
\\ %ADE 

\resultsrownobg
{MaskCLIP~\cite{zhou2022extract}$^\text{+(ref)\cite{cha2023learning}}$} %(OpenCLIP Laion 2B)}
{71.9} % VOC20
{27.4} % Context59
{18.6} % Stuff
{23.0} % City
{14.9} 
{24.0} % Context
{21.6} % Object
{41.3} % VOC
& 30.3
\\ %ADE

\resultsrownobg
{CLIP-DINOiser$^{*}$~\cite{wysoczanska2024clipdinoiser}}
{80.8} % VOC20
{36.0} % Context59
{\underline{24.9}} % Stuff
{\bf 39.4} % City
{20.5} 
{32.5} % Context
{34.7} % Object
{62.2} % VOC
& 41.4
\\ %ADE 

\resultsrownobg
{CLIP-DINOiser$^{*}$~\cite{wysoczanska2024clipdinoiser}$^\text{+(AnyUp)\cite{wimmer2025anyup}}$}
{81.6} % VOC20
{\underline{37.2}} % Context59
{\bf 25.4} % Stuff
{\underline{39.1}} % City
{\underline{20.9}} 
{\underline{33.8}} % Context
{\bf 35.2} % Object
{\bf 64.0} % VOC
& \bf 42.2
\\ %ADE 

\midrule

\rowcolor{Black!10!white} \multicolumn{10}{l}{\textcolor{Black!10!white}{aaa} \textit{Use \textbf{Frozen} CLIP with \textbf{Interpretability}}} \\

\resultsrownobg
{\rowcolor{ForestGreen!15!white} $\text{\ours}$}
% {\cmarkg}
% {\cmarkg}
%
% {\xmarkr}
% {CLIP}
{80.7} % VOC20
{36.5} % Context59
{24.2} % Stuff
{38.5} % City
{20.7} % ADE 
{33.1} % Context
{34.7} % Object 33.3
{62.2} % VOC 61.6
& 41.3
\\ %ADE 

\resultsrownobg
{\rowcolor{ForestGreen!15!white} $\text{\ours}$$^\text{+(AnyUp)\cite{wimmer2025anyup}}$}
% {\cmarkg}
% {\cmarkg}
%
% {\xmarkr}
% {CLIP}
{\bf 81.8} % VOC20
{\bf 37.6} % Context59
{24.4} % Stuff
{38.1} % City
{\bf 21.1} % ADE 
{\bf 34.3} % Context
{33.9} % Object
{\underline{63.6}} % VOC
& \underline{41.9}
\\ %ADE 

\bottomrule
\end{tabular} 

\label{table:main_nobg}
\label{table:main_bg}
\vspace{-10pt}
\end{table}

%% file: sec/6_conclusion.tex
\section{Discussion and Conclusion}
\label{sec:conclusion}

We proposed \ours, a concept-based representation for a language-aligned vision encoder that represents
spatially grounded and human-interpretable concepts of differing granularity.
As such, \ours sets a new standard in the field of concept-based interpretability by providing meaningful, well-grounded concepts that are well-organized into hierarchies with meaningful concept names.

This work also comes with limitations. For instance, despite being significantly improved as compared CBMs and at foundation model scale (\cref{fig:name_accuracy}), there is still room for improvement for concept naming.
There is also a plethora of evaluation strategies suggested for explanations, yet most of them are not directly applicable to concept-based explanations learned in an unsupervised manner. In this work, we focused on a set of interpretability metrics that cover a range of interpretability aspects and were established recently in the literature.

Building a highly diverse concept representation space by learning on corresponding large (web-scale) datasets would be an interesting direction for future work. Another highly interesting future application is enabling interpretable vision tokens to be used with large vision-language models (LVLMs) for tasks such as visual question answering (VQA).

In summary, \ours is \textbf{the first concept-based foundation model unlocking explanations for all its downstream vision tasks}.

%% file: sec/X_suppl.tex
\input{sec/X_suppl_cover}
\clearpage

\clearpage
\section{Evaluation Details}
\label{app:sec:experiment_details}
\subsection{Quantitative Interpretability Metrics}
\label{app:sec:experimental_details:interpretability_metrics}
To quantitatively evaluate the locality of our discovered concepts, we compare \ours against four existing approaches: DN-CBM (data-driven global concepts), MSAE (data-driven global concepts), PatchSAE (data-driven local concepts), and SALF-CBM (predefined local concepts). Given the substantial differences in concept vocabularies, sparsity patterns, and spatial representations across these methods, we develop a comprehensive evaluation framework with five complementary metrics inspired by Pham \etal~\cite{pham2025escaping}, adapted to enable fair comparison across diverse concept discovery approaches and diverse concept sets.
%%%%%%%%%%%%%%%%%%%%%%%%%%%%
\subsubsection{Locality Metric.}
This metric measures how well spatial activations of the concepts match human-annotated segments:
Let $\mathds{1}_c(i,j)$ and $\mathds{1}_l(i,j)$ denote binary indicators of whether concept $c$ or label $l$ is active at pixel $(i,j)$, and let $\mathds{1}_{c,l}(i,j) = \max\big(\mathds{1}_c(i,j), \mathds{1}_l(i,j)\big)$. For a concept $c$ and label $l$ on image $x$:
\begin{equation}
\text{IoU}_{c,l}(x) = \frac{\sum_{i,j} \mathds{1}_c(i,j) \cdot \mathds{1}_l(i,j)}{\sum_{i,j} \mathds{1}_{c,l}(i,j)}.
\end{equation}
The overall Locality score is:
\begin{equation}
\text{Locality} = \frac{1}{|L|} \sum_{l \in L} \frac{1}{|I_l|} \sum_{x \in I_l} \max_{c \in C} \text{IoU}_{c,l}(x),
\end{equation}
where $L$ is the set of labels and $I_l$ is the set of images containing label $l$. For each image, we compute per human-annotated label the maximum IoU over all concepts, average it over all images containing that label, and then average across labels.
To obtain $\mathds{1}_c$, we bilinearly upsample each concept map to the input resolution and keep, per concept, the highest-valued pixels covering $90\%$ of its activation energy. Following Pham \etal~\cite{pham2025escaping}, ground-truth masks are dilated with a $3\times3$ max-pooling filter, and images and masks share the same resize and center-crop to $224\times224$. The background label is excluded for Locality and Consistency but included for Impurity.
%%%%%%%%%%%%%%%%%%%%%%%%%%%%
\subsubsection{Consistency Metric.}
This metric evaluates whether concepts reliably capture the same semantic regions across images:
\begin{equation}
\text{Consistency} = \frac{1}{|L|} \sum_{l \in L} \max_{c \in C} \frac{1}{|I_l^c|} \sum_{i \in I_l^c} \text{IoU}_{c,l}(i),
\label{consistency-iou}
\end{equation}
where $I_l^c$ is the set of images where either concept $c$ or label $l$ is present. Images where only one of them is present contribute an IoU of $0$. For each concept-label pair, we compute the mean IoU across all images where the concept or the label is present, then take the maximum over concepts for each label. 
%%%%%%%%%%%%%%%%%%%%%%%%%%%%
\subsubsection{Impurity Metric.}
This metric measures the spatial focus of concepts within single labels:
\begin{equation}
\text{Impurity} = \frac{1}{|C_{active}|} \sum_{c \in C_{active}} \frac{1}{|I_c|} \sum_{i \in I_c} H(p_i^c),
\end{equation}
where $C_{active}$ is the set of concepts that activate in at least 5 images for PartImageNet and 20 images for COCO-Stuff, $I_c$ is the set of images where concept $c$ is active, and $H(p_i^c)$ is the entropy of the normalized activation distribution of concept $c$ across all labels (including background) in image $i$:
\begin{equation}
H(p_i^c) = -\sum_{l \in L \cup \{\text{bg}\}} p_i^{c,l} \log p_i^{c,l},
\end{equation}
where $p_i^{c,l} = \frac{\sum_{\text{pixels} \in S_i^l} B_i^c}{\sum_{\text{all pixels}} B_i^c}$. For each concept, we compute the entropy of its activations across all human-annotated labels (including background), then average over the images. For the final score we average over all concepts that activate at least in the specified number of images. Lower values indicate concepts that concentrate within specific semantic regions rather than spreading across multiple labels. Note that the metric is on a log scale.

\subsection{User Study}
\label{app:sec:experimental_details:user_study}
To evaluate concept quality from a human perspective, we performed a large scale user study. In this section, we describe the setup and provide additional analyses of the results.

\subsubsection{Study Setup.}
\label{app:sec:experimental_details:user_study:setup}

We conducted a study in Amazon Mechanical Turk~\cite{amazonmturk} comprising three tasks: evaluating (1) concept consistency, (2) concept naming quality, and (3) concept relationship quality.
Task 1 and task 2 consist of 1000 questions each, while task 3 contains 1100 questions. Each question is supposed to be labeled by five annotators. For each task, groups of 20 questions are created to form `Human Intelligence Tasks' (HITs) in Amazon Mechanical Turk. A HIT is the smallest task unit an annotator can complete, \ie for any task an annotator reads the instructions and completes 20 questions (and might optionally continue completing more HITs). To help annotation quality, we use annotators located in the US, with at least 10000 approved HITS, and with at least a 98\% HIT approval rate. We pay annotators \$2.00 per completed HIT (\ie \$0.10 per question), with an estimated time of five minutes for solving a HIT. All HITs inform annotators on how their responses will be used and require them to provide consent. In the following, we provide additional details about each task.

\begin{figure}[!ht]
    \centering
    \begin{subfigure}{0.49\linewidth}
        \includegraphics[width=\linewidth]{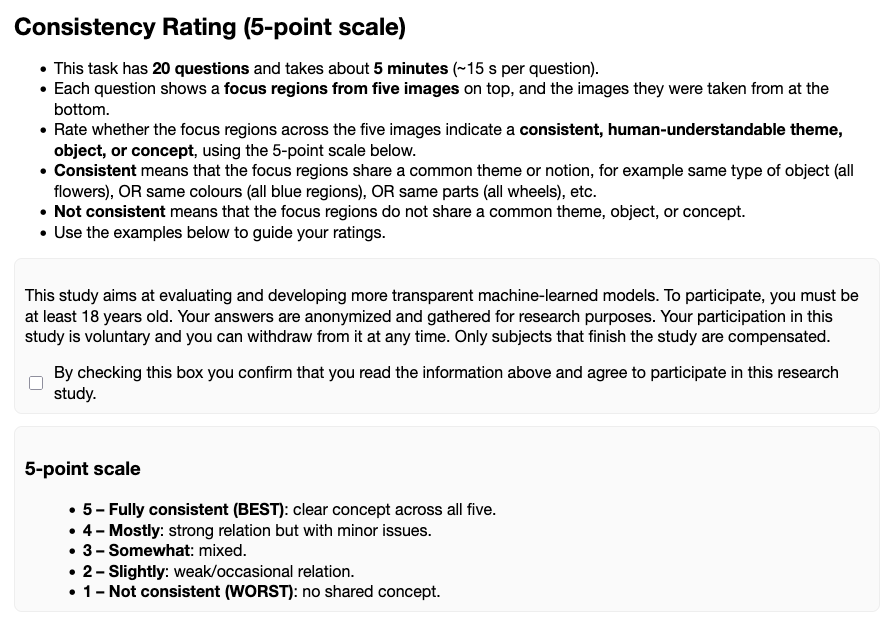}
        \vspace{10pt}
        \includegraphics[width=\linewidth]{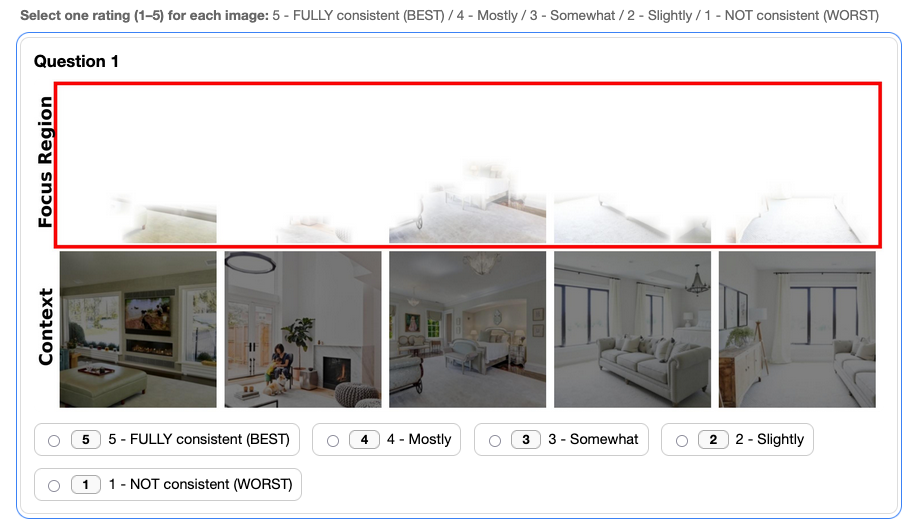}
        \caption{\emph{Top:} Task instructions. \emph{Bottom:} A question in the task.}
    \end{subfigure}
    \hfill
    \begin{subfigure}{0.49\linewidth}
        \includegraphics[width=\linewidth]{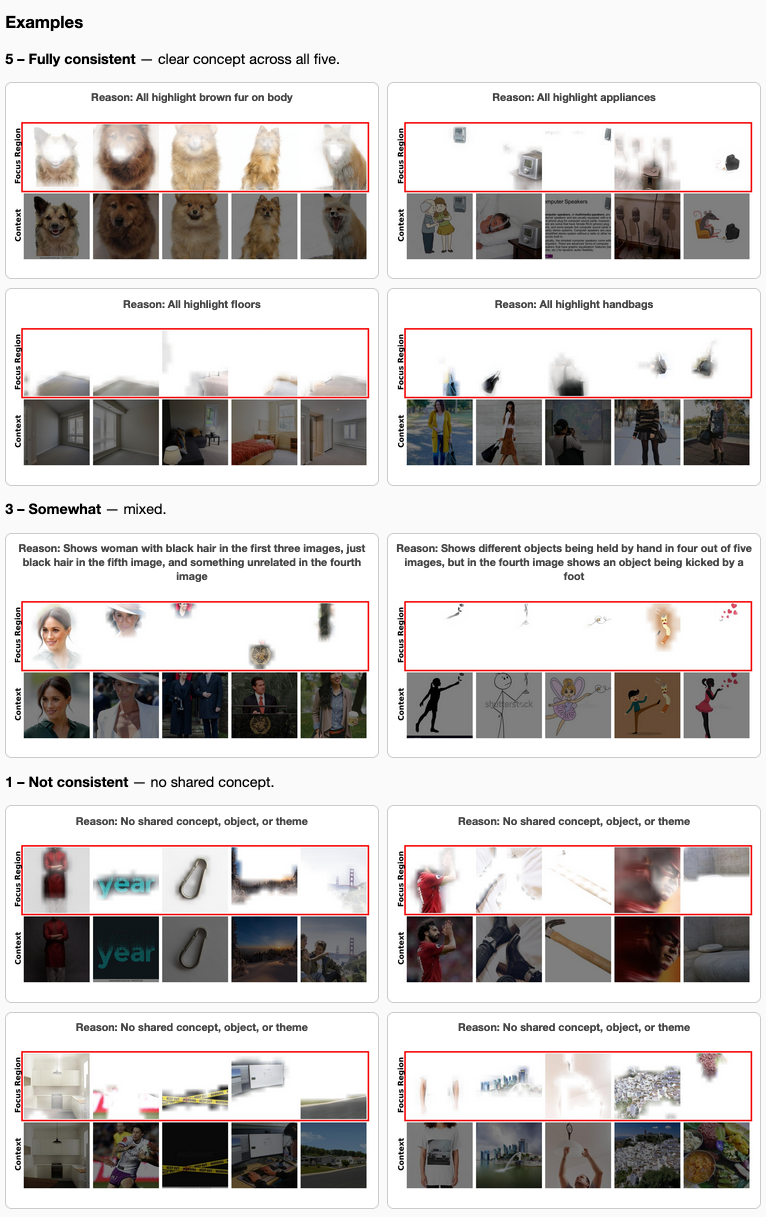}
        \caption{Examples provided after the instructions to guide annotators.}
    \end{subfigure}
    \caption{\textbf{Task 1 instructions and an example question.} See \cref{app:sec:experimental_details:user_study:setup} for details on the study setup.}
    \label{app:fig:user_study:task1}
\end{figure}

\paragraph{Task 1: Concept consistency.} In this task we ask annotators to rate the consistency of concepts from \ours on a five-point scale. Each question consists of the top five activating images and corresponding concept locality maps for an \ours SAE latent (a concept). We call concept locality maps "focus regions" to ensure easier understanding for layman. We bin latents into five equi-height bins based on frequency of their activation across images, and sample uniformly from each bin. We sample 908 concepts and additionally create 92 `random' questions for control, \ie questions where the 5 images and their focus regions are drawn iid from all SAE latents. For more details and an example, see \cref{app:fig:user_study:task1}.

\begin{figure}[!ht]
    \centering
    \begin{subfigure}{0.49\linewidth}
        \includegraphics[width=\linewidth]{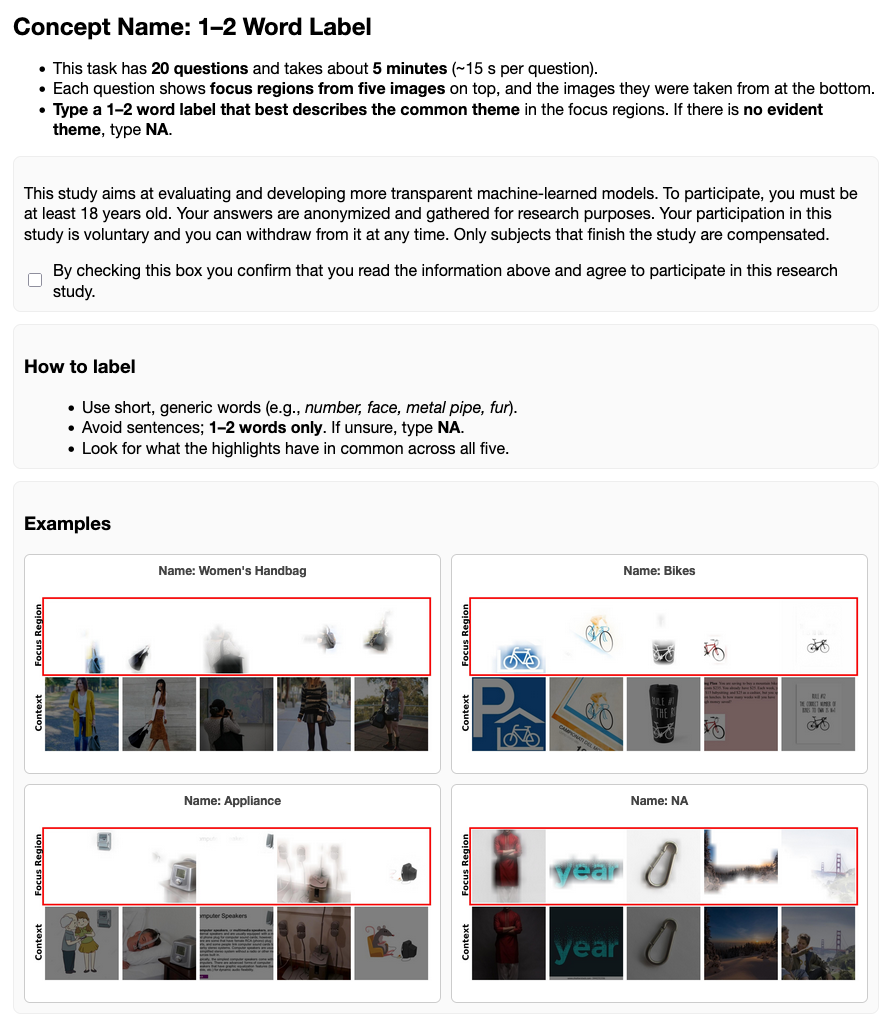}
        \caption{Task instructions along with examples to guide annotators.}
    \end{subfigure}
    \hfill
    \begin{subfigure}{0.49\linewidth}
        \includegraphics[width=\linewidth]{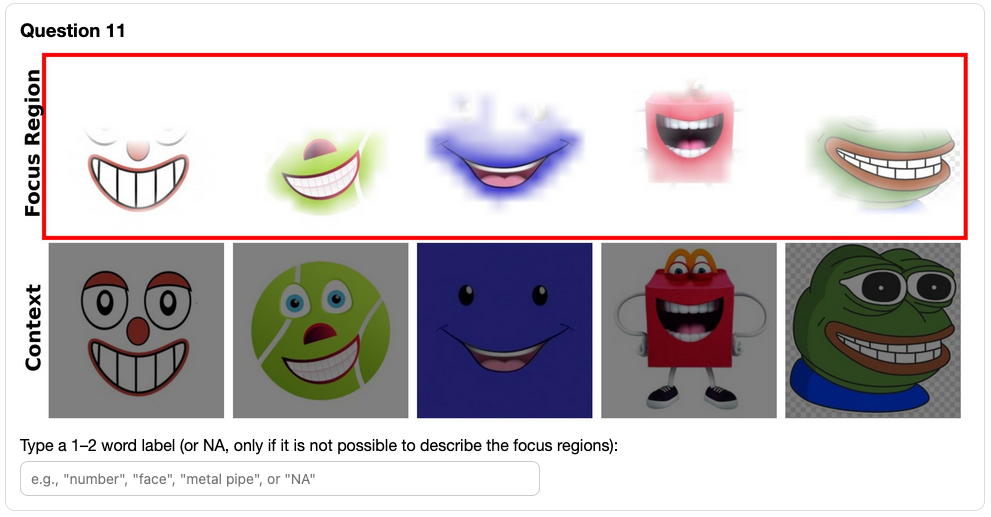}
        \caption{A question in the task.}
    \end{subfigure}
    \caption{\textbf{Task 2 instructions and an example question.} See \cref{app:sec:experimental_details:user_study:setup} for details on the study setup.}
    \label{app:fig:user_study:task2}
\end{figure}

\paragraph{Task 2: Concept naming quality.} In this task we ask annotators to provide a one to two word label for an \ours concept. Each question consists of the top five activating images and corresponding concept locality maps for an \ours SAE latent. We follow the same binning based on activation frequency as above and sample equally from each bin. We sample 908 concepts and additionally create 92 `random' questions for control, \ie questions where the 5 images and their focus regions are drawn iid from all SAE latents. If no name can be provided, annotators have the option to answer `NA'. For more details and an example, see \cref{app:fig:user_study:task2}.

\begin{figure}[!ht]
    \centering
    \begin{subfigure}{0.49\linewidth}
        \includegraphics[width=\linewidth]{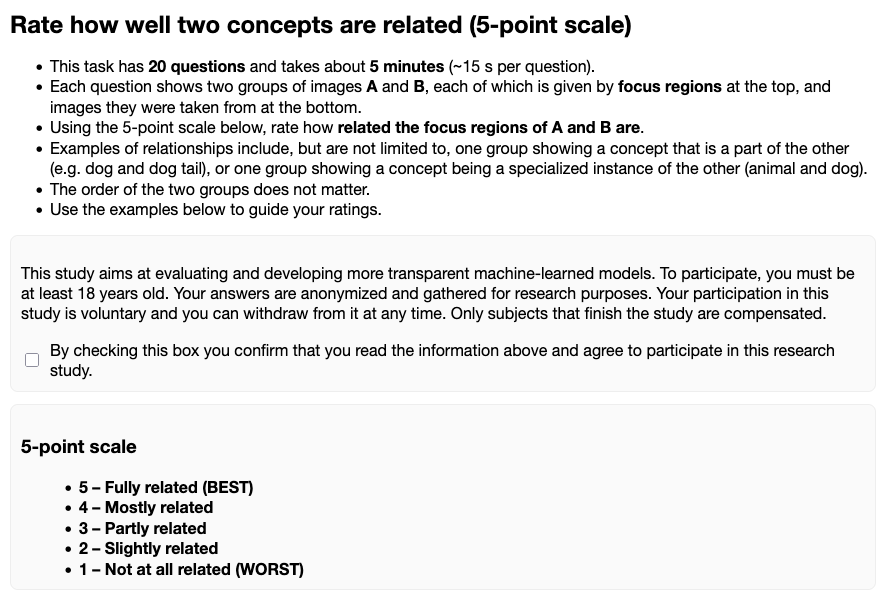}
        \vspace{10pt}
        \includegraphics[width=\linewidth]{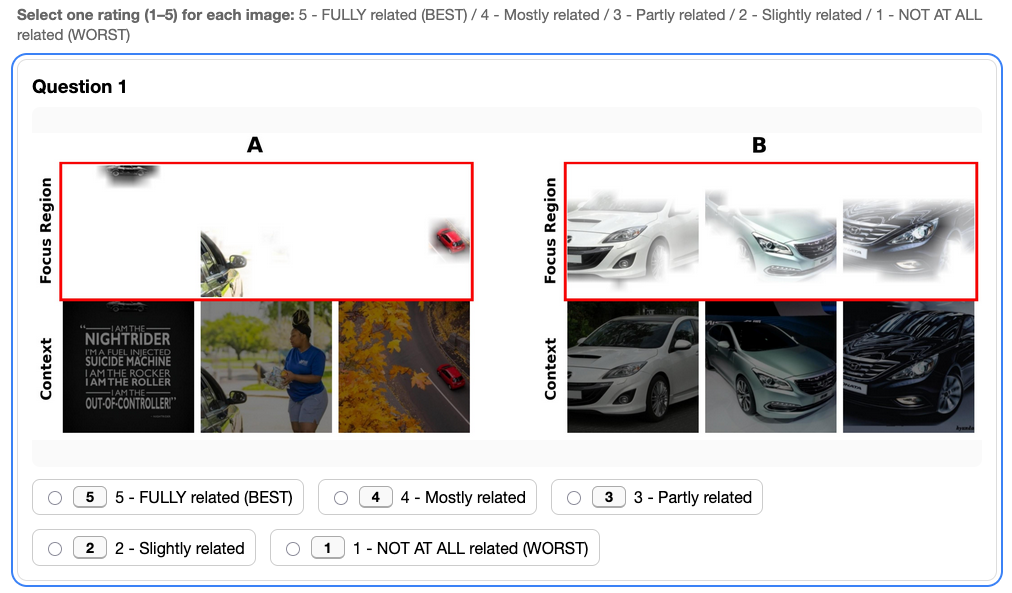}
        \caption{\emph{Top:} Task instructions. \emph{Bottom:} A question in the task.}
    \end{subfigure}
    \hfill
    \begin{subfigure}{0.49\linewidth}
        \includegraphics[width=\linewidth]{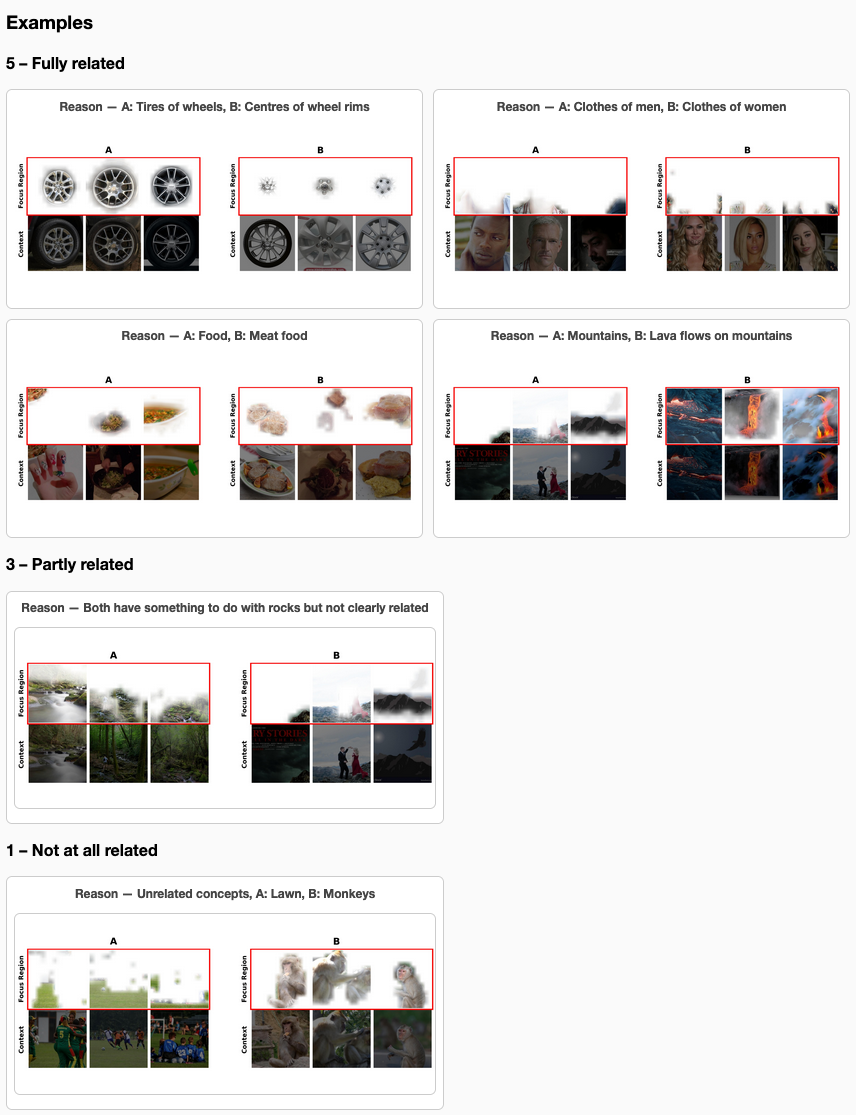}
        \caption{Examples provided after the instructions to guide annotators.}
    \end{subfigure}
    \caption{\textbf{Task 3 instructions and an example question.} See \cref{app:sec:experimental_details:user_study:setup} for details on the study setup.}
    \label{app:fig:user_study:task3}
\end{figure}

\paragraph{Task 3: Concept relationship quality.} In this task we ask annotators to rate a given relationship between two concepts from \ours on a five-point scale, and we use this task to evaluate the quality of hierarchical relationships from \ours. Each question consists of the top three activating images and corresponding concept locality maps for two SAE latents from \ours. We bin latents into five bins based on frequency of their activation across images, and sample equally from each bin. We sample 990 concept pairs that correspond to edges from the \ours concept hierarchies and additionally create 110 `random' relations across all concepts that do not have a real relation in our graph $\mathcal{G}$ for control. For more details and an example, see \cref{app:fig:user_study:task3}.

\subsubsection{Quality Filtering.}
We apply three types of filtering to obtain consensual and higher quality results.
For all tasks, we exclude those participants that did not answer the box asking for consent to being a participant. For task 1 and task 3, respectively, we filter out those participants that gave a score $>3$ for one of the random control questions, as they likely did not understand the task or answered randomly.
For task 2, we exclude those participants that gave NA as label for almost every question.
This leaves us with scores of 152 unique participants for task 1, 107 for task 2, and 158 for task 3.

\subsubsection{Statistics Stratified by Activation Frequency.}
\label{app:sec:experimental_details:user_study:results}

\begin{figure}
\centering

\begin{subfigure}[b]{0.55\textwidth}
    \centering
    \includegraphics[width=\textwidth]{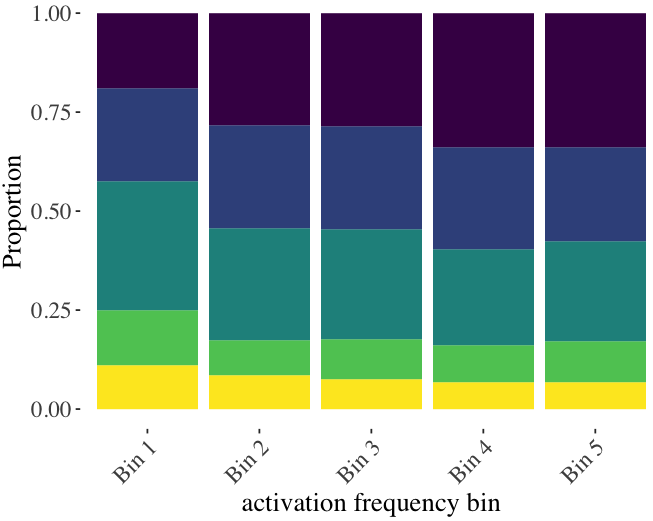}
    \caption{Task 1 -- Concept consistency per bin.}
    \label{app:fig:userstudy1perbin}
\end{subfigure}
\hfill
\begin{subfigure}[b]{0.44\textwidth}
    \centering
    \includegraphics[width=\textwidth]{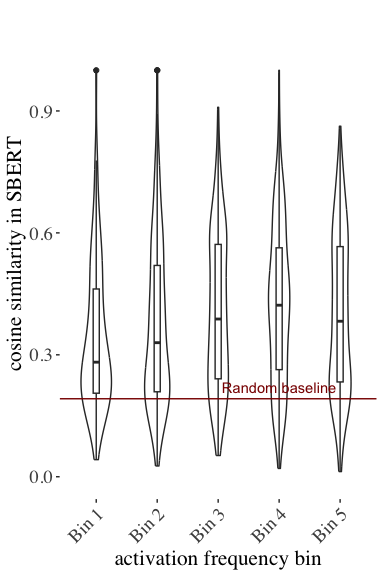}
    \caption{Task 2 -- Concept naming quality per bin.}
    \label{app:fig:userstudy2perbin}
\end{subfigure}

\caption{\textbf{Additional User Study Results.} We provide (a) concept consistency scores partitioned into concept activation frequency bins, with scores 1 (no consistency, yellow) to 5 (full consistency, purple) and (b) concept naming quality cosine similarity scores partitioned into concept activation frequency bins, red line indicates mean random similarity.}\label{app:fig:studyresults}
\end{figure}

To investigate how activation frequency of a concept, which is a reflection of its generality and correlates with the Matryoshka shells, impacts human interpretability, we investigate how stratification by activation frequency bins impacts study results. For both task 1 and 2 (see ~\cref{app:fig:studyresults}), we observe that results look largely consistent across bins with slight improvements in scores for less frequently activated bins. In other words, we see a slight (negative) correlation between activation frequency and scores. This is to be expected, as frequently activated, general concepts activate on more diverse images, often capture more abstract or high-level concepts that are seemingly less consistent compared to the more specific concepts, given the larger variation in images. Similarly, it is harder to identify a clear and crisp label for a high-level concept compared to a highly specific one.

\subsection{Relationship Evaluation Using an Automated Judge}
\label{app:sec:experimental_details:relationship_judge}
As discussed in \cref{sec:conceptquality}, we evaluate the accuracy of concept relationship directions provided by \ours using an automated judge. Specifically, we sample 1000 concept pairs that are assigned a parent-child relationship by \ours, and for each pair, we obtain the top-5 activating images and their corresponding activating regions. We construct a composite image for each concept, and provide them to a gpt-5.5~\cite{singh2025openai} judge, asking it to (1) evaluate if both concepts are consistent, and if so (2) the directionality of the relation between them, if any. In other words, given concept A marked as a parent of concept B, the model is asked to evaluate if A is indeed a parent of B as opposed to B being a parent of A. The prompt is provided in \cref{fig:system-prompt_relationship}.

\begin{figure}[!ht]
    \centering
    \includegraphics[width=\linewidth]{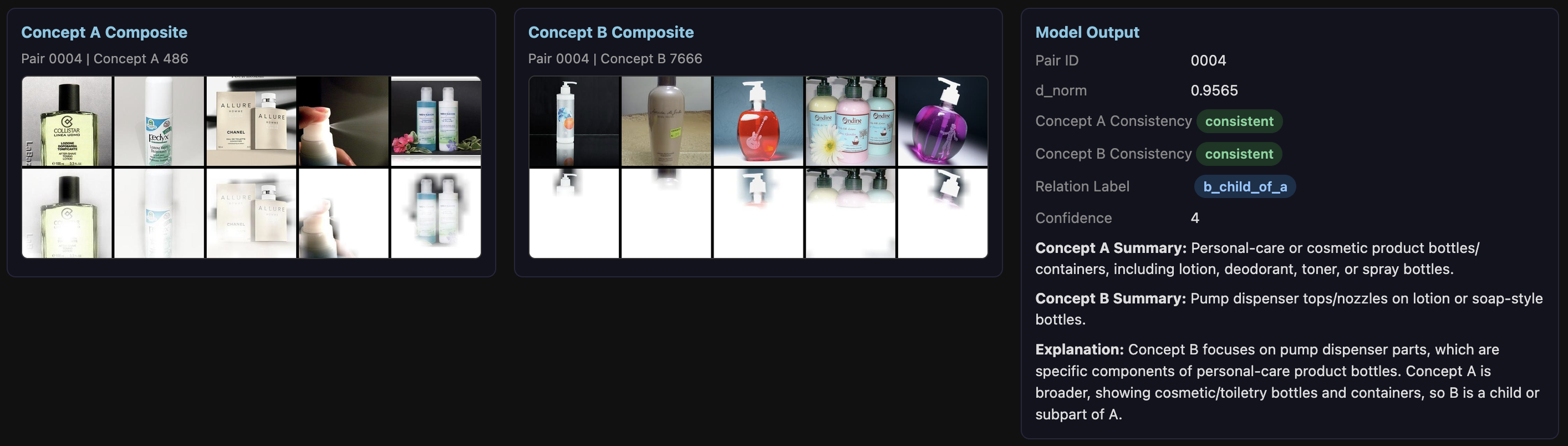}
    \caption{\textbf{Example of concept pairs and the output of the automated judge.} See \cref{app:sec:experimental_details:relationship_judge} for details.}
    \label{app:fig:relationship_example}
\end{figure}

We find that among consistent prompts with a clear relationship, the directionality provided by \ours is accurate for 79.79\% of pairs. Specifically, out of 1000 pairs, the judge found 727 to have both concepts consistent. Out of these, 376 pairs were found to have a clear directionality, and 300/376 had a directionality matching that provided by \ours. The strict filtering process ensures that the judge correctly understood the concepts before evaluating their relationship. A qualitative example of concept pairs and the output by the judge is provided in \cref{app:fig:relationship_example}.

\begin{figure}[t]
\centering
\begin{Verbatim}[
    frame=single,
    breaklines=true,
    breakanywhere=true,
    fontsize=\scriptsize,
    breaksymbolleft=,
    breaksymbolright=,
]
# Role
You are a visual evaluator whose task is to judge whether two visually inferred concepts stand in a parent-child relation, and in which direction.

# Instructions
You will be given exactly two images in this exact order:
1. a concept A composite
2. a concept B composite

Each composite corresponds to K = 5 examples arranged in a 2 x 5 grid:
- top row: the 5 original example images
- bottom row: the 5 corresponding focus images

Within each composite, each focus image is created by blending the original image with white according to a heatmap. Pixels with high heatmap strength remain close to the original image, while pixels with low heatmap strength become washed out toward white.

Your task:
1. Infer whether the concept A examples have a dominant visual concept, focusing primarily on the parts of the focus images that remain most visible and least washed out.
2. Infer whether the concept B examples have a dominant visual concept, focusing primarily on the parts of the focus images that remain most visible and least washed out.
3. If both concept A and concept B examples have a dominant visual concept, judge whether:
- `b_child_of_a`: concept B is a child of concept A.
- `a_child_of_b`: concept A is a child of concept B.
- `unclear_relation`: the evidence is insufficient to decide, or there is no clear parent-child relation in either direction.

A concept is a child of another if it is a specification or subpart of the other. For example, if concept A is "Dog" and concept B is "Golden Retriever", then B is a child of A. If concept A is "Dog" and concept B is "Tail", then B is a child of A. If concept A is "Dog" and concept B is "Car", then there is no parent-child relation.

Base your answer only on the visual evidence in the examples and their focus images. Focus mainly on what remains visible in the focus images. Use the original images mainly for context. Do not rely on assumptions that are not supported by the images.

Decision rule:
- Mark a concept as `consistent` if the examples support a dominant shared concept in the parts of the focus images that remain most visible.
- Mark a concept as `inconsistent` only if you cannot identify a dominant shared concept from the parts of the focus images that remain most visible.
- Do not treat washed-out or nearly white regions as the concept.
- If either concept A or concept B is inconsistent, set `relation_label` to `unclear_relation`.
- Even if a concept is inconsistent, still provide a short best-effort `concept_a_summary` or `concept_b_summary` describing the dominant pattern or why the concept is mixed.
- If either concept is inconsistent, use `relation_explanation` to explain that the relation cannot be judged reliably because the concept evidence is not coherent enough.

Output:
- Return JSON only.
- Use exactly these fields:
  - `concept_a_consistency`
  - `concept_b_consistency`
  - `concept_a_summary`
  - `concept_b_summary`
  - `relation_label`
  - `confidence`
  - `relation_explanation`

Note:
- `concept_a_consistency` and `concept_b_consistency` must each be either `consistent` or `inconsistent`.
- `relation_label` must be one of `b_child_of_a`, `a_child_of_b`, or `unclear_relation`.
- `confidence` must be an integer from 1 to 5, representing confidence in the overall judgment, especially the consistency labels and relation label.
- `concept_a_summary` and `concept_b_summary` should each be a short explanation of what the concept represents.
- `relation_explanation` should be a short explanation of the relation decision. If a concept is inconsistent, explain that clearly.
\end{Verbatim}
\caption{\textbf{Prompt provided to the automated judge to evaluate relationship directionality.} For details, see \cref{app:sec:experimental_details:relationship_judge}.}
\label{fig:system-prompt_relationship}
\end{figure}

\subsection{Naming Evaluation Using an Automated Judge}
\label{app:sec:experimental_details:naming_judge}

As discussed in \cref{sec:conceptquality} and \cref{fig:name_accuracy}, we evaluate the accuracy of names assigned by \ours in comparison to baselines using an automated judge. Specifically, we compare against DN-CBM~\cite{rao2024discover} and MSAE~\cite{zaigrajew2025interpreting}, since they both provide a concept decomposition that can be reconstructed to the original embedding, preserving the utility of the foundation model backbone. For each method, we obtain the top-10 concept names for each image from the MS COCO~\cite{lin2014microsoft} validation set, and evaluate \textit{whether the concept is present in the image}. For DN-CBM and MSAE, we additionally subtract the mean activation strength of each concept before ranking the top concepts per image, following the procedure used in those works.

Note that this evaluates concept presence unlike concept bottleneck evaluations (\eg \cref{fig:classification}), which instead evaluate how much a concept contributed to a specific downstream classification task. We provide pairs of images and individual concepts to a gpt-5-mini~\cite{singh2025openai} judge, and ask it to decide if that concept is present in the image. The prompt used is provided in \cref{fig:system-prompt}.

Additionally, since each method uses a different concept set and a different vocabulary, we also evaluate with a random baseline for each method. For this, we sample ten concepts uniformly at random for each image from within the concept sets of each method. This helps better understand the true performance of each method, since using a concept set that has poor diversity but contains concepts that may be found in typical COCO classes could show a high accuracy by random chance. Nevertheless, concepts by \ours are rated the most accurate by a significant margin, with a 42.8pp improvement over the random baseline (\cref{fig:name_accuracy}) as compared to 26.2pp for MSAE and 20.4pp for DN-CBM, respectively. For additional qualitative examples, see \cref{app:sec:namecompare}.

\begin{figure}[t]
\centering
\begin{Verbatim}[
    frame=single,
    breaklines=true,
    breakanywhere=true,
    fontsize=\small,
    breaksymbolleft=,
    breaksymbolright=,
]
# Role
You are a visual evaluator whose task is to judge whether an image contains a particular concept.

# Instructions
You will be given an:
- an image
- a text concept (a single word or short phrase)

Determine if there is anything in the image that looks like the concept.

Output:
- 1 if the image contains the concept.
- 0 otherwise.

Do not output anything else.
\end{Verbatim}
\caption{\textbf{Prompt provided to the automated judge to evaluate concept naming.} For details, see \cref{app:sec:experimental_details:naming_judge}.}
\label{fig:system-prompt}
\end{figure}

\subsection{Open-Vocabulary Segmentation Inference}
\label{app:sec:experimental_details:ovs}
We follow the standard MMSegmentation~\cite{mmseg2020} sliding window inference protocol. While images are generally resized to a shorter side of 448 pixels and processed in $448 \times 448$ windows. For high-resolution datasets (Cityscapes, ADE20K, and COCO-Stuff) we pass images close to their original resolution. This prevents the loss of fine details and improves the performance of both \ours and CLIP-DINOiser~\cite{wysoczanska2024clipdinoiser}. Additionally, we utilize the background filtering strategy from CLIP-DINOiser to suppress non-foreground regions on datasets with a background prompt.

%%%%%%%%%%%%%%%%%%%%%%%%%%%%%%%%%%
\clearpage

\section{Training Details}
\label{app:sec:training}
This section provides detailed training procedures for all components of \ours. We follow the notation established in the main paper (\cref{sec:method}), recap the CLIP-DINOiser training, with a focus on the lightweight convolution, in Sec.~\ref{app:sec:dinoiser_training}, discuss SAE training in Sec.~\ref{app:sec:sae_training}, hierarchy discovery in Sec.~\ref{app:sec:family_training}, concept naming in Sec.~\ref{app:sec:naming_details}, classifier training in Sec.~\ref{app:sec:probe_training}, training for image captioning in Sec.~\ref{app:sec:captioning}, open-vocabulary segmentation in Sec.~\ref{app:sec:Open-Vocabulary segmentation}, and hyperparameter ablation in Sec.~\ref{app:sec:hyperparameter_ablation}.

\subsection{CLIP-DINOiser Training}
\label{app:sec:dinoiser_training}

Following Wysoczańska \etal~\cite{wysoczanska2024clipdinoiser}, we train a lightweight 3×3 convolutional layer $g: \mathbb{R}^d \rightarrow \mathbb{R}^{d'}$ with $d = 512$ and $d' = 64$ to approximate DINO affinity patterns from CLIP patch tokens. 

\paragraph{Training objective.} The convolution $g$ is trained to predict binarized DINO~\cite{caron2021emerging} affinities. Given patch tokens $X$ from the final CLIP layer $L$, we compute predicted affinities as follows:
\begin{equation}
A^{\phi}=\frac{g(\phi^L(X))}{\left\Vert g(\phi^L(X))\right\Vert} \otimes \left(\frac{g(\phi^L(X))}{\left\Vert g(\phi^L(X))\right\Vert}\right)^T.
\end{equation}
The training loss is binary cross-entropy between $A^{\phi}$ and binarized DINO affinities $D = A^{\xi} > \gamma$:
\begin{equation}
\mathcal{L}^{\text{C}} = \sum_{p=1}^N \left[ D_{p} \log A^\phi_{p} + (1 - D_{p}) \log ( 1 - A^\phi_{p}) \right],
\end{equation}
where $\gamma = 0.2$ is the binarization threshold.

\paragraph{Training setup.} We train on 1,000 randomly sampled images from ImageNet~\cite{imagenet} under learning rate of $5e-5$ with batch size of 16 for 100 epochs.
The CLIP ViT-B/16 backbone $\phi$ and DINO ViT-B/16 remain frozen throughout training. For more details see Wysoczańska \etal~\cite{wysoczanska2024clipdinoiser}.

\subsection{Matryoshka SAE Training}
\label{app:sec:sae_training}

We train hierarchical Matryoshka Sparse Autoencoders with BatchTopK sparsity~\cite{bussmann2024batchtopk} on DINOised patch features $F^+$ extracted as described in \cref{sec:msae} of the main paper.

\paragraph{Architecture.} The SAE consists of encoder $\pi:\mathbb{R}^d\rightarrow\mathbb{R}^{m}$ and decoder $\pi^{-1}:\mathbb{R}^{m}\rightarrow\mathbb{R}^d$ where $d = 512$ is the CLIP embedding dimension and $m = 8192$ is the number of concepts (expansion factor of 16).

For a patch embedding $F_p^+ \in \mathbb{R}^d$, the encoder computes pre-activation features:
\begin{equation}
z_{\text{pre}} = W_{\text{enc}}^T (F_p^+ - b_{\text{dec}}) + b_{\text{enc}}\;,
\end{equation}
where $W_{\text{enc}} \in \mathbb{R}^{d \times m}$ are encoder weights, $b_{\text{enc}} \in \mathbb{R}^m$ is the encoder bias, and $b_{\text{dec}} \in \mathbb{R}^d$ is the decoder bias (shared with the decoder).

\paragraph{BatchTopK sparsity.} Following Bussmann \etal~\cite{bussmann2024batchtopk}, we apply batch-level top-$k$ activation, which retains exactly $B \cdot k$ neurons across a batch of size $B$:
\begin{equation}\label{app:eq:z}
z = \pi(F_p^+) = \text{ReLU}(\text{BatchTopK}(z_{\text{pre}}))
\end{equation}
where BatchTopK sets all but the top $B \cdot k$ pre-activations to zero across the entire batch. We use $k = 12$, meaning the SAE activates 12 concepts per patch on average.

\paragraph{Matryoshka structure.} Following \cref{sec:msae} of the main paper, we organize the $m = 8192$ concepts into six hierarchical groups with size ratios $[0.008, \allowbreak 0.03, \allowbreak 0.06, \allowbreak 0.12,  \allowbreak0.24, \allowbreak 0.543]$. Let $C = [1 \ldots m]$ be the index set and $l_j \in C$ the index defining Matryoshka shell $j$. The reconstruction loss from the main paper is:
\begin{equation}
\mathcal{L}_{\text{rec}}(F_p^+) = \sum_j \left\Vert \pi^{-1}(\pi(F_p^+)[1:l_j]) - F_p^+ \right\Vert_2^2\;,
\end{equation}
where $\pi(F_p)[1:l_j]$ sets all activations after index $l_j$ to zero.

\paragraph{Auxiliary loss for dead features.} To recover neurons that haven't activated recently (last 10,000,000 samples), we add an auxiliary loss~\cite{gao2024scaling}:
\begin{equation}
\mathcal{L}_{\text{aux}} = \frac{\|r - W_{\text{dec}}^T z_{\text{dead}}\|_2^2}{\|r - \bar{r}\|_2^2}\;,
\end{equation}
where $r = F_p^+ - \pi^{-1}(\pi(F_p))$ is the reconstruction residual, $\bar{r}$ is the batch-mean residual, and $z_{\text{dead}}$ are activations for dead features only. Hence the total loss is
\begin{equation}
\mathcal{L}_{\text{total}} = \mathcal{L}_{\text{rec}} + \alpha_{\text{aux}} \cdot \mathcal{L}_{\text{aux}} \,
\end{equation}
with $\alpha_{\text{aux}} = \frac{1}{32}$.

\paragraph{Training setup.} We train on all DINOised patch embeddings extracted from CC12M~\cite{changpinyo2021conceptual}. Each image yields $N = 196$ patches (14×14 grid), providing approximately 2 billion training samples. We use Adam optimizer~\cite{kingma2014adam}, with learning rate of $1e-4$. We set the batch size to 16,268 from 83 images and train for 3 epochs. We keep 10\% of the patches as held-out validation set. Our training converges within 3 epochs with fraction of variance explained reaching approximately 73\% and number of dead features dropping below 10.

\subsection{Hierarchy Discovery}
\label{app:sec:family_training}

After SAE training, we discover concept relationships by analyzing patch-level co-occurrence patterns as described in \cref{sec:coocc} of the main paper. Note that during inference, we consider a concept $j$ to be active if its activation exceeds an estimated threshold, which is derived to mimic the BatchTopK sparsity enforced during training.

\paragraph{Confidence matrix computation.} For each concept pair $(i,j)$, we compute the weighted conditional probability $C_{ij}$ as defined in \cref{hierarchy_matrix} using the SAE training dataset (CC12M~\cite{changpinyo2021conceptual}).

\paragraph{Graph construction.} We construct the hierarchy graph $\mathcal{G} = (V, E)$ where $V = \{c_1, c_2, \dots, c_m\}$, i.e., vertices are concepts, and
 $(i,j) \in E$ if $C_{ij} \geq \tau$.
This graph $\mathcal{G}$ defines our hierarchies, where each edge represents a parent to child (coarse to fine-grained) relationship. In this work we use $\tau=0.75$. The threshold is an SAE-dependent hyperparameter determined via qualitative validation of the resulting hierarchies, selecting the lowest value that avoids spurious child relationships. We find that this value is stable and transfers well across different SAEs and datasets. In practice, $\mathcal{G}$ is typically a Directed Acyclic Graph (DAG) with potentially multiple parents per child.

\subsection{Concept Naming}
\label{app:sec:naming_details}

Following \cref{sec:labeling} of the main paper, we assign concept names using hierarchy-aware vector reconstruction.

\begin{figure*}[t]
\centering
\includegraphics[width=.95\linewidth]{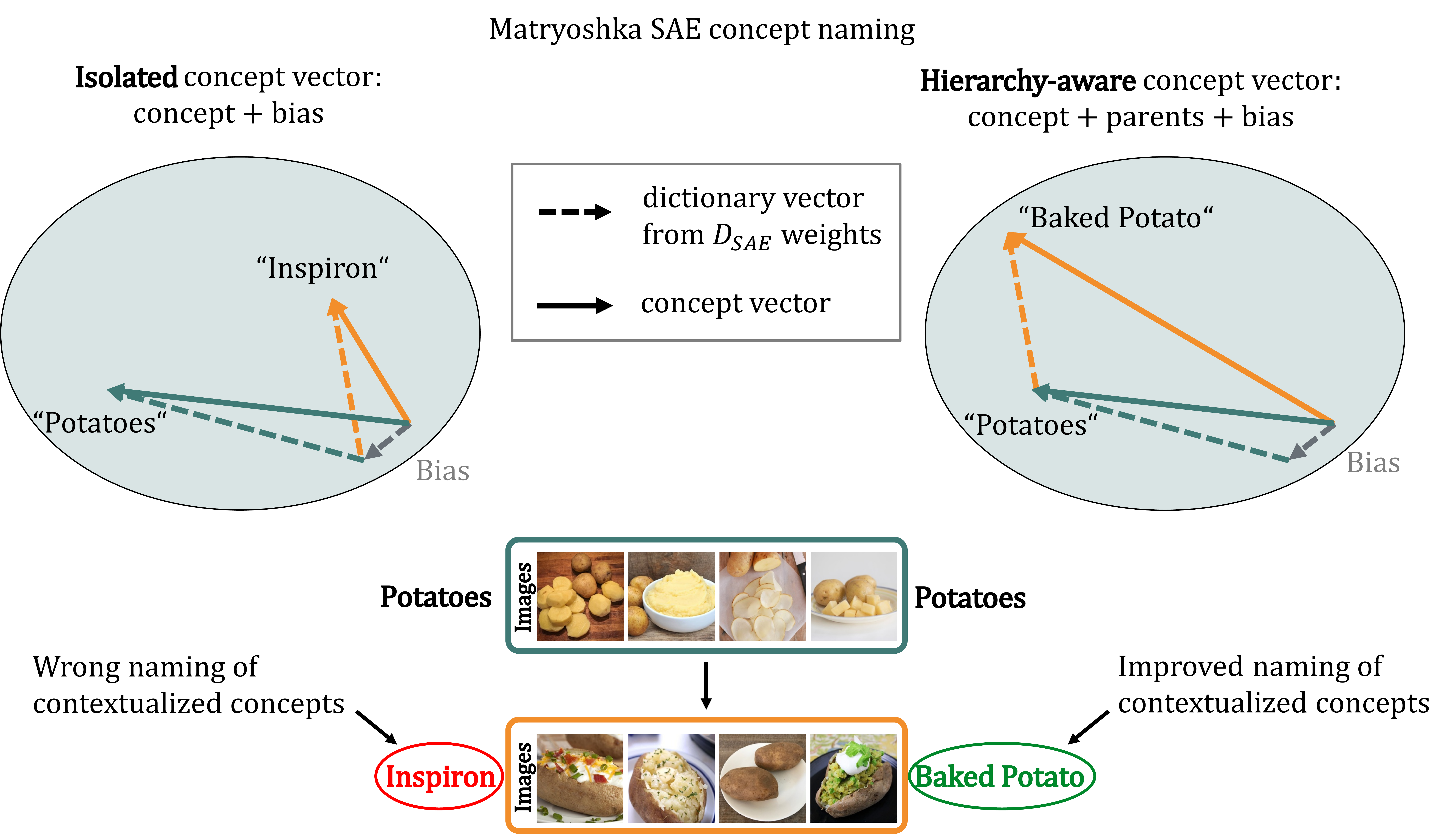}
\caption{\textbf{Adapted concept vector reconstruction.} In Matryoshka SAEs, dictionary vectors across different Matryoshka shells are residual by design. This means fine-grained concepts (e.g. "Baked Potato") are contextualized by coarser parent concept (e.g., "Potatoes"). Consequently, evaluating a fine-grained dictionary vector in isolation yields a misaligned representation (left), often resulting in erroneous textual labels such as "Inspiron" (bottom left). To resolve this, we reconstruct the concept vector by summing the target's dictionary vector with the vectors of all its parent concepts (right), which are identified as ancestors in our relationship graph (\cref{sec:coocc}). This hierarchy-aware reconstruction is essential for assigning accurate, meaningful text labels to fine-grained concepts (bottom right).}
\label{fig:concept_reconstruction}
\end{figure*}

\paragraph{Vector reconstruction.} For fine-grained concepts in later Matryoshka groups, dictionary vectors $\pi^{-1}_i$ alone are insufficient as they encode residuals from parent concepts (see \cref{fig:concept_reconstruction}). We reconstruct hierarchy-aware concept vectors:
\begin{equation}
v_{c_i} = b + \alpha \cdot ( w_{ii} \cdot \pi^{-1}_i + \sum_{a \in \mathcal{G}, a \rightarrow i} D_{ia} \cdot w_{ia} \cdot \pi^{-1}_a ),
\end{equation}
where:
\begin{itemize}
    \item $b$ is the SAE decoder bias $b_{\text{dec}}$
    \item $\pi^{-1}_i$ is the dictionary vector (decoder weights for concept $i$)
    \item $a \rightarrow i$ denotes parent $a$ of concept $i$ in graph $\mathcal{G}$
    \item $D_{ia}$ is the confidence from the co-occurrence matrix
    \item $w_{ia}$ is the mean activation of concept $a$ in the top-30 most activating patches of concept $i$
    \item $\alpha = 1.33$ is an empirically chosen coefficient to strengthen the concept direction relative to the decoder bias
\end{itemize}
% The coefficient $\alpha > 1$ ensures concept $i$'s dictionary vector receives stronger weight than parent vectors, maintaining specificity while incorporating parent context.

\paragraph{Vocabulary matching.} Given reconstructed vector $v_{c_i}$, we assign names as in \cref{sec:labeling}:
\begin{equation}
l_i = \text{argmax}_{v \in V} \cos(v_{c_i}, \mathcal{T}(v)) ,
\end{equation}
where $\mathcal{T}$ is the CLIP text encoder and $V$ is our vocabulary.\\
As such, we get meaningful labels across concepts of different granularity. This assumes that the query vocabulary $V$ is rich enough. We found that vocabularies used in concurrent works~\cite{zaigrajew2025interpreting,rao2024discover} are often too limited for fine-grained concept labeling.

\paragraph{Vocabulary construction.}
To ensure comprehensive coverage, we construct our vocabulary by combining multiple resources. 
Concretely, our vocabulary includes (1) \texttt{Common words} as the 20k most frequent English words from \cite{oikarinen2022clip}, (2) a filtered 40k-word dataset from Brysbaert \etal~\cite{brysbaert2014concreteness}, in which each word is annotated with a concreteness score between 1 (abstract) and 5 (highly concrete) and we retain only words with a mean concreteness above 2.5 as concreteness correlates strongly with imageability, (3)  multi-word expressions from Muraki \etal~\cite{muraki2023concreteness}, filtering with a concreteness threshold of 2.0, 
and (4) LAION expressions following the approach of Bhalla \etal~\cite{bhalla2024interpreting}, extracting the 40k most frequent uni-words from LAION-400M, filtering NSFW captions using the given NSFW flag. Furthermore, we include the bi-words extracted by Bhalla \etal~\cite{bhalla2024interpreting}.
After gathering all resources, we merge them into a single comprehensive candidate vocabulary that balances coverage, concreteness, and fine-grained expressivity and provides a good resource for meaningful concept naming.

\paragraph{Vocabulary Templates.}
To improve labeling accuracy, we use sentence templates to encode the query vocabulary. However using the standard Zero-shot templates biases the labeling toward nouns, as CLIP's text encoder is sensitive to grammatical context. To allow concepts to be labeled by nouns, verbs, or adjectives, we introduce three distinct template sets, one for each part of speech. Each set contains ten abstract templates $\{\tau^{(p)}_k\}_{k=1}^{10}$, where $p \in \{\text{noun}, \text{verb}, \text{adj}\}$. Example templates include patterns such as ``a photo of a \{\}'' (noun), ``a photo of someone \{\}'' (verb), or ``something that is \{\}'' (adjective). We deliberately avoid overly specific templates to prevent unintentionally biasing the meaning of vocabulary items.

Given a vocabulary item $v \in V$, we generate its template-conditioned embeddings by applying the CLIP text encoder $\mathcal{T}$ to each templated prompt:
\begin{equation}
E_p(v)
= \frac{1}{10} \sum_{k=1}^{10}
\frac{\mathcal{T}\!\left(\tau^{(p)}_k(v)\right)}
{\left\lVert \mathcal{T}\!\left(\tau^{(p)}_k(v)\right)\right\rVert_2},
\qquad 
p \in \{\text{noun},\text{verb},\text{adj}\}.
\end{equation}
Each $E_p(v)$ is then normalized to unit length. For every concept vector $v_{c_i}$, we compute the cosine similarity to all template-conditioned embeddings:
\begin{equation}
s_{i}(v)
= \max_{p \in \{\text{noun},\text{verb},\text{adj}\}}
\cos\!\left(v_{c_i},\, E_p(v)\right).
\end{equation}
Finally, we assign the vocabulary label that achieves the highest similarity:
\begin{equation}
l_i = \operatorname*{arg\,max}_{v \in V} s_i(v).
\end{equation}
This template-based strategy enables each concept to select the grammatical form that best fits its semantics, yielding more accurate and fine-grained names than noun-only matching.

\clearpage
\subsection{Probing for Classification}
\label{app:sec:probe_training}

For downstream classification tasks, we train a linear probe on image-level concept activations while keeping all other components frozen (i.e., not fine-tuning).
\paragraph{Concept aggregation.} We aggregate patch-level concept activations to image-level by combining max pooling and mean pooling:
\begin{equation}
\hat{c}^{\text{max}}_i = \max_{p \in \{1, \ldots, N\}} z_{i,p},
\qquad
\hat{c}^{\text{mean}}_i = \frac{1}{N} \sum_{p=1}^{N} z_{i,p},
\end{equation}
and form the final image-level concept vector by concatenating both:
\begin{equation}
\hat{c}_i =
\left[
\hat{c}^{\text{max}}_i | \hat{c}^{\text{mean}}_i
\right],
\end{equation}
where $z_{i,p} \in \mathbb{R}^m$ are the sparse concept activations given by our SAE concept mapping (see \cref{app:eq:z}) for patch $p$ of image $i$, and $N = 196$ is the number of patches. We find that using only max-pooled concepts yields comparable performance: on ImageNet we match the reported accuracy, and on Places365, we observe only a $0.4\%$ reduction.

\paragraph{Linear probe.} We train a sparse linear classifier $h: \mathbb{R}^m \to \mathbb{R}^{|Y|}$ with $\ell_1$ regularization:
\begin{equation}
\mathcal{L}_{\text{probe}} = \text{CE}(h(\hat{c}_i), y_i) + \lambda \|\omega\|_1 ,
\end{equation}
where $\omega$ are the probe weights, $y_i$ is the class label of image $i$, CE is cross-entropy loss, and $\lambda$ controls the L1 sparsity loss.
\paragraph{Training setup.} We use AdamW optimizer~\cite{kingma2014adam} with $\beta_1 = 0.9$, $\beta_2 = 0.999$ and set the learning rate to $1e-4$. We use batch size of 1024 and train for 100 epochs, taking the highest accuracy checkpoint. For the hyper-parameter search, we perform the following sweep over a held-out subset ($10\%$ of training data).\[
\text{batch size} \in \{256,\, 512,\, 1024\},
\qquad\qquad
\text{learning-rate} \in \{1e-3,\, 5e-4,\, 1e-4\},
\qquad
\]
\[
\lambda_{\mathrm{L1}} \in \{0.1,\, 1.0,\, 2.0,\, 4.0,\, 8.0\}.
\]

\noindent All other components (CLIP encoder $\phi$, DINOiser $g$, SAE $\pi$ and $\pi^{-1}$) remain frozen.

At inference, we do not apply the activation threshold learned during BatchTopK training, this keeps our setup comparable to the other CBMs.

\subsection{Image Captioning}
\label{app:sec:captioning}
To explore the capabilities of \ours for downstream captioning task, we followed the setup of LLaVA training ~\cite{liu2023InstructTune}, mapping the dense output tokens to input words for a language model via an MLP adapter. Specifically, we took the 14$\times14$ dense tokens after SAE reconstruction and used a 2-layer MLP to project them into 196 input words for an instruction-tuned Gemma-2 model~\cite{gemmateam2024gemma2improvingopen} with 2 billion parameters.

\paragraph{Training.} We trained our captioning setup (the adapter + the Gemma model) on CC12M DreamLIP synthetic captions~\cite{changpinyo2021conceptual,Zheng2024DreamLIP}. Specifically, for every image we randomly sampled one of the long synthetic captions from the dataset. We used an effective batch-size 96 for 30K steps. We split the training into two stages, where the first 13K steps only trained the MLP adapter under cosine learning-rate schedule (1e-2$\rightarrow$1e-5) and in the second stage, the MLP and Gemma were jointly trained for 17K steps under fixed learning-rate of 1e-5. We used LoRA~\cite{hu2022lora} (r=16, $\alpha$=32) for updating Gemma parameters. Overall, we explored a lightweight captioning setup, with each training taking about 22 hours on a single H100 GPU. We note that we kept the backbone vision model, as well as the concept representation (SAE) frozen (i.e., no fine-tuning to this task).

We performed the above procedure both for \ours, as well as the same opaque backbone~\cite{wysoczanska2024clipdinoiser} without the concept representation of the SAE. The results from \cref{fig:captioning}, which were demonstrated on unseen COCO~\cite{lin2014microsoft} test-set (under Karpathy split) demonstrated that \ours achieves the same performance in terms of captioning, while additionally allowing for steering the generations through the named concepts~\cref{fig:captioning,fig:steer-supplement}. 

\subsection{Open-Vocabulary Segmentation}
\label{app:sec:Open-Vocabulary segmentation}
\subsubsection{Patch-wise Concept Contributions.}
In our standard inference approach, the vision encoder processes the image into a grid of spatial patch tokens. Passing these tokens through the SAE yields a sparse concept activation vector for each individual patch. To perform segmentation, we reconstruct the dense CLIP embedding for each patch using the SAE decoder and compute its cosine similarity against the text embeddings of the target vocabulary.Because the reconstructed embedding is a linear combination of the active concepts, we can directly trace the model's decision back to individual concepts. We define the patch-wise concept contribution to a specific segmentation label as the product of the activation strength of the concept within that patch and the cosine similarity between the concept's dictionary vector (its corresponding weights in the SAE decoder) and the text embedding of the predicted label. This allows us to explicitly rank which concepts drove the segmentation prediction for any given patch.
\subsubsection{Pixel-wise Concept Contributions.}
To achieve finer, pixel-perfect segmentation boundaries while preserving interpretability, we integrate AnyUp~\cite{wimmer2025anyup}. Rather than upsampling the final segmentation logits or the reconstructed dense features, we apply the spatial upsampling directly to the sparse concept activations. This transforms the coarse grid of patch-level concept activations into a high-resolution map, providing a distinct concept activation vector for every single pixel in the image. Once we have these pixel-level concept activations, we pass them through the frozen SAE decoder to reconstruct the dense CLIP embedding at the pixel resolution. The segmentation prediction is then made by computing the cosine similarity between this pixel-level reconstructed embedding and the text vocabulary. Crucially, because the upsampling occurs purely within the concept space, the interpretability mechanism remains entirely intact at a much higher resolution. The pixel-wise concept contribution is computed exactly as it is in the patch-wise setting: by multiplying the upsampled pixel-level concept activation by the cosine similarity between the concept's dictionary vector and the predicted text label. This enables us to generate highly detailed explanation maps, where the decision for every individual pixel can be decomposed into its top contributing, human-understandable concepts.
%%%%%%%%%%%%%%%%%%%%%%%%%%%%%%%%%%%%%%%%%%%%%%%%%%%%%%%%%%%%%%%%

\subsection{Hyperparameter and Architectural Ablation}
\label{app:sec:hyperparameter_ablation}
\label{app:sec:ext_ablations}

\begin{table}[h]
\centering
\footnotesize %
\caption{\textbf{Dictionary metrics as in Zaigrajew \etal~\cite{zaigrajew2025interpreting} on ImageNet.} We find that concept sparsity helps produce highly interpretable concept-based explanations. Therefore, we select the sparsest possible configuration (BatchTopK of 12) that still maintains strong and stable SAE reconstruction.}

\begin{tabular}{@{}llccccccc@{}}
\toprule
\textbf{Dict. Size} &\textbf{BatchTopK} &\textbf{L0} $\uparrow$ & \textbf{FVU} $\downarrow$ & \textbf{CS} $\uparrow$ & \textbf{CKNNA} $\uparrow$ & \textbf{DO} $\downarrow$ & \textbf{NDN} $\downarrow$ \\
\midrule
8k& 12&0.998  & 0.148  & 0.923  & 0.675  & 0.0020 & 6 \\
4k& 12& 0.997 & 0.160  & 0.917 & 0.676  & 0.0034 & 4 \\
16k& 12& 0.999 & 0.143  & 0.926 & 0.679  & 0.0040 & 472 \\
8k& 6& 0.999 & 0.187  & 0.902 & 0.635  & 0.0042 & 1266 \\
8k& 32&  0.995 & 0.108  & 0.945 &  0.731 & 0.0029 & 2 \\
8k& 64& 0.991 & 0.078  & 0.961 & 0.713  & 0.0028 & 10 \\
8k& 128& 0.982 & 0.043 & 0.979 & 0.493  & 0.0028 & 159 \\
8k& 256& 0.966 & 0.012  & 0.994 & 0.095  & 0.0030 & 2453 \\
\bottomrule
\end{tabular}

\label{tab:dict_metrics}
\end{table}

% 
%%%%%%%%%%%%%%%%%%%%%%%%%%%%%%%%%%%%%%%%%%%%%%%%%%%%%%%%%%%%%%%%
% ==============================================================================
% APPENDIX SECTION & TABLE
% ==============================================================================
% \subsection{Extended Architectural Ablations}
% \label{app:sec:ext_ablations}

In this section, we present comprehensive hyperparameter and structural ablations of CFM. First, in \cref{tab:dict_metrics}, we ablate the dictionary size and BatchTopK threshold, measuring performance across various standard dictionary metrics to validate our default configuration. 

Next, we expand upon the structural ablations introduced in the main text. To ensure maximum readability, we decouple the evaluation of downstream task performance (Classification and Open-Vocabulary Segmentation) from the interpretability metrics (Locality, Consistency, Impurity, and $C^2$ Score). We divide these experiments into two major parts: assessing CFM's ability to scale to modern, large vision backbones (\cref{tab:supp_siglip_downstream} and \ref{tab:supp_siglip_interp}) and ablating the specific architectural choices of our framework built on standard CLIP (\cref{tab:supp_arch_downstream} and \ref{tab:supp_arch_interp}). All SAEs in this section are trained exclusively on ImageNet rather than CC12M.

\noindent \textbf{Backbone Scaling.} Scaling CFM to the SigLIP-2 backbone yields substantially better classification accuracy than our ViT-B/16 version ($78.8\% \rightarrow 86.2\%$), nearly matching the performance of the base SigLIP model while uniquely providing interpretability. For open-vocabulary segmentation, while spatial locality is slightly noisier compared to the ViT-B/16 variant, applying CFM still drives a massive improvement over MaskCLIP and matches the performance of CLIP-DINOiser on SigLIP-2. Crucially, CFM achieves this competitive segmentation performance while remaining fully interpretable, unlike those completely opaque baselines. Overall, this demonstrates that CFM generalizes well, preserving high downstream accuracy even when applied to larger backbones and different vision architectures.\\
\noindent \textbf{Guided Pooling.} Removing guided pooling reduces the model to patchSAE with task modulations. This causes a severe drop in segmentation performance and concept consistency, proving the mechanism is essential for spatially grounded tasks. \\
\noindent \textbf{Sparsity Factor ($k$).} Modifying $k$ introduces a trade-off: denser activations (e.g., $k{=}128$) improve spatial locality, while enforcing high sparsity (e.g., $k{=}6$) forces more semantically consistent concepts. Downstream performance remains largely unaffected. \\
\noindent \textbf{SAE Type.} Replacing Matryoshka with standard $\ell_1$ performs poorly on interpretability. While TopK and BatchTopK match downstream performance, they fundamentally lack the ability to extract semantic hierarchies.

% ------------------------------------------------------------------------------
% 1. SIGLIP SCALING: DOWNSTREAM PERFORMANCE
% ------------------------------------------------------------------------------
\begin{table*}[h]
\centering
\scriptsize
\setlength{\tabcolsep}{3pt}
\caption{\textbf{Backbone Scaling (SigLIP-2) - Downstream Performance.} We evaluate classification and Open-Vocabulary Segmentation (OVS) on the modern, 400M parameter SigLIP-2 SO/14 backbone. Best results are \textbf{bold}, second best are \underline{underlined}.}
\begin{tabular}{@{}l|cc|ccccccccc@{}}
\toprule
\multirow{2}{*}{\textbf{Method}} & 
\multicolumn{2}{c|}{\textbf{Classification} $\uparrow$}& 
\multicolumn{9}{c}{\textbf{Open-vocabulary segmentation} $\uparrow$} \\

\cmidrule(lr){2-3} \cmidrule(lr){4-12}

& \textbf{IMN} & \textbf{Places} & 
\textbf{V20} & \textbf{C59} & \textbf{Stf} & \textbf{Cty} & \textbf{ADE} & 
\textbf{Ctx} & \textbf{Obj} & \textbf{VOC} & \textbf{Avg} \\
\midrule
Base SigLIP-2  & \textbf{87.6} & \textbf{57.4} & - & - & - & - & - & - & - & - & - \\
MaskCLIP~\cite{zhou2022extract} & - & - & 72.3 & 22.4 & 16.5 & 21.1 & 16.4 & 19.8 & 16.4 & 28.2 & 26.6 \\
CLIP-DINOiser~\cite{wysoczanska2024clipdinoiser} & 86.4 & 57.3 & \textbf{86.1} & \textbf{29.8} & \textbf{20.8} & \textbf{33.0} & \textbf{21.9} & \textbf{26.3} & 27.2 & 33.9 & \textbf{34.9} \\
CFM (SigLIP-2) & \underline{86.2} & \underline{57.3} & \underline{85.3} & \underline{29.5} & \underline{20.7} & \underline{31.6} & \underline{21.2} & \underline{26.2} & \textbf{28.0} & \textbf{35.3} & \underline{34.7} \\
\bottomrule
\end{tabular}
\label{tab:supp_siglip_downstream}
\end{table*}

% ------------------------------------------------------------------------------
% 2. SIGLIP SCALING: INTERPRETABILITY
% ------------------------------------------------------------------------------
\begin{table*}[h]
\centering
\scriptsize
\setlength{\tabcolsep}{6pt}
\caption{\textbf{Backbone Scaling (SigLIP-2) - Interpretability Metrics.} CFM yields well-grounded, consistent concepts even when scaling to a much larger vision backbone.}
\begin{tabular}{@{}l|cc|cc|cc|c@{}}
\toprule
\multirow{2}{*}{\textbf{Method}} & 
\multicolumn{2}{c|}{\textbf{Locality} $\uparrow$} & 
\multicolumn{2}{c|}{\textbf{Consistency} $\uparrow$} & 
\multicolumn{2}{c|}{\textbf{Impurity} $\downarrow$} &
\multicolumn{1}{c}{\textbf{C$^2$} $\uparrow$}\\

\cmidrule(lr){2-3} \cmidrule(lr){4-5} \cmidrule(lr){6-7} \cmidrule(l){8-8}

& \textbf{Part} & \textbf{Stuff} & 
\textbf{Part} & \textbf{Stuff} & 
\textbf{Part} & \textbf{Stuff} & \textbf{Score} \\
\midrule
CFM (SigLIP-2) & 39.4 & 41.2 & 14.9 & 11.8 & 0.298 & 0.613 & 0.503 \\
\bottomrule
\end{tabular}
\label{tab:supp_siglip_interp}
\end{table*}

% ------------------------------------------------------------------------------
% 3. ARCHITECTURE ABLATIONS: DOWNSTREAM PERFORMANCE
% ------------------------------------------------------------------------------
\begin{table*}[h]
\centering
\scriptsize
\setlength{\tabcolsep}{3pt}
\caption{\textbf{Architecture Ablations (CLIP ViT-B/16) - Downstream Performance.} We independently remove guided pooling, alter the sparsity factor ($k$), and test alternative SAE types. Best results are \textbf{bold}, second best are \underline{underlined}. Note that the classification results in this table are with the OpenCLIP CLIP ViT-B/16  backbone, in contrast results in \cref{fig:classification} are done with the OpenAI CLIP ViT-B/16 backbone for clear comparability with existing CBMs.}
\begin{tabular}{@{}l|cc|ccccccccc@{}}
\toprule
\multirow{2}{*}{\textbf{Method}} & 
\multicolumn{2}{c|}{\textbf{Classification} $\uparrow$}& 
\multicolumn{9}{c}{\textbf{Open-vocabulary segmentation} $\uparrow$} \\

\cmidrule(lr){2-3} \cmidrule(lr){4-12} 

& \textbf{IMN} & \textbf{Places} & 
\textbf{V20} & \textbf{C59} & \textbf{Stf} & \textbf{Cty} & \textbf{ADE} & 
\textbf{Ctx} & \textbf{Obj} & \textbf{VOC} & \textbf{Avg} \\
\midrule

\rowcolor{Black!10!white} 
\multicolumn{12}{l}{\textbf{Default Setup}} \\
CFM (Default) & \underline{78.8} & 55.6 & \textbf{81.1} & \textbf{36.8} & 24.5 & 38.6 & \textbf{20.6} & \textbf{33.3} & 34.1 & \textbf{62.2} & \textbf{41.4} \\
\midrule

\rowcolor{Black!10!white} 
\multicolumn{12}{l}{\textbf{Guided Pooling}} \\
CFM w/o guid. pool. & 77.7 & 55.0 & 63.0 & 26.1 & 17.9 & 27.2 & 14.9 & 23.5 & 17.3 & 36.4 & 28.3 \\
\midrule

\rowcolor{Black!10!white} 
\multicolumn{12}{l}{\textbf{Sparsity Factor ($k$)}} \\
CFM w/ $k$=6  & 78.5 & 55.5 & 80.7 & \underline{36.5} & 23.9 & 38.0 & 20.0 & \underline{33.2} & 32.7 & 61.8 & 40.9 \\
CFM w/ $k$=32  & \textbf{78.9} & \underline{55.8} & 80.4 & 36.3 & 24.3 & 38.9 & 20.3 & 32.8 & 34.0 & \underline{62.1} & 41.1 \\
CFM w/ $k$=64  & \underline{78.8} & \textbf{55.9} & 80.2 & 36.1 & 24.5 & \underline{39.3} & 20.3 & 32.7 & \underline{34.5} & 61.8 & 41.2 \\
CFM w/ $k$=128 & 78.3 & 55.7 & 80.6 & 35.9 & \underline{24.6} & \underline{39.3} & \underline{20.5} & 32.5 & 34.4 & \textbf{62.2} & \underline{41.3} \\
\midrule

\rowcolor{Black!10!white} 
\multicolumn{12}{l}{\textbf{SAE Type (vs Matryoshka)}} \\
CFM w/ L1 SAE & 77.4 & 55.1 & \underline{80.8} & 36.0 & \textbf{24.7} & \textbf{39.4} & \underline{20.5} & 32.5 & \textbf{34.7} & 61.9 & \underline{41.3} \\
CFM w/ TopK SAE & 78.7 & \underline{55.8} & 80.5 & 36.3 & \underline{24.6} & 38.8 & 20.4 & 33.0 & 34.2 & \textbf{62.2} & \underline{41.3} \\
CFM w/ BatchTopK & \textbf{78.9} & 55.6 & 80.5 & 36.3 & 24.4 & 38.2 & \textbf{20.6} & 33.0 & 34.1 & 62.0 & 41.1 \\
\bottomrule
\end{tabular}
\label{tab:supp_arch_downstream}
\end{table*}

% ------------------------------------------------------------------------------
% 4. ARCHITECTURE ABLATIONS: INTERPRETABILITY
% ------------------------------------------------------------------------------
\begin{table*}[h]
\centering
\scriptsize
\setlength{\tabcolsep}{6pt}
\caption{\textbf{Architecture Ablations (CLIP ViT-B/16) - Interpretability Metrics.} Evaluation of spatial locality, consistency, and impurity across architectural configurations. Best results are \textbf{bold}, second best are \underline{underlined}. Note that the SAE training dataset in this ablation is ImageNet, hence we get here an increased C2 score because it is also evaluated on ImageNet.}
\begin{tabular}{@{}l|cc|cc|cc|c@{}}
\toprule
\multirow{2}{*}{\textbf{Method}} & 
\multicolumn{2}{c|}{\textbf{Locality} $\uparrow$} & 
\multicolumn{2}{c|}{\textbf{Consistency} $\uparrow$} & 
\multicolumn{2}{c|}{\textbf{Impurity} $\downarrow$} &
\multicolumn{1}{c}{\textbf{C$^2$} $\uparrow$}\\

\cmidrule(lr){2-3} \cmidrule(lr){4-5} \cmidrule(lr){6-7} \cmidrule(l){8-8}

& \textbf{Part} & \textbf{Stuff} & 
\textbf{Part} & \textbf{Stuff} & 
\textbf{Part} & \textbf{Stuff} & \textbf{Score}\\
\midrule

\rowcolor{Black!10!white} 
\multicolumn{8}{l}{\textbf{Default Setup}} \\
CFM (Default) & 43.4 & 44.3 & \underline{15.4} & \underline{13.2} & 0.338 & 0.536 & 0.561\\
\midrule

\rowcolor{Black!10!white} 
\multicolumn{8}{l}{\textbf{Guided Pooling}} \\
CFM w/o guid. pool. & 43.2 & 44.6 & 11.8 & 8.1 & \textbf{0.219} & \textbf{0.449} & 0.442\\
\midrule

\rowcolor{Black!10!white} 
\multicolumn{8}{l}{\textbf{Sparsity Factor ($k$)}} \\
CFM w/ $k$=6  & 39.9 & 41.4 & \textbf{16.0} & \textbf{14.5} & 0.355 & 0.552 & \textbf{0.570}\\
CFM w/ $k$=32  & 47.5 & 47.2 & 14.0 & 11.5  & 0.328 & \underline{0.524} & 0.557\\
CFM w/ $k$=64  & \underline{48.9} & \underline{48.4} & 13.0 & 11.4 & 0.329 & 0.531 & 0.556\\
CFM w/ $k$=128 & \textbf{50.2} & \textbf{48.9} & 13.1 & 11.4 & 0.332 & 0.557 & 0.550\\
\midrule

\rowcolor{Black!10!white} 
\multicolumn{8}{l}{\textbf{SAE Type (vs Matryoshka)}} \\
CFM w/ L1 SAE & 48.0 & 47.4 & 2.8 & 2.3 & 0.688 & 1.200 & 0.315\\
CFM w/ TopK SAE & 45.8 & 46.6 & 14.3 & 12.6 & 0.342 & 0.583 & 0.543\\
CFM w/ BatchTopK & 44.2 & 44.6 & 14.6 & 12.9 & \underline{0.324} & 0.526 & \underline{0.567}\\
\bottomrule
\end{tabular}
\label{tab:supp_arch_interp}
\end{table*}
%%%%%%%%%%%%%%%%%%%%%%%%%%%%%%%%%%%%%%%%%%%%%%%%%%%%%%%%%%%%%%%% 
\clearpage
\section{Additional Results}
\label{app:sec:results}
In this section we extend the results provided in the main paper. We begin with concept visualizations for different concept hierarchies in \cref{app:sec:families-viz}, for different matryoshka fractions in \cref{app:sec:rand-viz}, and for accuracy of naming in \cref{app:sec:namecompare}. We also provide further results on quantitative metrics in \cref{app:sec:polysemanticity_evaluation,app:sec:comparison_to_post_hoc_attribution}. We then further demonstrate the downstream interpretability and utility of \ours, with interpretable classification (\cref{app:sec:classification-viz}), interpretable open-vocabulary segmentation (\cref{app:sec:segmentation-viz}), and steerable captioning (\cref{app:sec:captioning-viz}).

\subsection{Concept Hierarchies}
\label{app:sec:families-viz}
\begin{figure*}[b!]
\centering
\includegraphics[width=\linewidth]{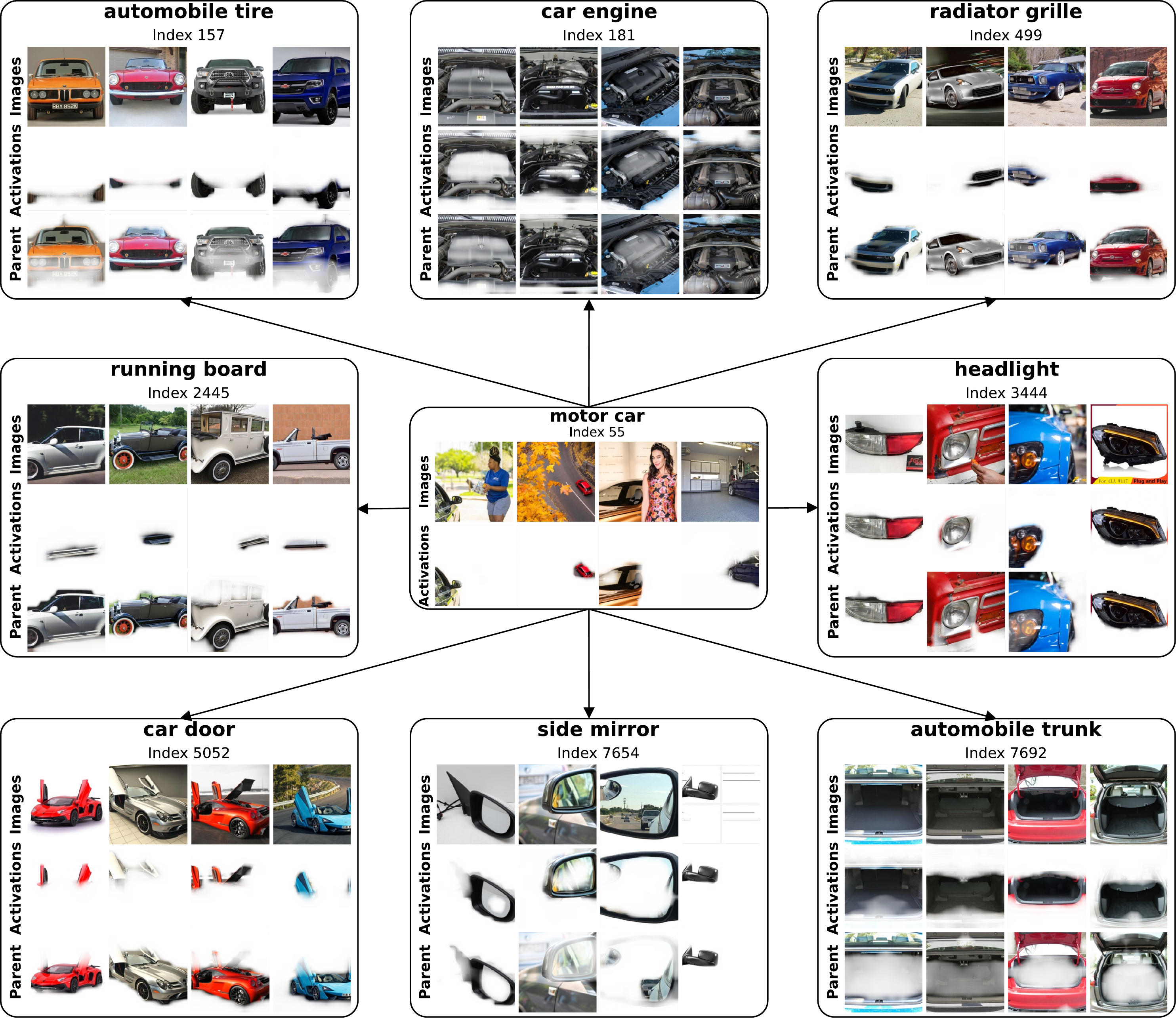}
\caption{\textbf{Hierarchies deconstruct complex objects into to their parts.} Here we show a subgraph of the ‘motor car’ hierarchy to illustrate that concept relationships can be formed as partonomies (part-based taxonomies). At the top of each concept visualization, we show the name and index of the concept, followed by a row of the top-activating images and their corresponding concept activation regions. For concepts with parents, we also show the parent activation on the same image below.}\label{app:fig:family1}
\end{figure*}

\begin{figure*}[h]
\centering
\includegraphics[width=\linewidth]{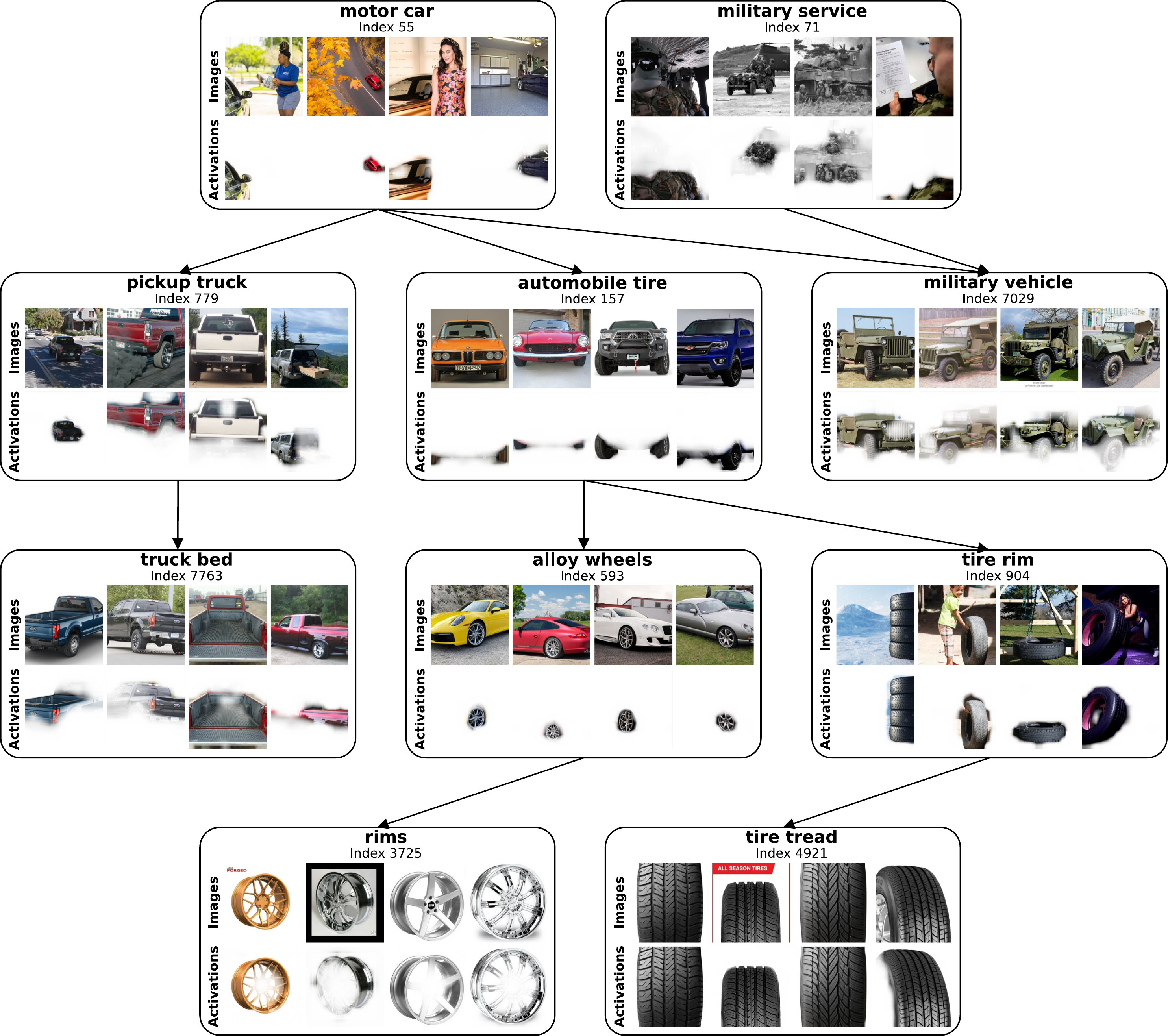}
\caption{\textbf{Concept hierarchies have a DAG structure.} We show a subgraph of the "motor car" hierarchy that shows the complexity of discovered hierarchies with several levels in the hierarchy and concepts having multiple parents (see 'military vehicle' with two parents 'motor car and 'military service'). Furthermore, the hierarchy also reveals progressively finer-grained concepts, such as 'alloy wheels' (third row) and the even finer subpart 'rim' (fourth row).} \label{app:fig:family2}
\end{figure*}

\begin{figure*}[t]
\centering
\includegraphics[width=\linewidth]{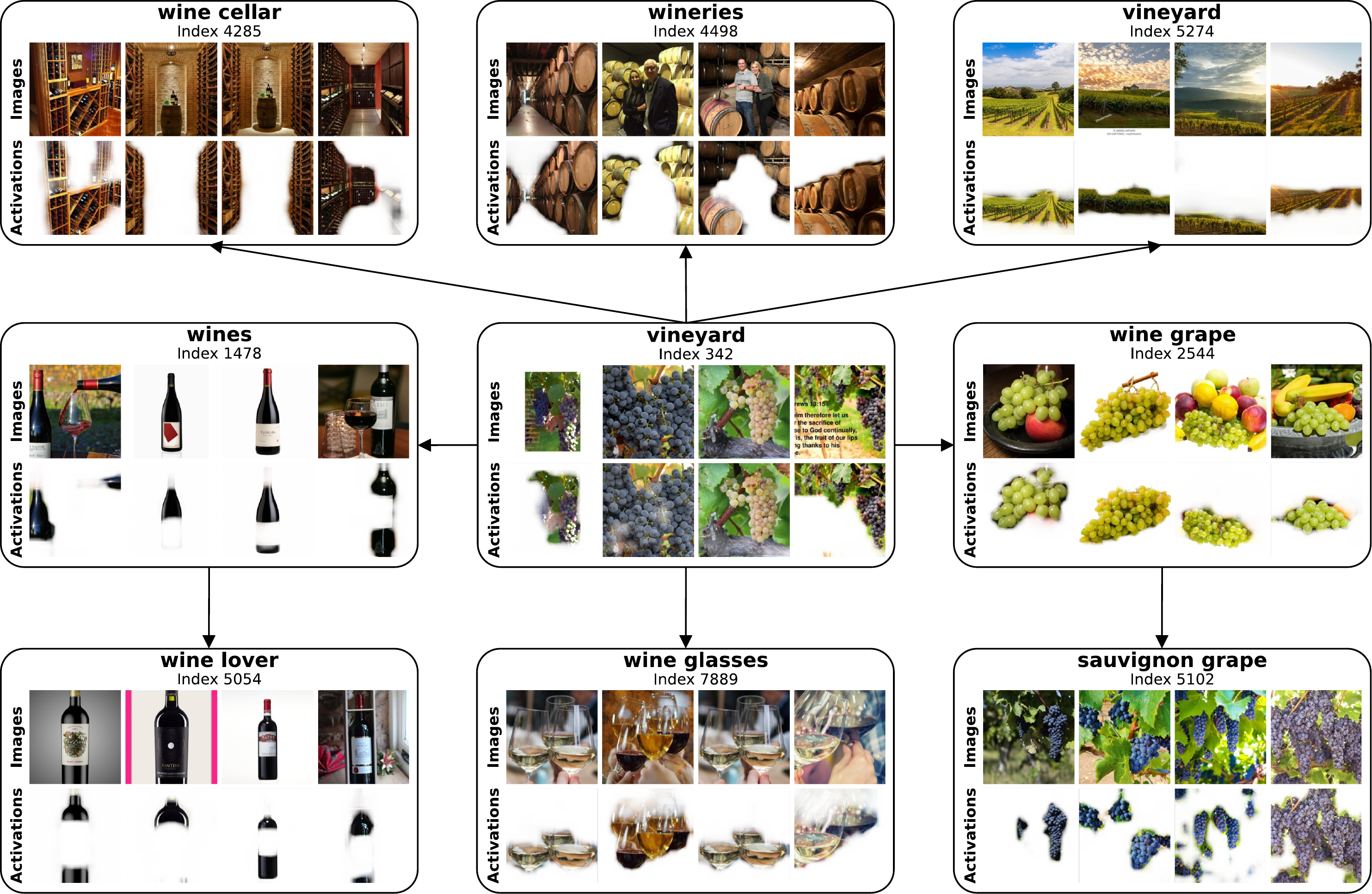}
\caption{\textbf{Hierarchies can be semantically highly coherent.} We show the entire "vineyard" hierarchy, i.e. the concept "vineyard" with all of its descendants and ancestors in the hierarchy graph $\mathcal{G}$. A hierarchy of acquired taste.} \label{app:fig:family3}
\end{figure*}

\begin{figure*}[h!]
\centering
\includegraphics[width=\linewidth]{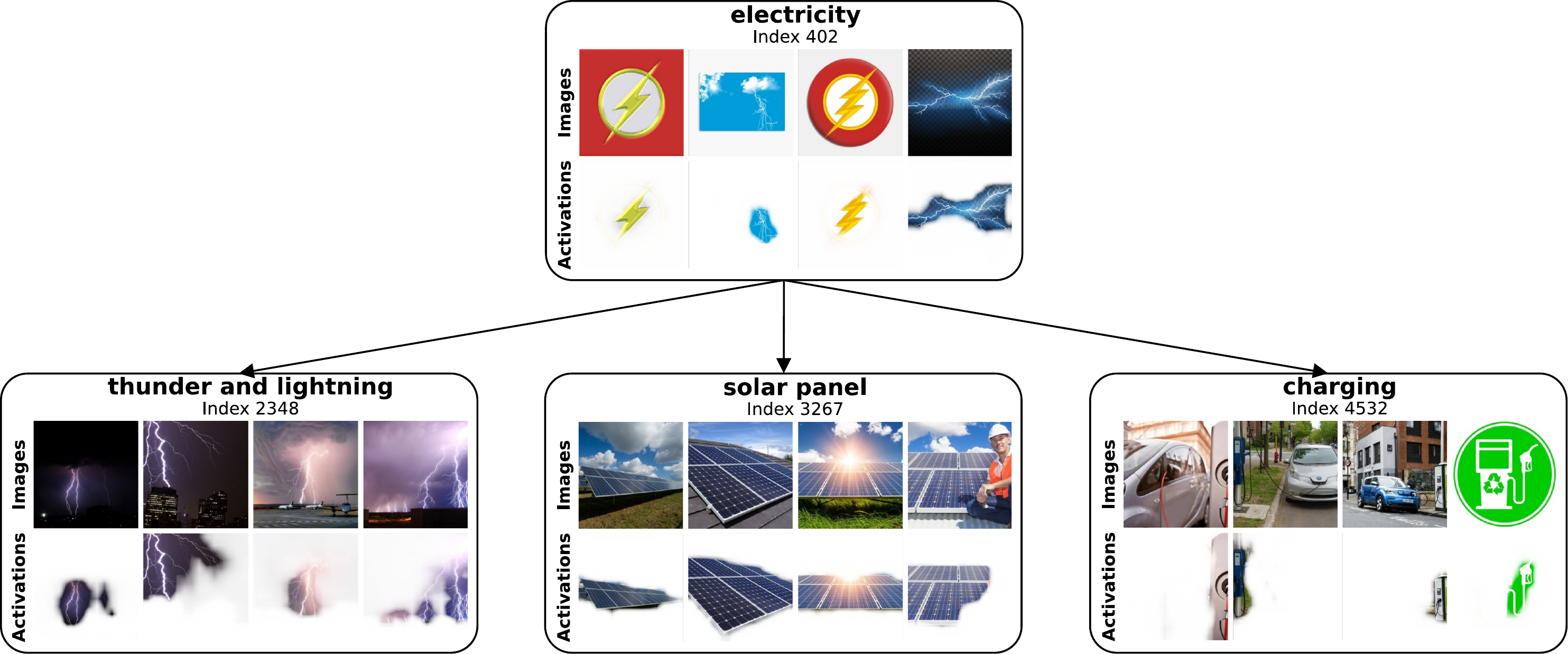}
\caption{\textbf{Hierarchies can have abstract semantic relationships, without any common visual patterns.} We show the full "electricity" hierarchy. Note that children like "solar panel" or "charging" do not share visual patterns with the top-activating examples of the parent "electricity", but instead are connected through an abstract semantic relationship.} \label{app:fig:family6}
\end{figure*}

\begin{figure*}[t]
\centering
\includegraphics[width=\linewidth]{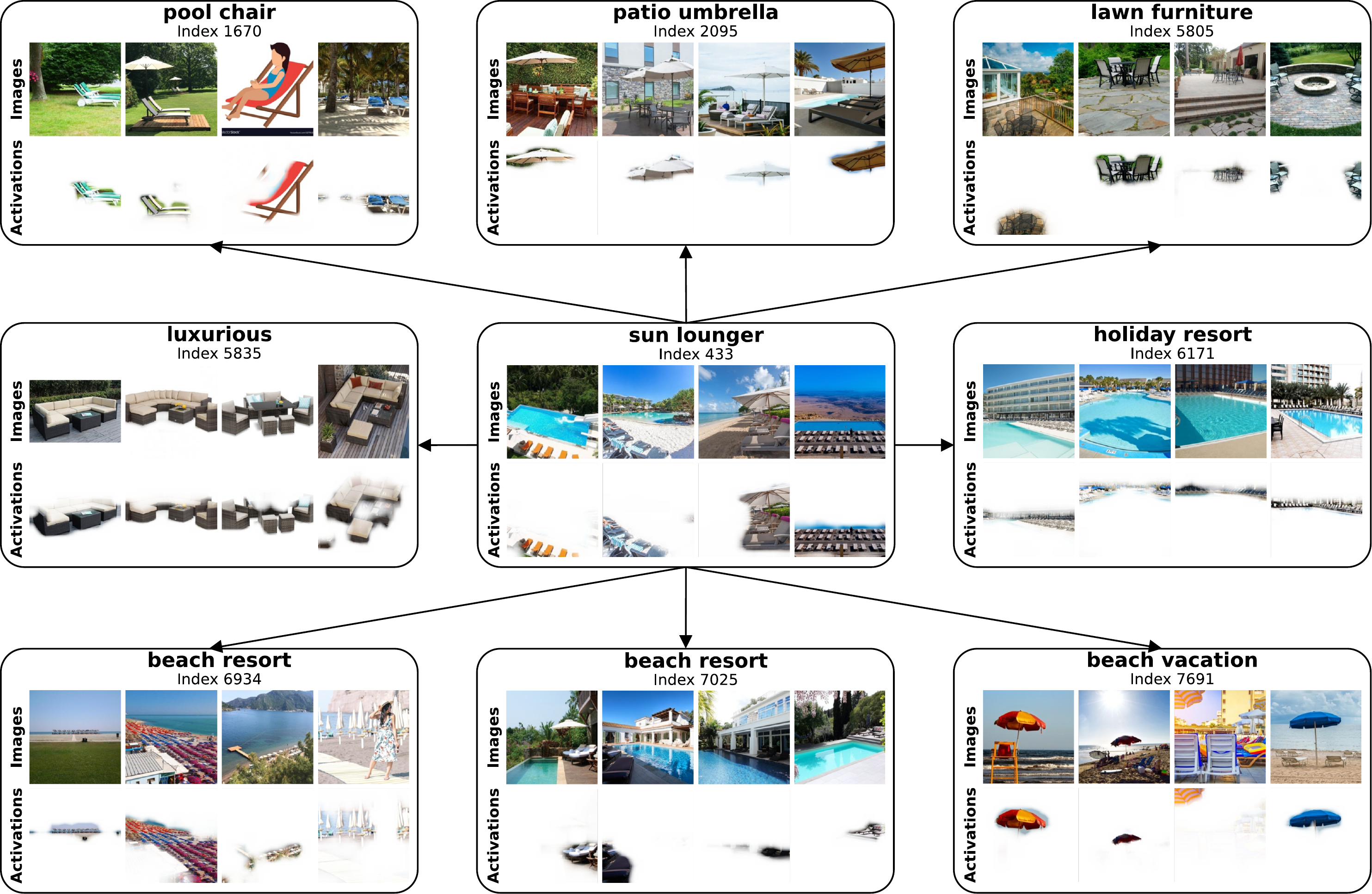}
\caption{\textbf{Concept naming can be limited in the context of hierarchies.} We show the entire "sun lounger" hierarchy with all of its descendants in the hierarchy graph $\mathcal{G}$. We observe that while the concept relations do make sense and concepts are spatially well grounded, concept naming is not ideal with multiple descendants receiving overly general names.} \label{app:fig:family4}
\end{figure*}

\begin{figure*}[h!]
\centering
\includegraphics[width=\linewidth]{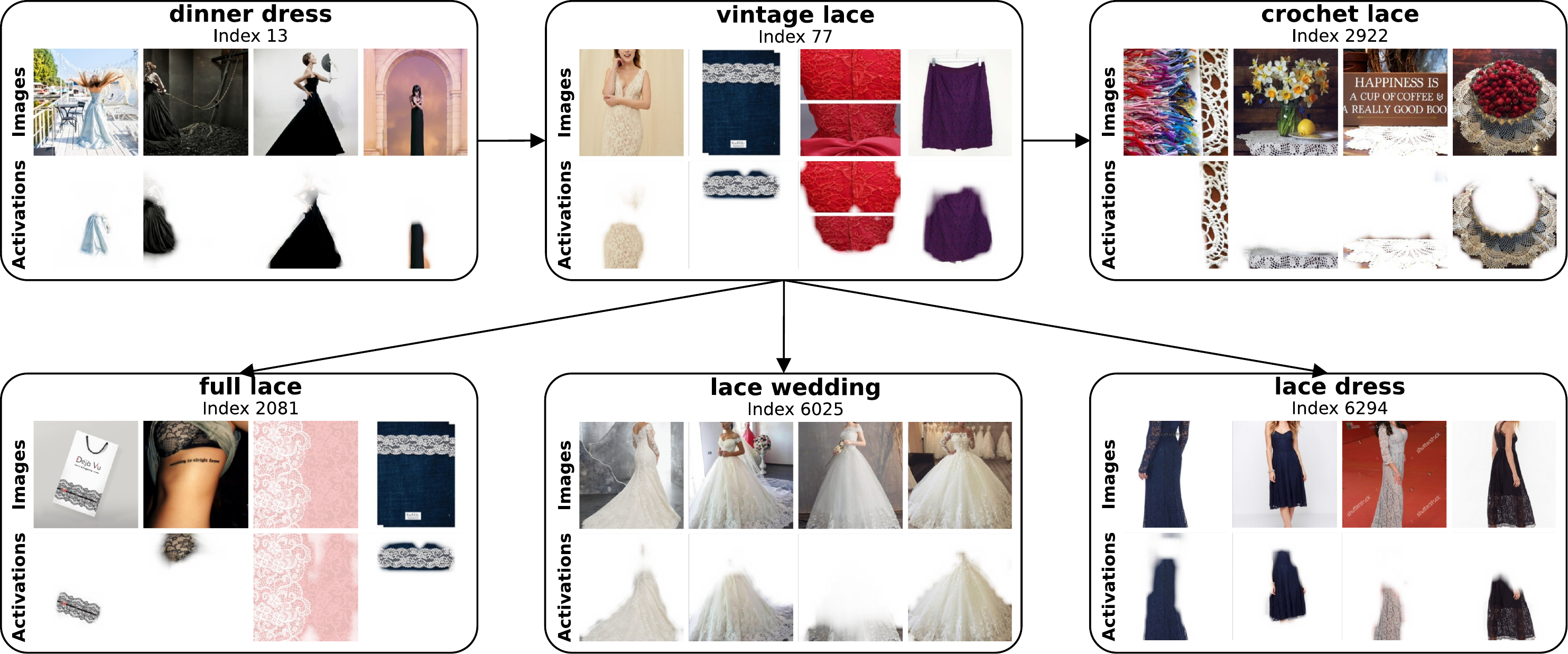}
\caption{\textbf{Hierarchies can be based on material properties.} We show the entire "vintage lace" hierarchy from the hierarchy graph $\mathcal{G}$. We find both specifications as well as sub-\textit{material} relationships ("is made of").} \label{app:fig:family5}
\end{figure*}

In \cref{app:fig:family1,app:fig:family2,app:fig:family3,app:fig:family4,app:fig:family5,app:fig:family6} we provide more curated examples from our concept relation graph that show how hierarchies can deconstruct complex objects into their parts, that these hierarchies can have several levels of hierarchy and a concept can share multiple parents, that relationship can be based on concepts describing properties of another concept (e.g., material), and that relations can be abstract semantics (e.g., electricity, solar panel, and charging). With \cref{app:fig:family4} we also show an example where relations are visually coherent, but naming can be limited, potentially because the concept vocabulary does not contain the relevant name or the text and image encoder of CLIP are not aligned for the specific fine-grained concept.

\clearpage

\clearpage


\clearpage


\clearpage
\subsection{Concept Consistency Visualization}
\label{app:sec:rand-viz}
\begin{figure*}[h!]
\centering
\includegraphics[width=\linewidth]{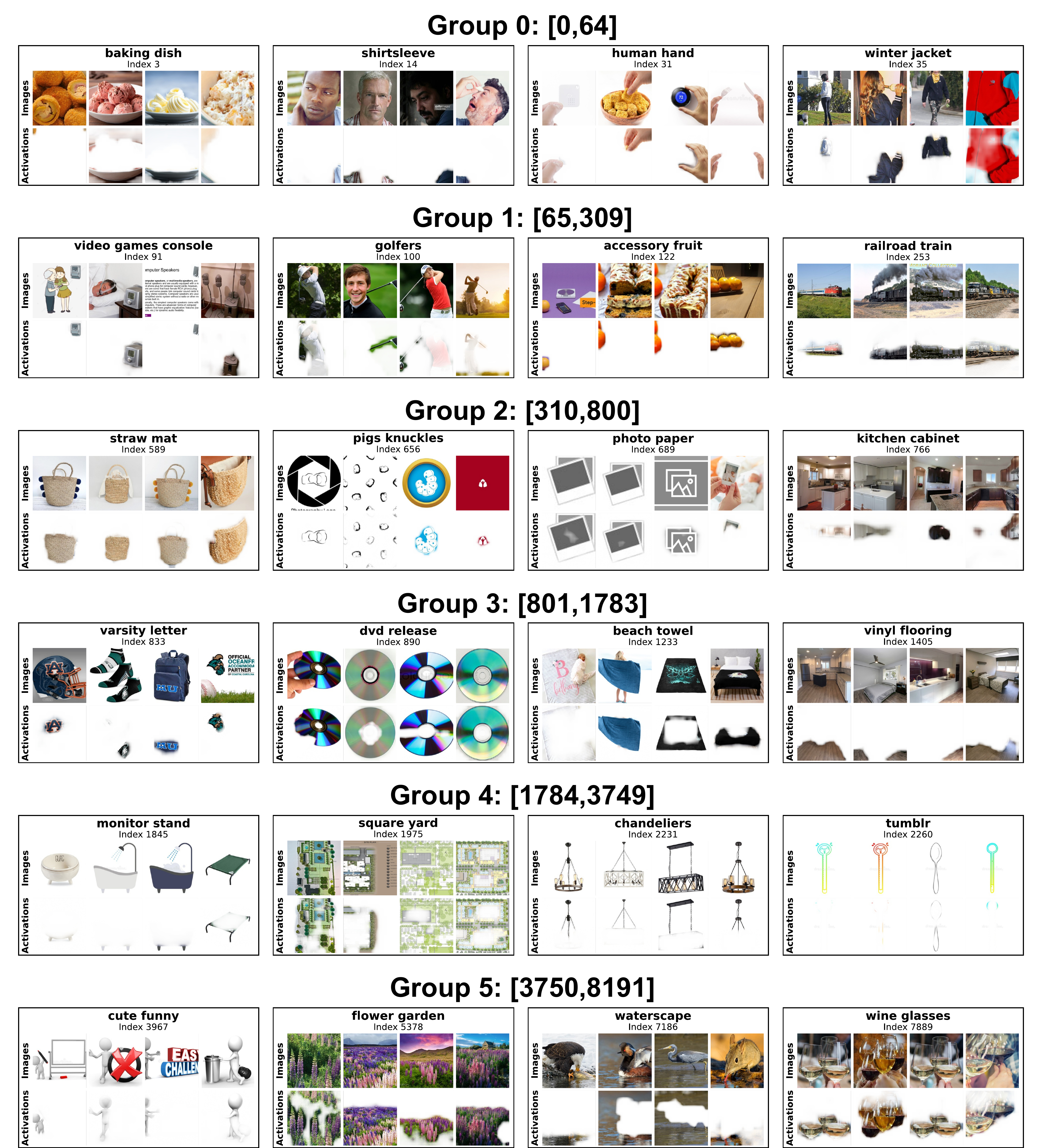}
\caption{\textbf{Randomly sampled Concepts per Matryoshka group.} We show a random selection of concepts in each Matryoshka shell, defined by a range of neuron indices. By design of the Matryoshka training, the lower the range, the more "general" the information. Here, one can also observe several automated concept naming failure patterns, as discussed in the main text. These include \textit{near-misses} (closely related but inaccurate names like ``accessory fruit'' or ``baking dish''); \textit{spurious correlations}; 
\textit{lexical gaps}, where geometric or textural features lacking direct language correspondence are assigned object names with similar shapes (e.g., ``pigs knuckles''); and misnamings due to an \textit{incomplete candidate vocabulary} pool.
}\label{app:fig:concepts_matryoshka}
\end{figure*}

We further show randomly sampled concepts for each Matryoshka group in \cref{app:fig:concepts_matryoshka} and observe that concepts are easy to understand, spatially well located, and correspond to frequent and less frequent concepts.

% \clearpage
\subsection{Concept Name Comparison}
\label{app:sec:namecompare}

\begin{figure}[h]
    \centering
    \includegraphics[width=0.8\linewidth]{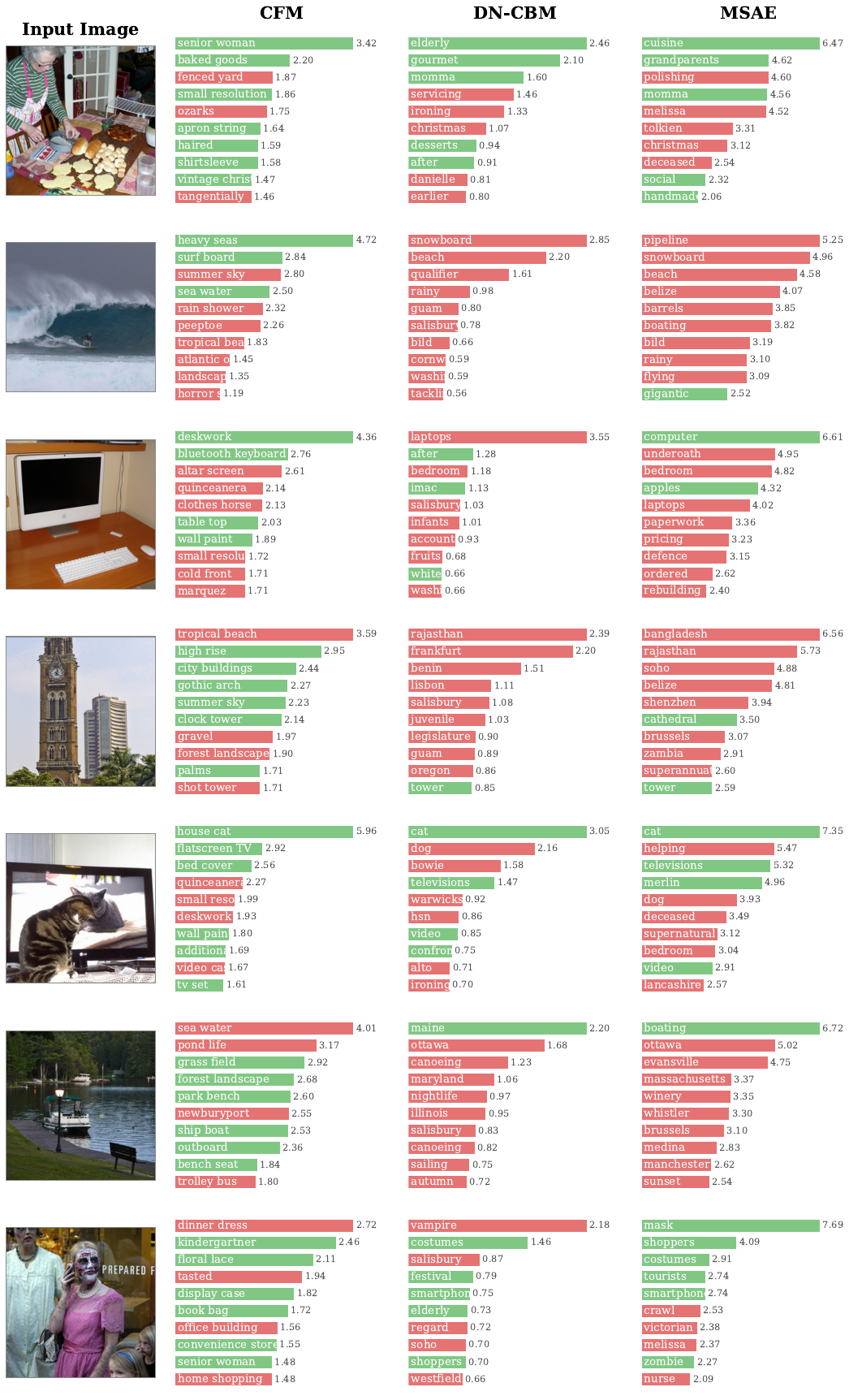}
    \caption{\textbf{Randomly sampled examples with extracted names.} We show additional randomly sampled examples for the evaluation shown in \cref{fig:name_accuracy}. We find that on the whole, \ours provides more accurate names.}
    \label{fig:supp:naming_examples}
\end{figure}

We provide additional examples of concepts extracted by \ours as compared to baselines in \cref{fig:supp:naming_examples}. This agrees with our observations in \cref{sec:conceptquality} and \cref{fig:name_accuracy}, where we show that \ours provides more accurate names as compared to baselines. For full experimental details, see \cref{app:sec:experimental_details:naming_judge}.

\subsection{Polysemanticity Evaluation}
\label{app:sec:polysemanticity_evaluation}
To quantitatively evaluate the reduction in semantic entanglement, we utilize the Concept Polysemanticity Entropy (CPE) score introduced by Yu \etal~\cite{yu2025coe}. The CPE metric operates on a logarithmic scale to measure concept polysemanticity, where lower scores indicate purer, more monosemantic representations. For this analysis, we compare the standard neurons from the patch tokens of the last transformer layer against our extracted CFM concepts. In the standard CPE evaluation protocol for convolutional architectures like ResNets, images are masked using the spatial neuron activations before being passed to the Vision-Language Model (VLM) to extract concept atoms. However, standard Vision Transformer (ViT) patch tokens generally exhibit poor spatial grounding, making this masked evaluation unreliable. Consequently, for ViTs, the standard baseline computes the CPE using entirely unmasked (non-local) images. For our extracted CFM concepts, we compute the CPE in both settings: with spatial masking (local) and without masking (non-local). Our results demonstrate that CFM successfully resolves polysemanticity. The standard ViT neurons yield a non-local CPE of 0.926. In contrast, our CFM concepts achieve a non-local CPE of 0.869, indicating that our formulation inherently reduces semantic ambiguity. Furthermore, when leveraging the strong spatial grounding of CFM to apply precise concept masks, the CPE drops even further to 0.814. This confirms that \ours provides disentangled concepts with accurate spatial localization.
\clearpage
\subsection{Concept Consistency Evaluation}
\label{app:sec:comparison_to_post_hoc_attribution}

\begin{figure}
    \centering
    \includegraphics[width=0.8\linewidth]{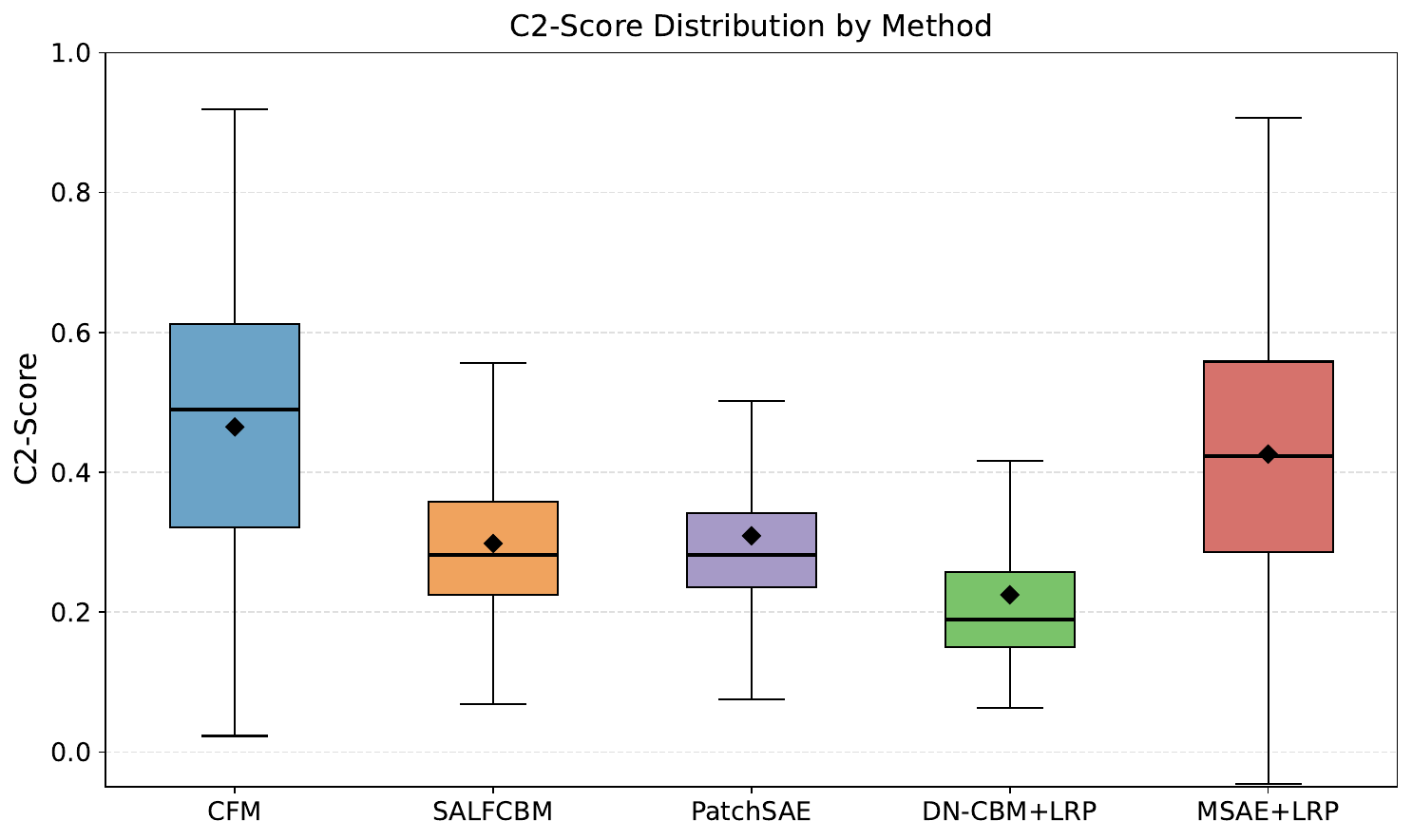}
    \caption{We evaluate \ours against other baselines in terms of C$^2$-Score~\cite{parchami2025fact} and observe that it outperforms methods with inherent spatial concept activations (SALFCBM and PatchSAE). It also outperforms DN-CBM and performs similar to MSAE, all while not requiring a post-hoc attribution method (LRP) with additional backward passes.}
    \label{app:fig:c2}
\end{figure}

We provide additional results for evaluation of concept consistency, localization, and impurity, as discussed in \cref{sec:conceptquality} and \cref{fig:interpret_local} in the main paper. Specifically, we compare across two additional baselines---MSAE~\cite{zaigrajew2025interpreting} and DN-CBM~\cite{rao2024discover}, that do not directly provide image level grounding. Therefore, we use layer-wise relevance propagation (LRP)~\cite{chefer2021generic} to obtain pixel-level attributions, and evaluate with our localization, consistency, and impurity metrics, as shown in \cref{app:tab:locmetrics}. We find that \ours outperforms both across metrics. We also compare across methods on the C$^2$-Score in \cref{app:fig:c2}, and find that \ours significantly outperforms most baselines. This is despite the fact that unlike LRP, \ours does not require an additional backward pass.

\begin{table}[h]
\centering
\scriptsize
\vspace{-10pt}
\caption{\textbf{\ours discovers well-grounded and consistent concepts.} We extend \cref{fig:interpret_local}-left from the main paper and compare the inherent spatial grounding of \ours against global concept models (\dncbm and \msae), for which we apply a post-hoc attribution method~\cite{chefer2021generic}. We observe that \ours outperforms both across locality, consistency, and impurity metrics.}
\label{app:tab:locmetrics}
\begin{tabular}{lcccccc}
\toprule
& \multicolumn{2}{c}{\textbf{Loc.} $\uparrow$} & \multicolumn{2}{c}{\textbf{Cons.} $\uparrow$} &\multicolumn{2}{c}{ \textbf{Impur.} $\downarrow$} \\
\textbf{Methods}  & Part & Stuff & Part & Stuff & Part & Stuff \\ \midrule
\dncbm~\cite{rao2024discover} + LRP~\cite{chefer2021generic}& 18.7 & 17.9 & 9.3 & 5.7 & 0.761 & 1.235              \\
\msae~\cite{zaigrajew2025interpreting} + LRP~\cite{chefer2021generic}& 20.9 & 20.0 & 10.6 & 7.9 & 0.833 & 1.198   \\
\rowcolor{ForestGreen!15!white} \ours & \textbf{44.3} & \textbf{44.5} & \textbf{15.3} & \textbf{13.1} & \textbf{0.333} & \textbf{0.539} \\
\bottomrule
\end{tabular}

\end{table}

\clearpage
\subsection{Interpretable Classification}
\label{app:sec:classification-viz}
\begin{figure}[t]
    \centering
    \begin{subfigure}{.49\linewidth}
        \centering%
        \includegraphics[width=\linewidth ]{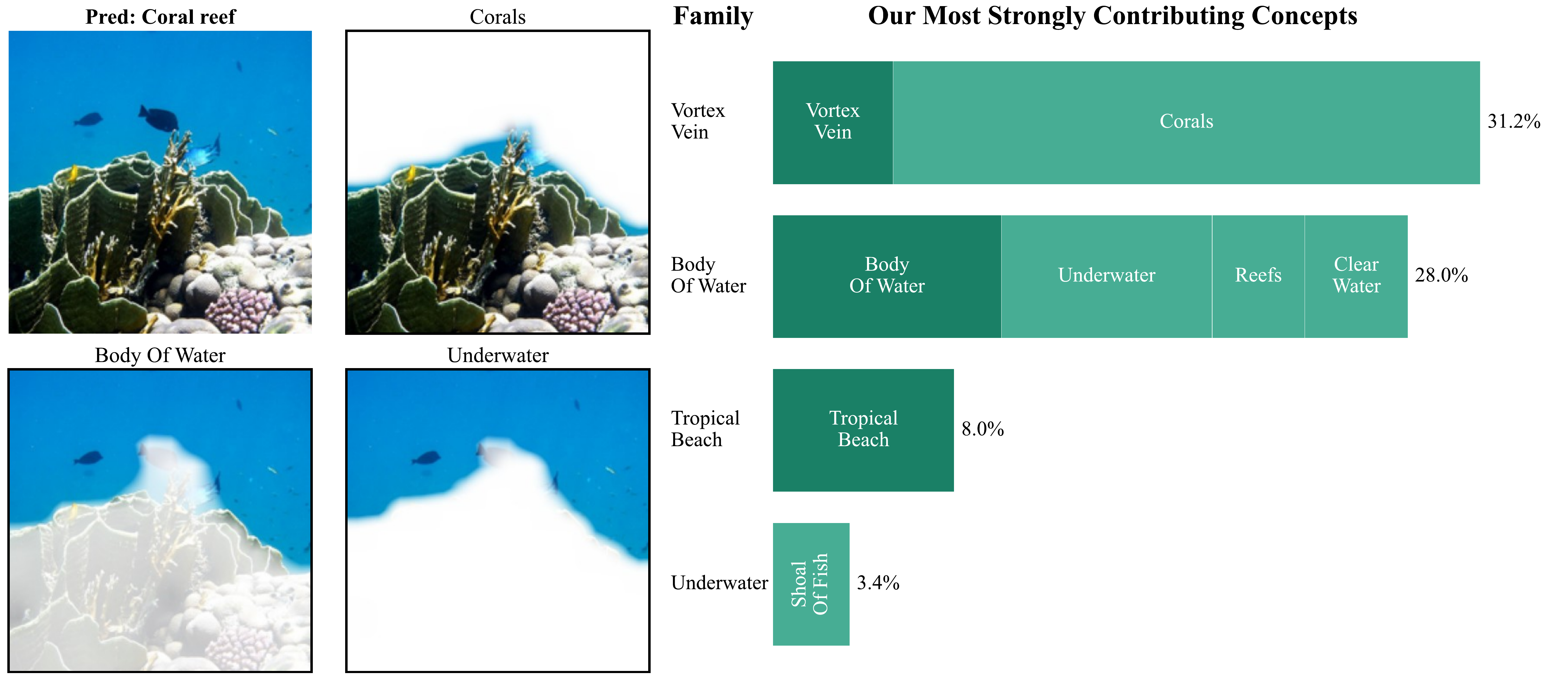}%
    \end{subfigure}%
    \begin{subfigure}{.49\linewidth}
        \centering%
        \includegraphics[width=\linewidth ]{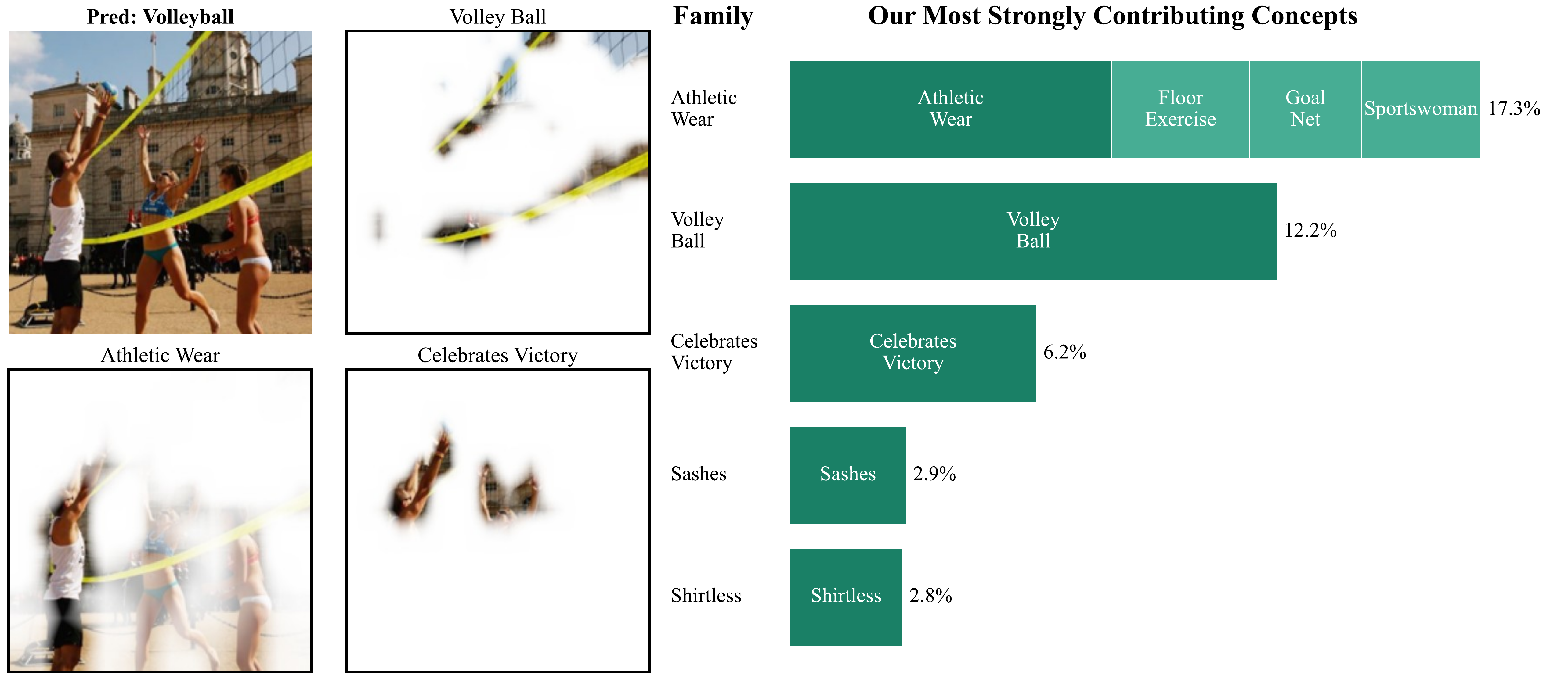}%
    \end{subfigure}%
    \vspace{1.0em}
    \begin{subfigure}{.49\linewidth}
        \centering%
        \includegraphics[width=\linewidth ]{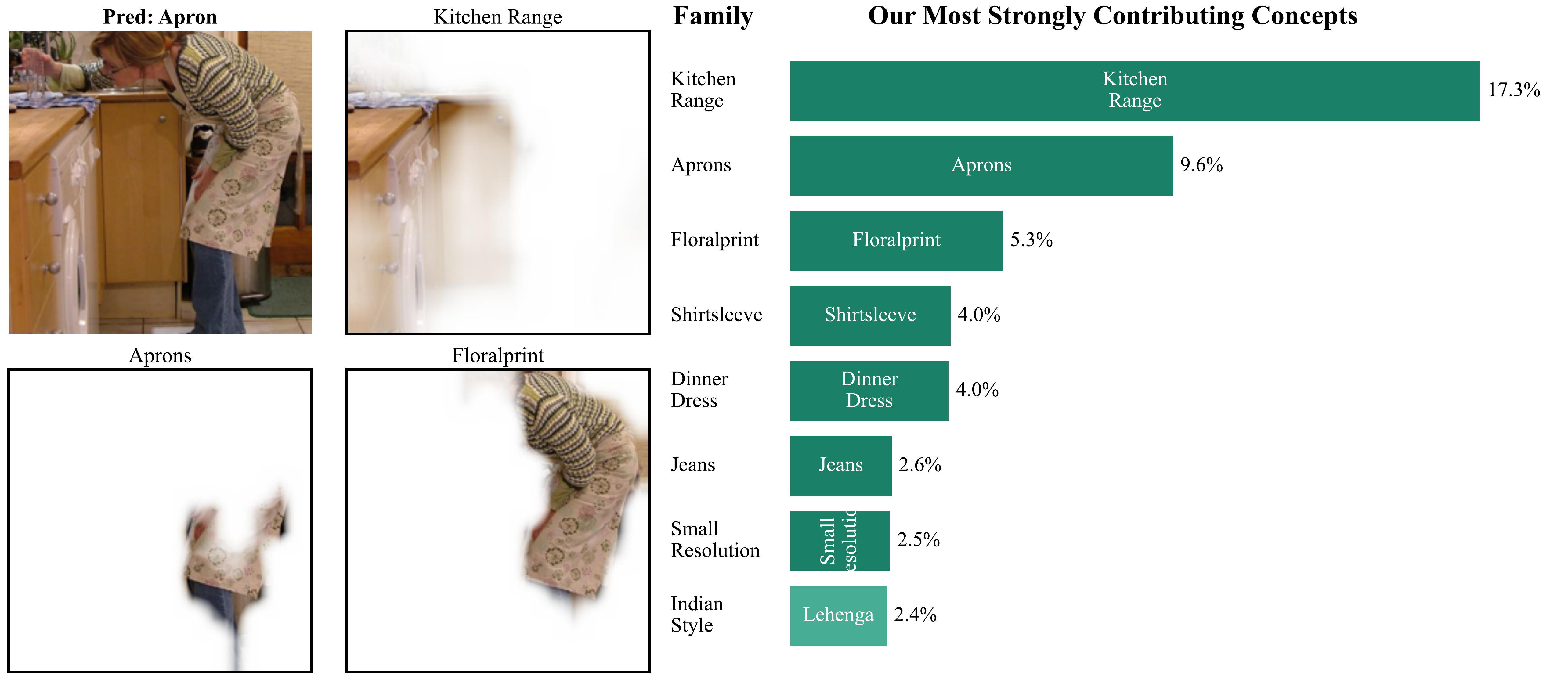}%
    \end{subfigure}%
    % \vspace{0.2em}
    \begin{subfigure}{.49\linewidth}
        \centering%
        \includegraphics[width=\linewidth ]{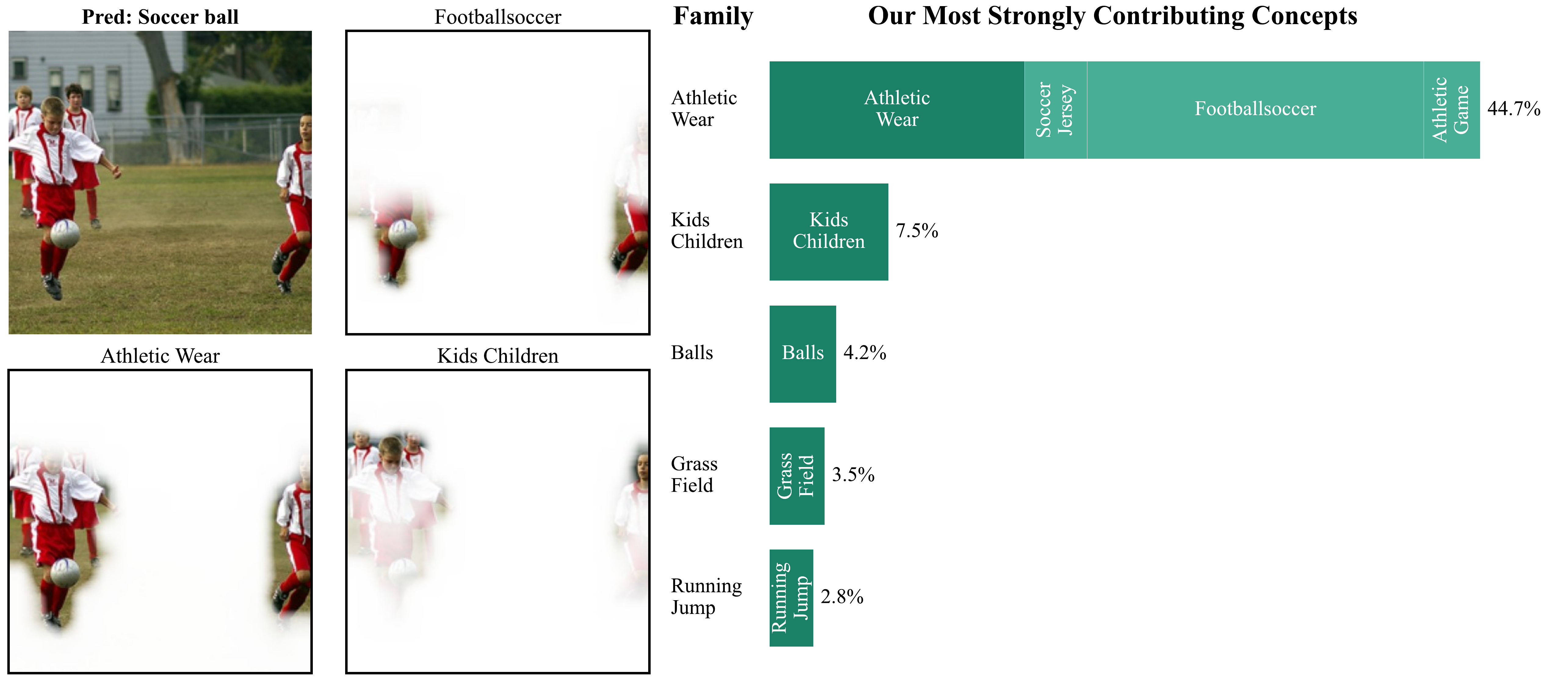}%
    \end{subfigure}%
    
    \caption{\textbf{Spatial grounding of concept-based explanations for ImageNet.}  We provide example images, their predicted class, along with fine-grained concept-based explanations showing contributions to prediction as bar plots, and spatial grounding in terms of per-patch concept activations.}\label{app:fig:class_exp_imagenet}
\end{figure}

\begin{figure}
    \centering
    % Top subfigure
    \begin{subfigure}{.49\linewidth}
        \centering%
        \includegraphics[width=\linewidth ]{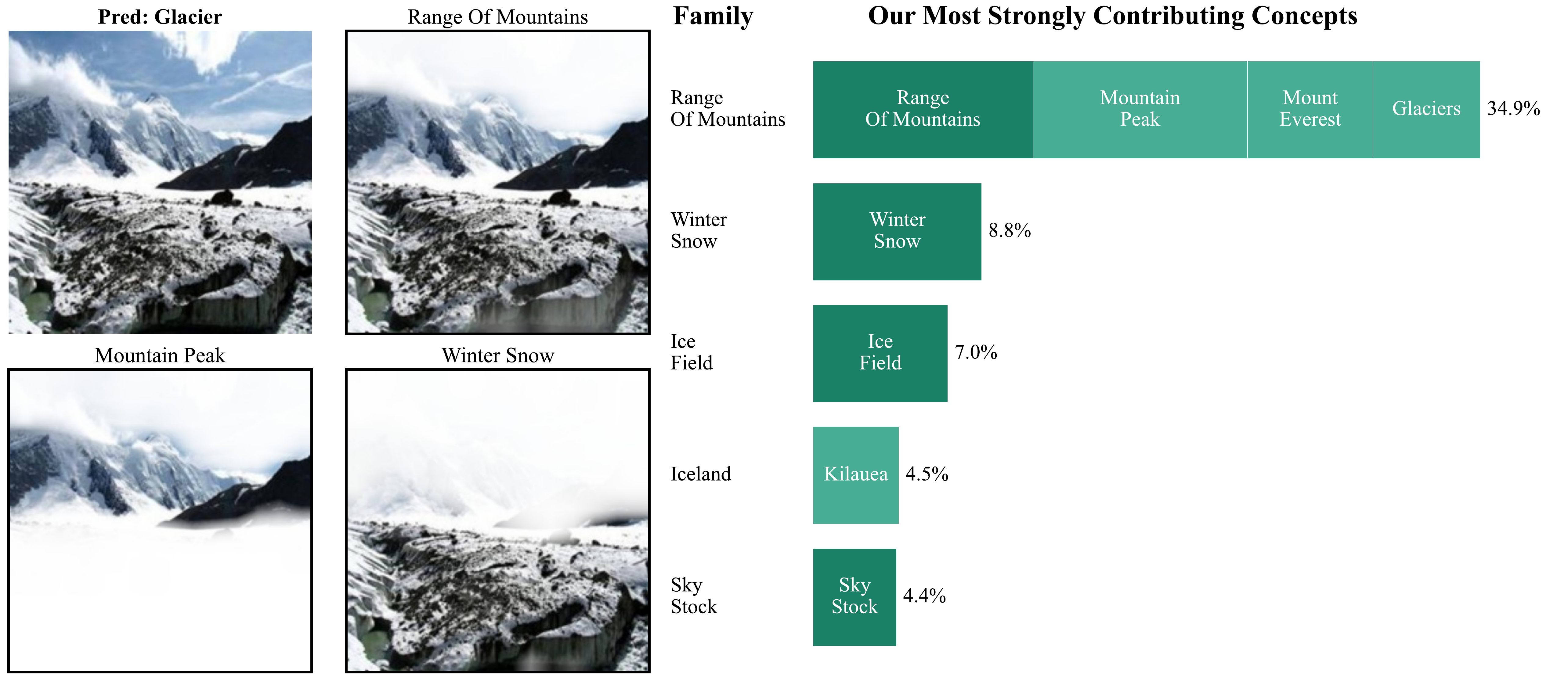}%
    \end{subfigure}%
    \begin{subfigure}{.49\linewidth}
        \centering%
        \includegraphics[width=\linewidth ]{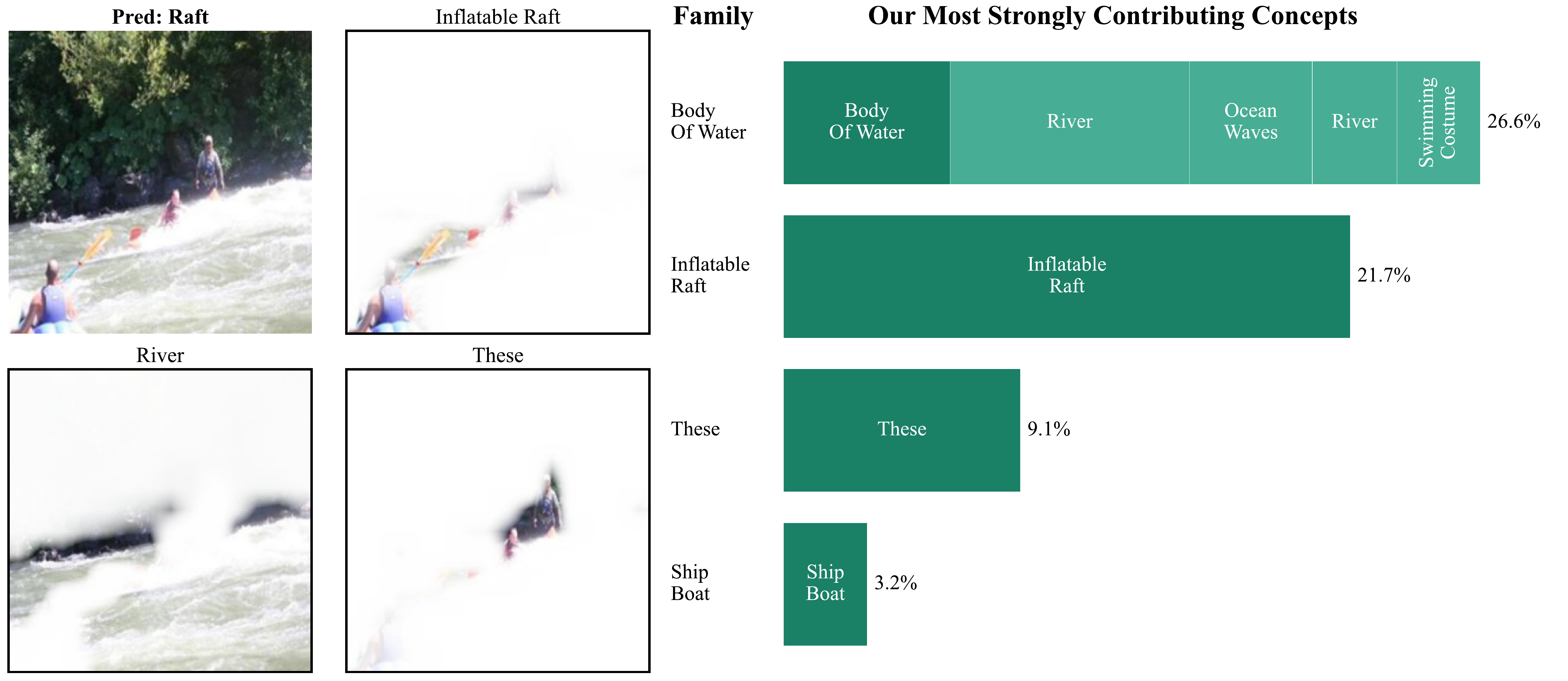}%
    \end{subfigure}%
    \vspace{1.0em}
    \begin{subfigure}{.49\linewidth}
        \centering%
        \includegraphics[width=\linewidth ]{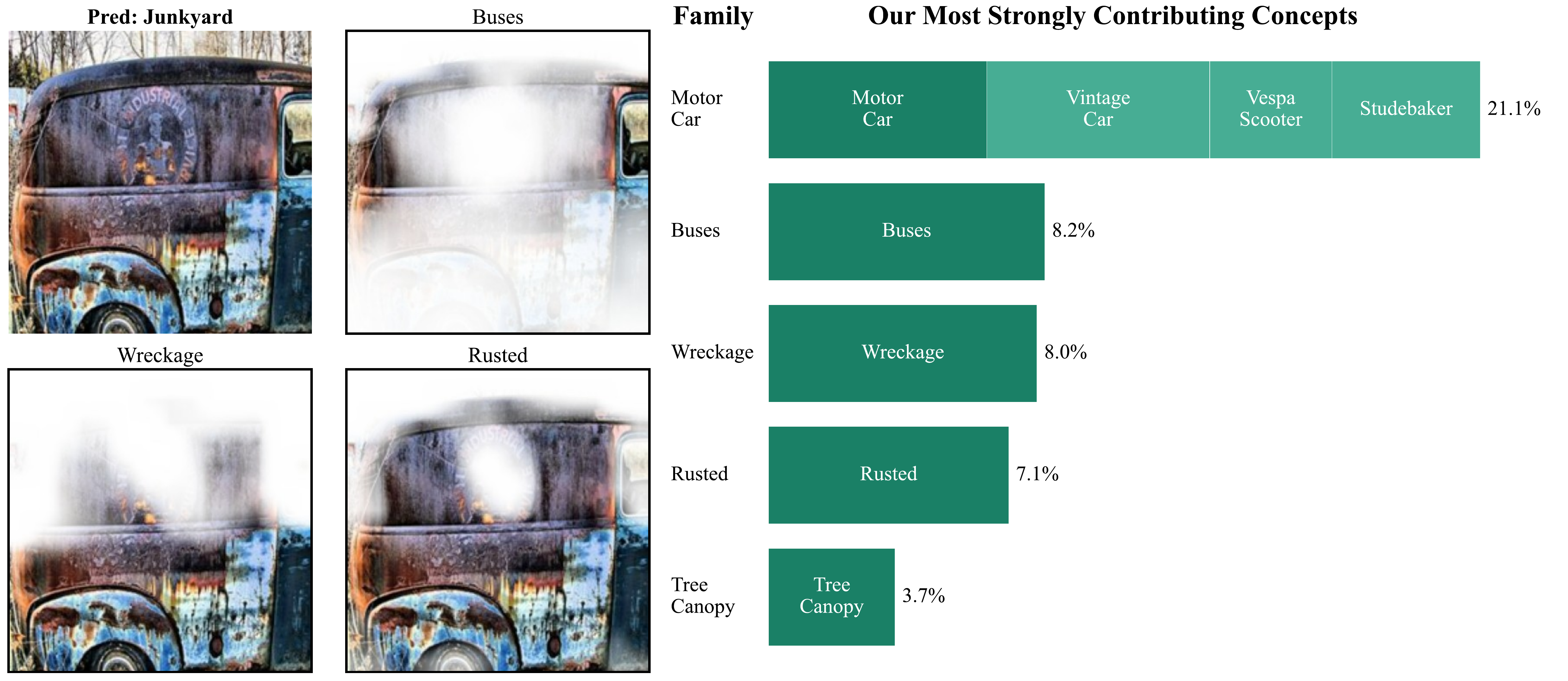}%
    \end{subfigure}%
    % \vspace{0.2em}
    \begin{subfigure}{.49\linewidth}
        \centering%
        \includegraphics[width=\linewidth ]{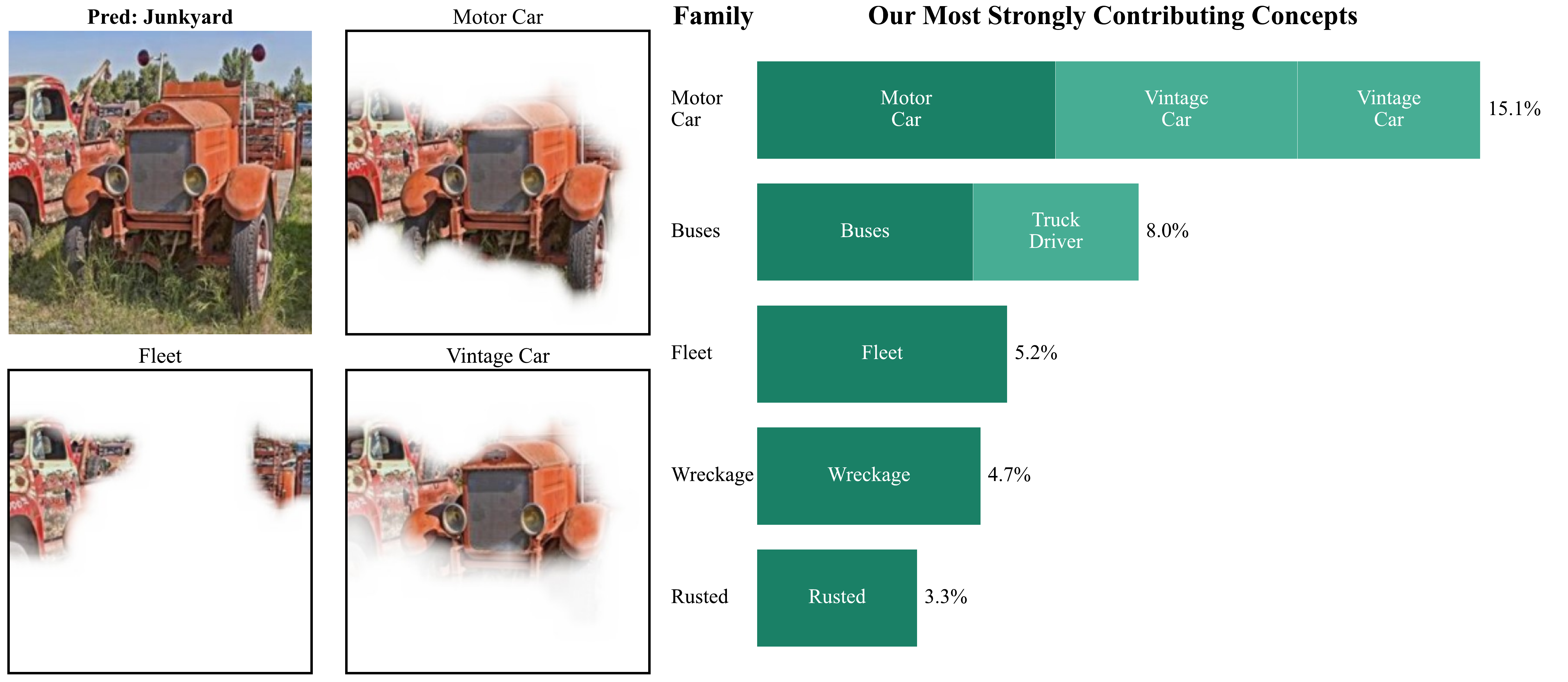}%
    \end{subfigure}%
    
    \caption{\textbf{Spatial grounding of concept-based explanations for Places365.} We provide example images, their predicted class, along with fine-grained concept-based explanations showing contributions to prediction as bar plots, and spatial grounding in terms of per-patch concept activations. See also Figure 8 of Rao \etal~\cite{rao2024discover} for a comparison to existing methods.}\label{app:fig:class_exp_places}
\end{figure}

Here, we provide additional qualitative results explanations for classifications on ImageNet (\cref{app:fig:class_exp_imagenet}) and Places365 (\cref{app:fig:class_exp_places}) and observe that explanations are rooted in spatially well-grounded fine-grained relevant concepts. Moreover, in contrast to many existing works, the top concepts cover a large fraction of the overall relevance for a classification, meaning the explanations are succinct, which is a natural consequence of using (batch) top-k SAE architectures.

\subsection{Interpretable Open-Vocabulary Segmentation} 
\label{app:sec:segmentation-viz}
\begin{figure}[htbp]
    \centering
    \includegraphics[width=\textwidth]{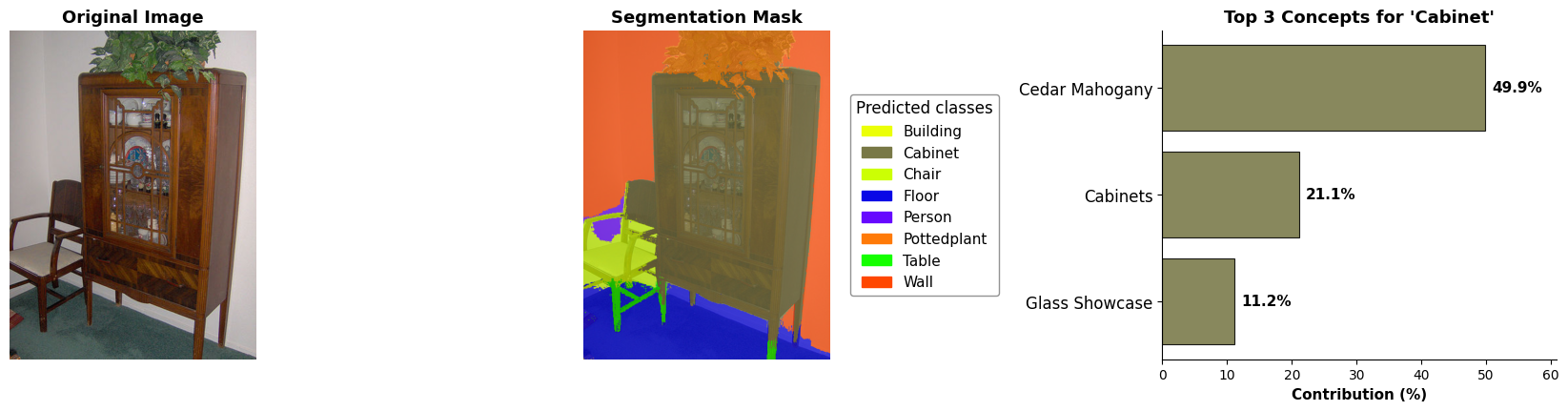}
    \includegraphics[width=\textwidth]{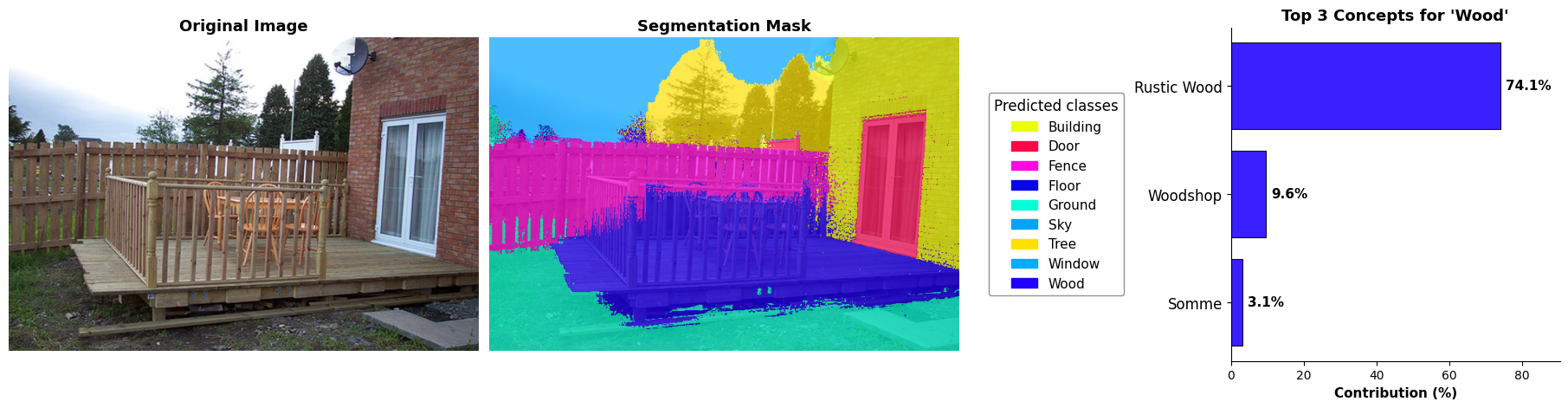}
    \includegraphics[width=\textwidth]{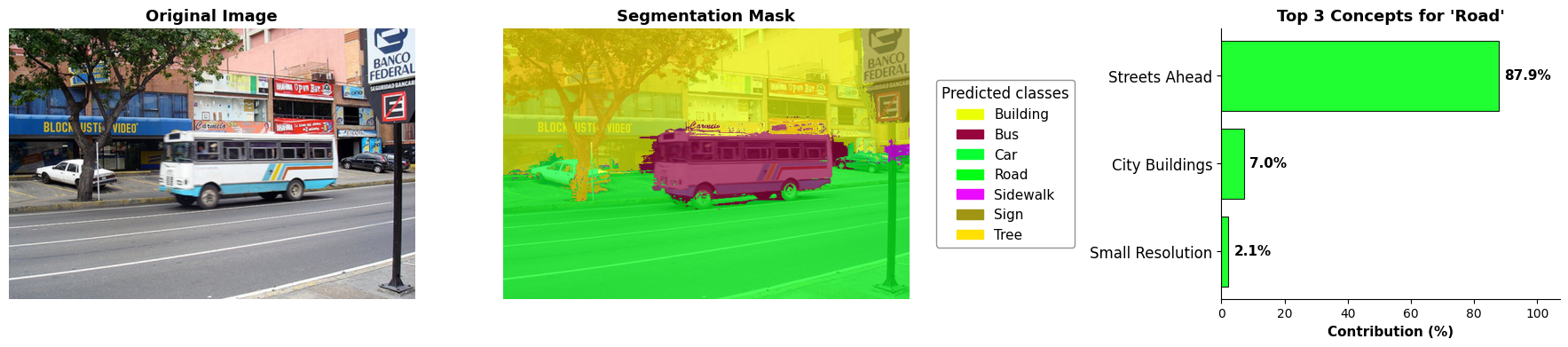}
    \includegraphics[width=\textwidth]{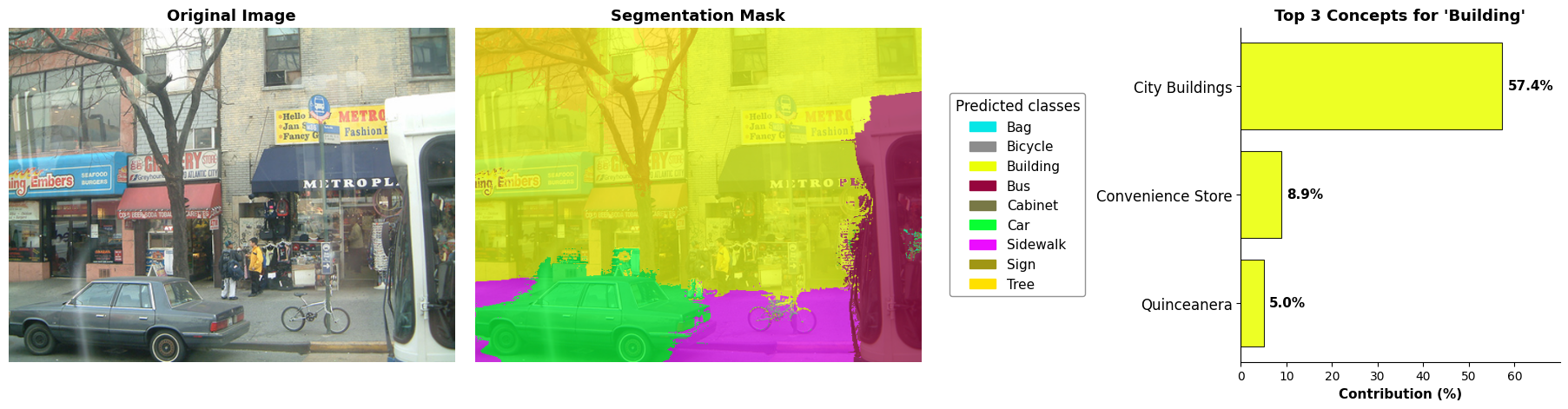}
    \includegraphics[width=\textwidth]{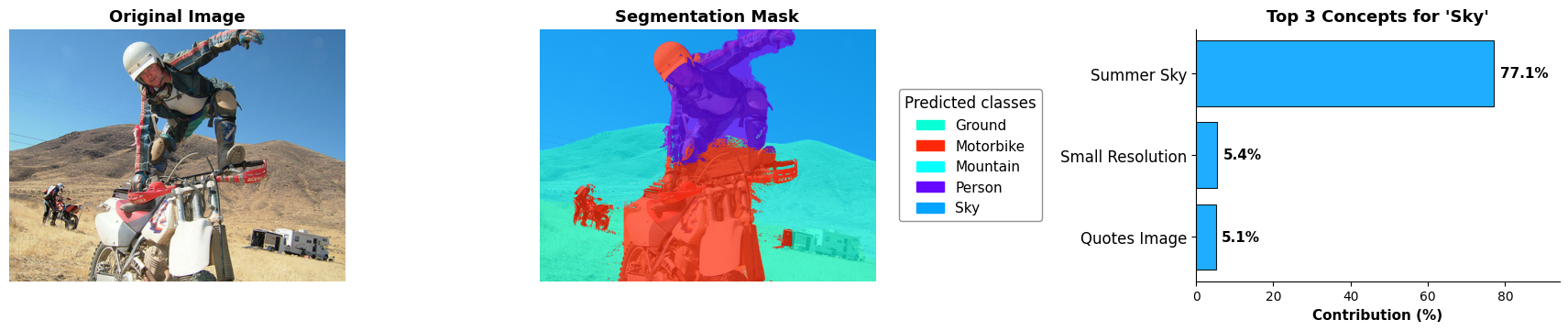}
    
    \caption{\textbf{$\ours^\text{+(AnyUp)}$ provides high-quality, interpretable open-vocabulary segmentation at the pixel level.} We show five randomly sampled scenes from the Pascal Context-59 dataset. For each example (row): Left: the original input image. Middle: the pixel-level segmentation mask produced by CFM. Right: a bar plot detailing the top contributing concepts for the largest predicted segment in the scene. The concept contributions are computed directly at the pixel level as the product of the spatially upsampled concept activation and its cosine similarity with the target label embedding. This illustrates how \ours yields highly transparent, fine-grained explanations for its dense predictions across diverse real-world scenes.}
    \label{app:fig:context59_qualitative}
\end{figure}
\begin{figure*}[h]
\centering
\includegraphics[width=\linewidth]{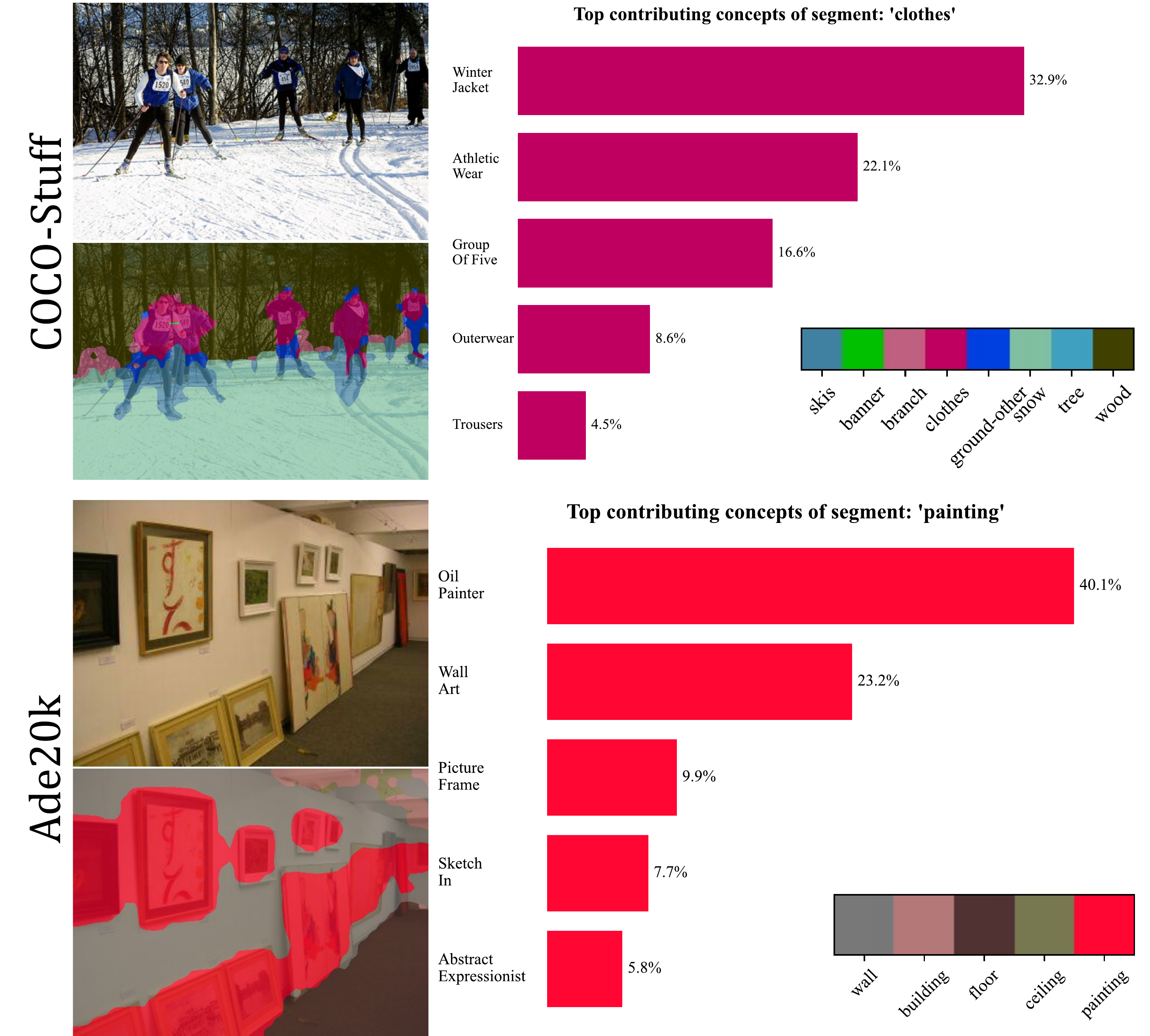}
\caption{\textbf{\ours enables interpretable open-vocabulary segmentation.} Top left: input image. Bottom left: segmentation produced by \ours when provided with the full set of dataset label names as text queries. Top right: bar plot showing the top contributing concepts for a selected segmentation label, where contribution is computed as the product of concept activation and its cosine similarity with the label embedding. Bottom right: all segmentation labels that the model identifies as present in the image. This illustrates how \ours produces both pixel-level predictions and transparent concept-level explanations of why each label is selected.}\label{app:fig:seg_exp}
\end{figure*}
\begin{figure*}[h]
\centering
\includegraphics[width=\linewidth]{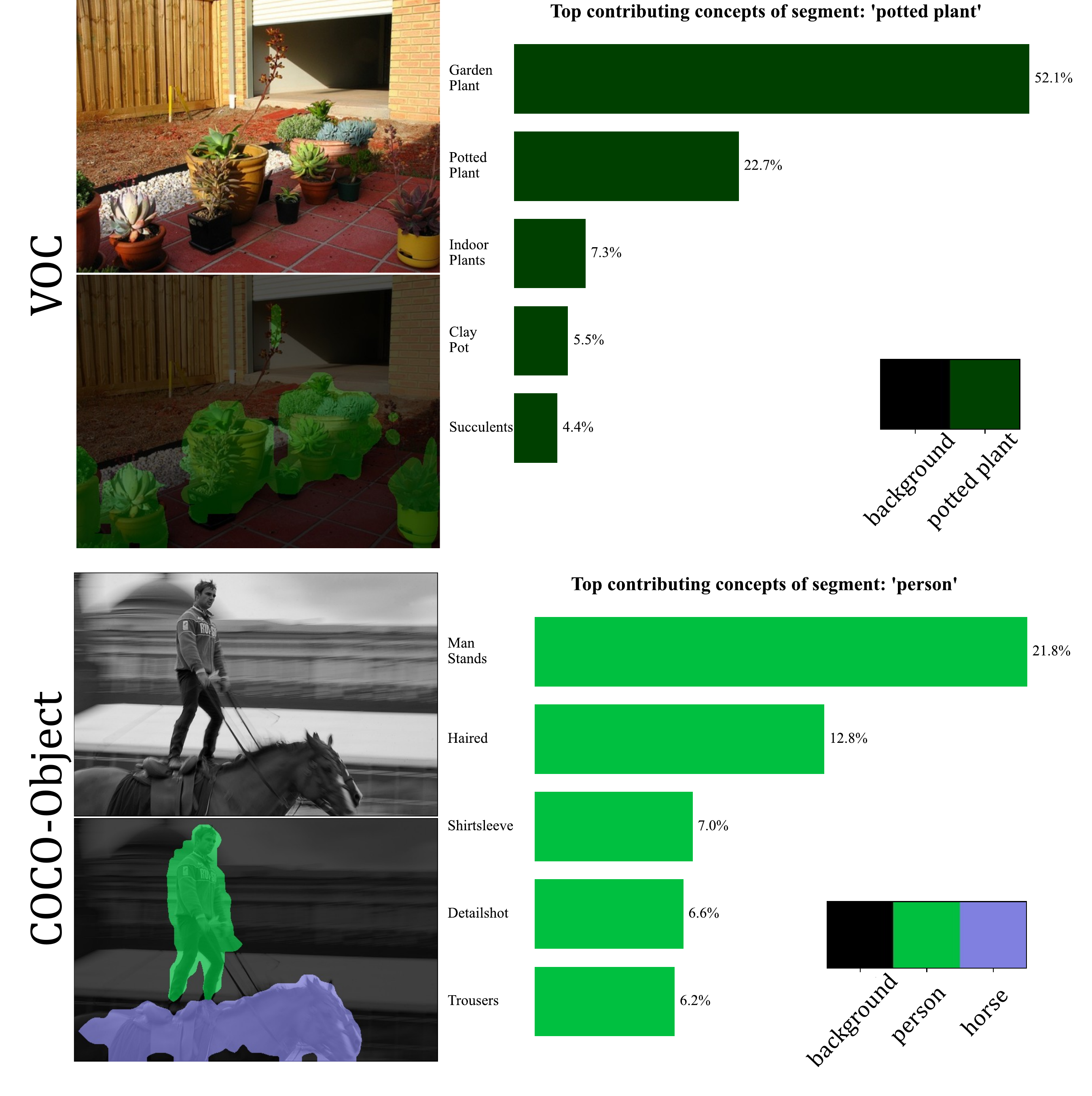}
\caption{\textbf{\ours enables interpretable open-vocabulary segmentation with background removal.}
We demonstrate segmentation results when providing \ours with the full set of dataset label names together with an additional background prompt. The background label is used to filter out non-foreground regions, allowing the model to focus on semantically meaningful segments. As in the main results, \ours produces both pixel-level segmentations and concept-level explanations through contribution scores derived from concept activations and their cosine similarity with the corresponding label embeddings.}\label{app:fig:seg_exp_bkg}
\end{figure*}

In \cref{app:fig:context59_qualitative}, we present randomly sampled qualitative explanations for the Pascal Context-59 dataset, specifically showcasing our method's pixel-level Open-Vocabulary Segmentation (OVS) prediction and explanation capabilities. 
Furthermore, \cref{app:fig:seg_exp,app:fig:seg_exp_bkg} provide curated results demonstrating our patch-level OVS performance with interpretable explanations across PASCAL-VOC \cite{everingham2011pascal}, COCO-Object \cite{lin2014microsoft}, COCO-Stuff \cite{caesar2018coco}, and ADE20K \cite{zhou2019semantic}. Crucially, unlike existing approaches, our method allows us to explicitly decompose these segmentation logits into individual concept-wise contributions for any given text label, providing highly transparent decision rationales.
\clearpage

\clearpage

\subsection{Steerable Captioning} 
\label{app:sec:captioning-viz}
\begin{figure*}[h!]
\centering
\includegraphics[width=0.9\linewidth]{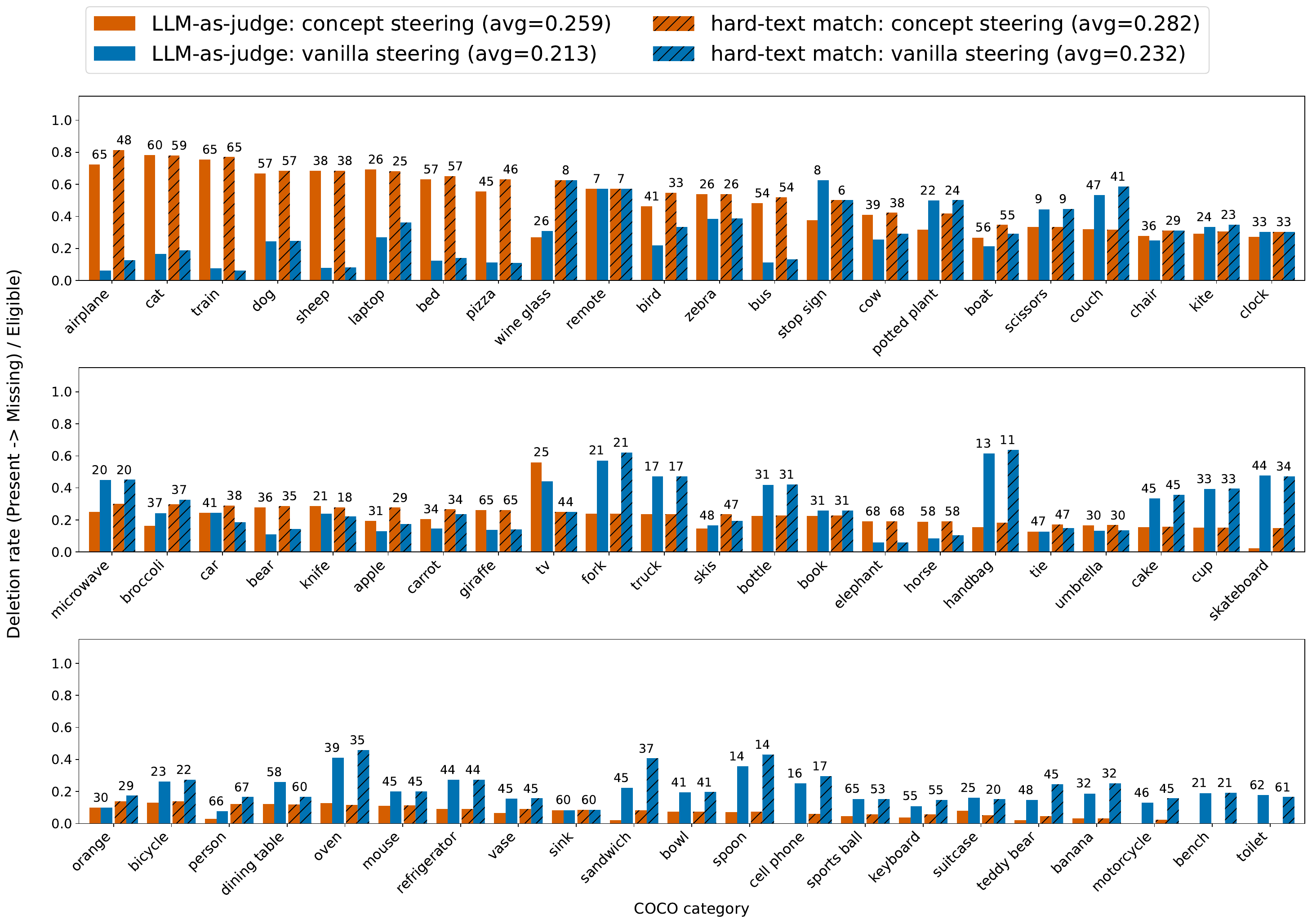}
\caption{\textbf{Quantitative Steering Evaluation.} For every COCO category (split to three subplots) we evaluate the success rate (y-axis) of the category not appearing on the output caption, when performing concept-based steering with \ours (orange bars) or CLIP-based steering (baseline; blue bars). We observe that the performance varies across categories, with \ours's automated concept-based steering being more successful on average by 5\%. }
\label{fig:steer-quant-supp}
\vspace{-0.8in}
\end{figure*}

\begin{figure*}[h!]
\centering
\vspace{-0.3in}
\includegraphics[width=\linewidth]{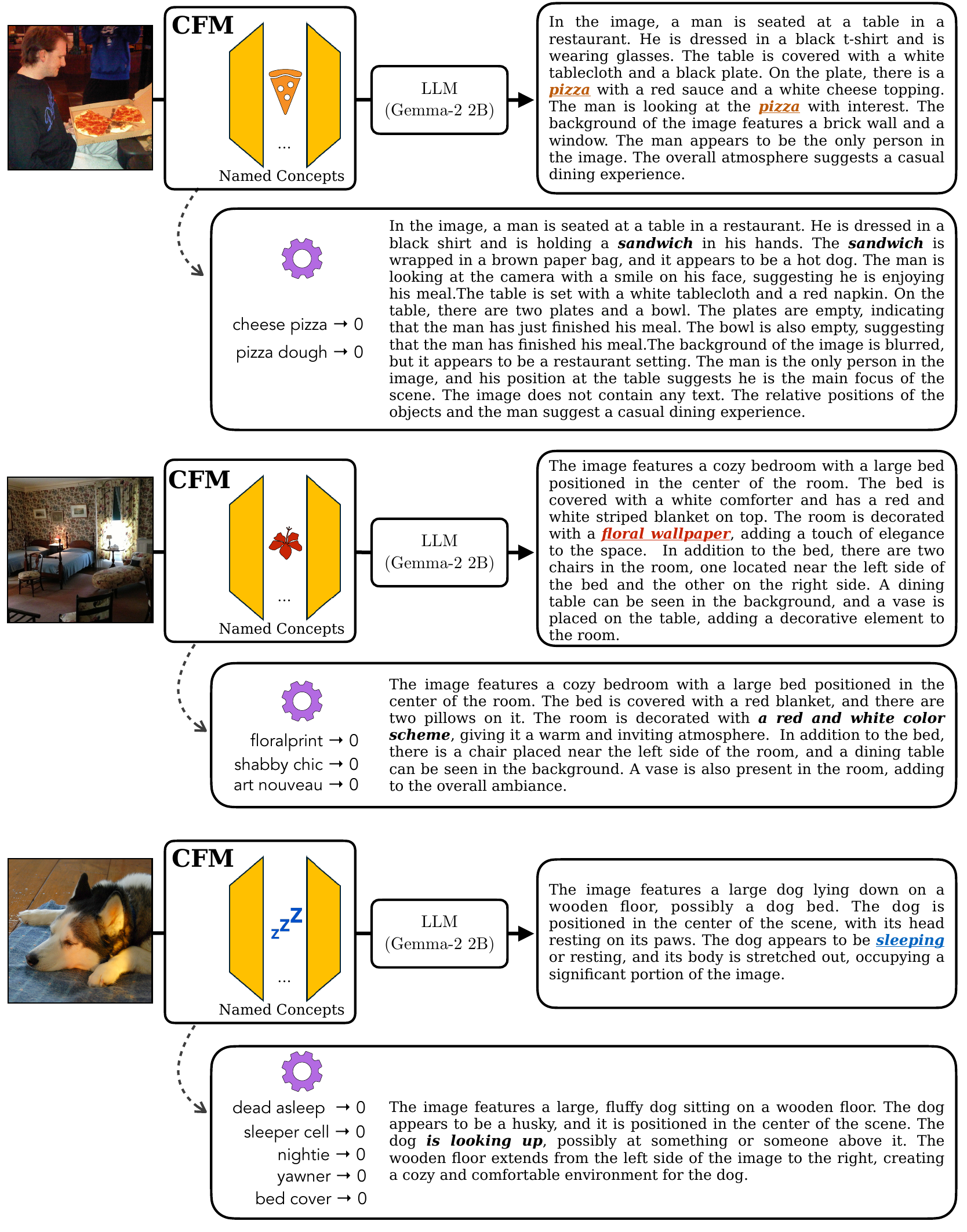}
\caption{\textbf{Steering with \ours.} In each example (major rows), we steer the output caption through intervention on the concept activations (left), \eg on the first row the pizza-related concepts are disabled. For each case we asked Google-Gemini to find the relevant concepts among the activated names. We observe that the intervention is effective in steering the output caption, highlighting the utility of a concept-based bottleneck representation and the high quality of the concept naming.}
\label{fig:steer-supplement}
\vspace{-0.6in}
\end{figure*}

In this section we extend the downstream captioning steering results from \cref{fig:captioning} with quantitative evaluation and additional qualitative samples. In \cref{fig:steer-quant-supp} we target every COCO class category as an object that we would like to remove from the output caption. In \cref{fig:steer-supplement}, we show additional samples from the COCO dataset~\cite{lin2014microsoft}, and report the model captions before and after steering. In both \cref{fig:steer-quant-supp,fig:steer-supplement} the steered concepts were selected by asking Google-Gemini to search through activated concept names for relevant names.

\myparagraph{Quantitative Evaluation.} In \cref{fig:steer-quant-supp}, we target every COCO class category with 80 samples per class and aim to remove the main category from the output caption while quantitatively evaluating the success rate. For this, we compare the initial caption of the LLM (before any steering) with our automated concept-based steering as well as a CLIP-based steering on the target word itself as a baseline. To check for presence of words in the output caption, we evaluate using both text matching (solid bars) and LLM-as-judge (hatched bars). We measure the success rate (y-axis) only on images where the original caption explicitly mentioned the object (the reported number on top of every bar).

In \cref{fig:steer-quant-supp}, we observe that the success rate of concept steering for both the baseline (blue bars) and our concept-based steering (orange bars) significantly differs across categories. Across many categories, CFM outperformed the baseline, e.g. ‘train’ (76.9\% vs. 6.2\%) and ‘cat’ (78.0\% vs. 18.6\%). There were also failure cases such as ‘fork’ (24\% vs. 61.9\%). Averaged across all images, our automated concept-level steering outperformed the baseline by 5 percentage points.

\myparagraph{Qualitative Evaluation.} We extend the qualitative results for captioning and steering shown in the main
paper with additional samples. In \cref{fig:steer-supplement}, we show test samples from COCO dataset \cite{lin2014microsoft}, and report the model captions before and after steering. We observe that CFM is able to successfully provide concept-level steering and change the downstream output caption of the LLM.

%% file: sec/X_suppl_cover.tex
\clearpage
\appendix
\crefalias{section}{appendix}
% \crefname{subsection}{App.}{Apps.}
% \Crefname{subsection}{App.}{Apps.}
% \setcounter{page}{1}
% 

\renewcommand\thesection{\Alph{section}}
\renewcommand{\theHsection}{appendix.\Alph{section}}
\numberwithin{equation}{section}
\numberwithin{figure}{section}
\numberwithin{table}{section}
\renewcommand{\thefigure}{\thesection\arabic{figure}}
\renewcommand{\thetable}{\thesection\arabic{table}}
\crefname{appendix}{Sec.}{Secs.}

{\onecolumn 
{\begin{center}
\Large\bf
{\ours: Language-aligned Concept Foundation
Model for Vision}\\[1em]
\large
Appendix
\end{center}
}
\newcommand{\additem}[2]{%
\item[\textbf{(\ref{#1})}] 
    \textbf{#2} \dotfill\makebox{\textbf{\pageref{#1}}
    }
}

\newcommand{\myindent}{.5em}
\newcommand{\addsubitem}[2]{%
\vspace{.5em}
    \textbf{(\ref{#1})}
        \hspace{\myindent} #2 \\    
}

\newcommand{\adddescription}[1]{\vspace{.1em}
\begin{adjustwidth}{0cm}{0cm}
#1
\end{adjustwidth}
}
\setlist[itemize]{noitemsep,leftmargin=*,topsep=0em} %
\setlist[enumerate]{noitemsep,leftmargin=*,topsep=0em}

\noindent In this appendix, we provide additional details on \ours, covering model training, experimental metrics, user study design and analysis, and additional qualitative results.
In particular, \Cref{app:sec:experiment_details} is dedicated to evaluation details, where we formally define the interpretability metrics reported in \cref{fig:interpret_local} in the main manuscript, discuss details of how we conducted the user study in Amazon MTurk, provide additional analysis of results stratified by concept activations frequency and results for concept naming quality, provide details about automated evaluations including prompts, and provide details about the open-vocabulary segmentation inference procedure.

In \Cref{app:sec:training}, we provide all necessary technical details to reproduce our results, including formal definitions of CLIP-DINOiser, its training objective, how Matryoshka SAEs were trained including the choice of hyperparameters, and formal details on hierarchy discovery and concept naming used in \ours. Lastly we provide training details for the classification, image captioning tasks, and the open-vocabulary segmentation tasks.

In \Cref{app:sec:results}, we provide additional results spanning all aspects of \ours, including concept relations, spatial grounding of concepts, and downstream tasks. We \textbf{highly encourage the reader to take a look at these results}. These are curated examples to show the diversity and granularity of concepts as well as concept hierarchies that \ours can find, but also to show the subtle limitations when it comes to naming fine-grained concepts. The following table of contents allows for fast access to the individual sections.

\vspace{0.3in}

\begin{adjustwidth}{1.2cm}{1.2cm}
\begin{enumerate}[label={({\arabic*})}, topsep=1em, itemsep=.2em]

    \additem{app:sec:experiment_details}{Evaluation Details}\\[0.4em]
        \addsubitem{app:sec:experimental_details:interpretability_metrics}{Quantitative Interpretability Metrics}\
        \addsubitem{app:sec:experimental_details:user_study}{User Study}\
        \addsubitem{app:sec:experimental_details:relationship_judge}{Relationship Evaluation Using an Automated Judge}\
        \addsubitem{app:sec:experimental_details:naming_judge}{Naming Evaluation Using an Automated Judge}\
        \addsubitem{app:sec:experimental_details:ovs}{Open-Vocabulary Segmentation Inference}\
    \additem{app:sec:training}{Training Details}\\[0.4em]
        \addsubitem{app:sec:dinoiser_training}{CLIP-DINOiser Training}\
        \addsubitem{app:sec:sae_training}{Matryoshka SAE Training}\
        \addsubitem{app:sec:family_training}{Hierarchy Discovery}\
        \addsubitem{app:sec:naming_details}{Concept Naming}\
        \addsubitem{app:sec:probe_training}{Probing for Classification}\
        \addsubitem{app:sec:captioning}{Image Captioning}\
        \addsubitem{app:sec:Open-Vocabulary segmentation}{Open-Vocabulary Segmentation}\
        \addsubitem{app:sec:hyperparameter_ablation}{Hyperparameter and Architectural Ablation}\
        
    \additem{app:sec:results}{Additional Results}\\[0.4em]
        \addsubitem{app:sec:families-viz}{Concept Hierarchies}\
        \addsubitem{app:sec:rand-viz}{Concept Consistency Visualization}\
        \addsubitem{app:sec:namecompare}{Concept Name Comparison}\
        \addsubitem{app:sec:polysemanticity_evaluation}{Polysemanticity Evaluation}\
        \addsubitem{app:sec:comparison_to_post_hoc_attribution}{Concept Consistency Evaluation}\
        \addsubitem{app:sec:classification-viz}{Interpretable Classification}\
        \addsubitem{app:sec:segmentation-viz}{Interpretable Open Vocabulary Segmentation}\
        \addsubitem{app:sec:captioning-viz}{Steerable Captioning}\
        
\end{enumerate}
\end{adjustwidth}
}